\documentclass[10pt,twocolumn,letterpaper]{article}

\usepackage[pagenumbers]{cvpr} %

\makeatletter
\@namedef{ver@everyshi.sty}{}
\makeatother
\usepackage{tikz}

\usepackage{graphicx}
\usepackage{amsmath}
\usepackage{amssymb}
\usepackage{booktabs}
\usepackage{algorithm}
\usepackage{algorithmicx}
\usepackage{algpseudocode}

\usepackage[pagebackref,breaklinks,colorlinks]{hyperref}

\usepackage[capitalize]{cleveref}
\crefname{section}{Sec.}{Secs.}
\Crefname{section}{Section}{Sections}
\Crefname{table}{Table}{Tables}
\crefname{table}{Tab.}{Tabs.}

\def\cvprPaperID{11775} %
\def\confName{CVPR}
\def\confYear{2023}

\usepackage{latexsym,eucal,bbm,color}
\usepackage{url}
\usepackage{comment}
\usepackage{float}
\usepackage{tcolorbox}
\usepackage{booktabs}
\usepackage{blindtext}
\usepackage{multirow}
\usepackage{booktabs}
\usepackage{array}
\usepackage{soul}
 
\usepackage[normalem]{ulem}
\usepackage{stackengine}
\usepackage{parskip}
\usepackage{bm}
\usepackage{balance}
\usepackage{comment}
\usepackage{soul}
\usepackage{pgffor}

\usepackage{caption}

\def \bx{\boldsymbol{x}}

\def \bz{\boldsymbol{z}}

\def \bv{\boldsymbol{v}}

\def \bS{\boldsymbol{S}}

\def \bv{\boldsymbol{v}}
\def \bb{\boldsymbol{b}}

\def \bq{\boldsymbol{q}}

\def \bw{\boldsymbol{w}}

\def \bA{\boldsymbol{A}}

\def \bW{\boldsymbol{W}}

\def\R{{\mathbb R}}

\def\Indic{\mathbbm{1}}

\newcommand{\diag}{\text{diag}}

\usepackage{listings}

\definecolor{codegreen}{rgb}{0,0.6,0}
\definecolor{codegray}{rgb}{0.5,0.5,0.5}
\definecolor{codepurple}{rgb}{0.58,0,0.82}
\definecolor{backcolour}{rgb}{0.95,0.95,0.92}

\lstdefinestyle{mystyle}{
  backgroundcolor=\color{backcolour}, commentstyle=\color{codegreen},
  keywordstyle=\color{magenta},
  numberstyle=\tiny\color{codegray},
  stringstyle=\color{codepurple},
  basicstyle=\ttfamily\footnotesize,
  breakatwhitespace=false,         
  breaklines=true,                 
  captionpos=b,                    
  keepspaces=true,                 
  numbers=left,                    
  numbersep=5pt,                  
  showspaces=false,                
  showstringspaces=false,
  showtabs=false,                  
  tabsize=2
}

\begin{document}

\title{SplineCam: Exact Visualization and Characterization of Deep \\ Network Geometry and Decision Boundaries}

\author{Ahmed Imtiaz Humayun\\
Rice University\\
{\tt\small imtiaz@rice.edu}
\and
Randall Balestriero\\
Meta AI, FAIR\\
{\tt\small rbalestriero@fb.com}
\and
Guha Balakrishnan\\
Rice University\\
{\tt\small guha@rice.edu}
\and
Richard Baraniuk\\
Rice University\\
{\tt\small richb@rice.edu}
}
\maketitle

\begin{abstract}
   Current Deep Network (DN) visualization and interpretability methods rely heavily on data space visualizations such as scoring which dimensions of the data are responsible for their associated prediction or generating new data features or samples that best match a given DN unit or representation. In this paper, we go one step further by developing the first provably exact method for computing the geometry of a DN's mapping -- including its decision boundary -- over a specified region of the data space. 
   By leveraging the theory of Continuous Piece-Wise Linear (CPWL) spline DNs, \textbf{SplineCam} exactly computes a DN's geometry without resorting to approximations such as sampling or architecture simplification. 
   SplineCam applies to any DN architecture based on CPWL activation nonlinearities, including (leaky) ReLU, absolute value, maxout, and max-pooling and can also be applied to regression DNs such as implicit neural representations. 
   Beyond decision boundary visualization and characterization, SplineCam enables one to compare architectures, measure generalizability, and sample from the decision boundary on or off the data manifold.
   Project website: \href{https://bit.ly/splinecam}{bit.ly/splinecam}.
\end{abstract}

\section{Introduction}
\label{sec:intro}

Deep learning and in particular Deep Networks (DNs) have redefined the landscape of machine learning and pattern recognition \cite{lecun2015deep}. Although current DNs employ a variety of techniques that improve their performance, their core operation remains unchanged, primarily consisting of sequentially mapping an input vector $\bx$ to a sequence of $L$ {\em feature maps} $\bz^\ell$, $\ell=1,\dots,L$ by successively applying simple nonlinear transformations,
as in
\begin{equation}
    \bz^{\ell}= \sigma \left(\bW^{\ell}\bz^{\ell-1}+\bb^{\ell}\right), 
    \quad \ell=1,\dots,L
    \label{eq:no_BN}
\end{equation}
starting with $\bz^0=\bx$. Here $\bW^\ell$ and $\bb^\ell$ denotes the weight matrix and the bias vector for layer $\ell$, and $\sigma$ is an activation operator that applies an element-wise nonlinear activation function. One popular choice for $\sigma$ is the Rectified Linear Unit (ReLU) \cite{glorot2011deep} that takes the elementwise maximum between its entry and $0$.
The parametrization of $\bW^\ell,\bb^{\ell}$ controls the type of layer, e.g., circulant matrix for convolutional layer.

\begin{figure}[t!]
    \centering
    \includegraphics[width=.47\linewidth]{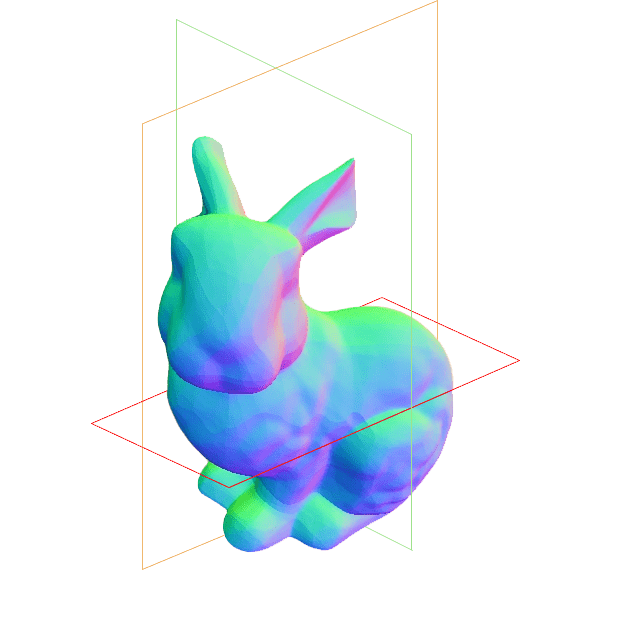}
    \includegraphics[width=.47\linewidth]{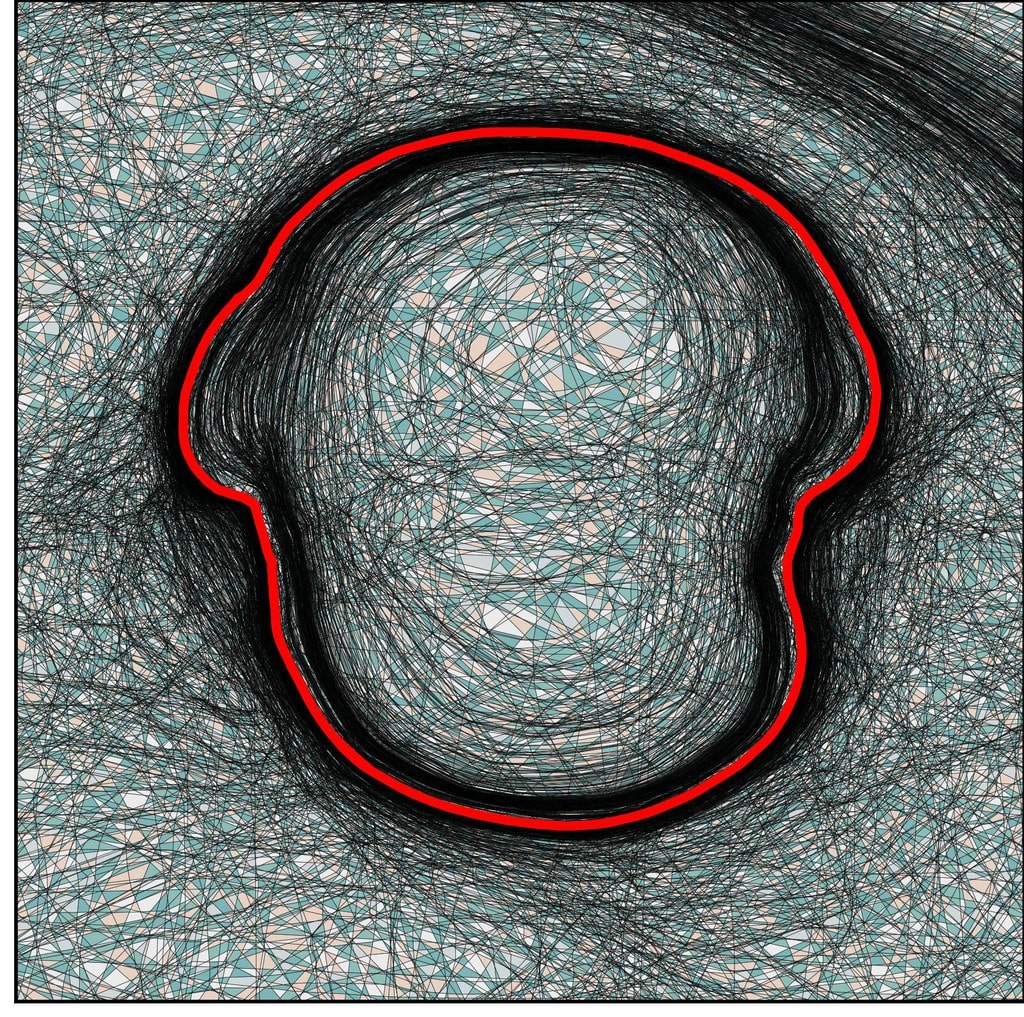}
    \includegraphics[width=.47\linewidth]{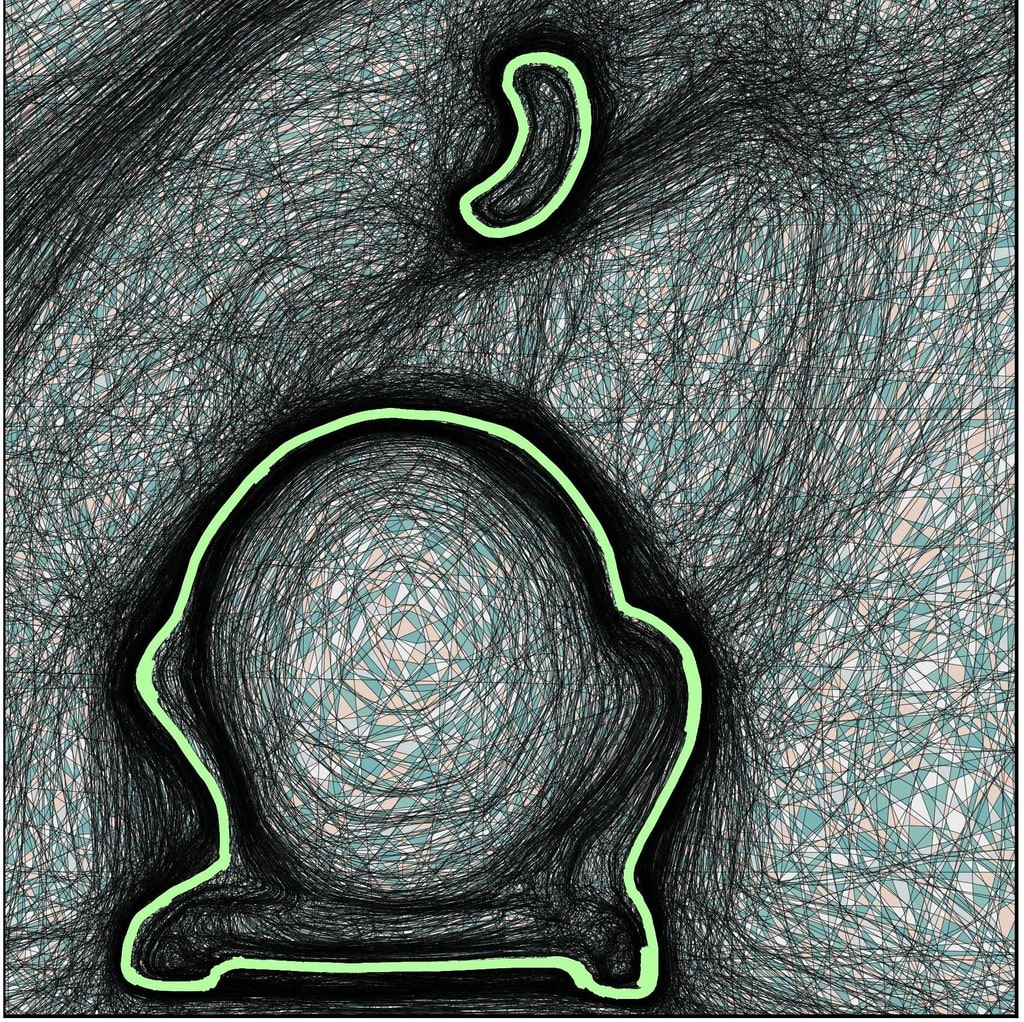}
    \includegraphics[width=.47\linewidth, angle=90]{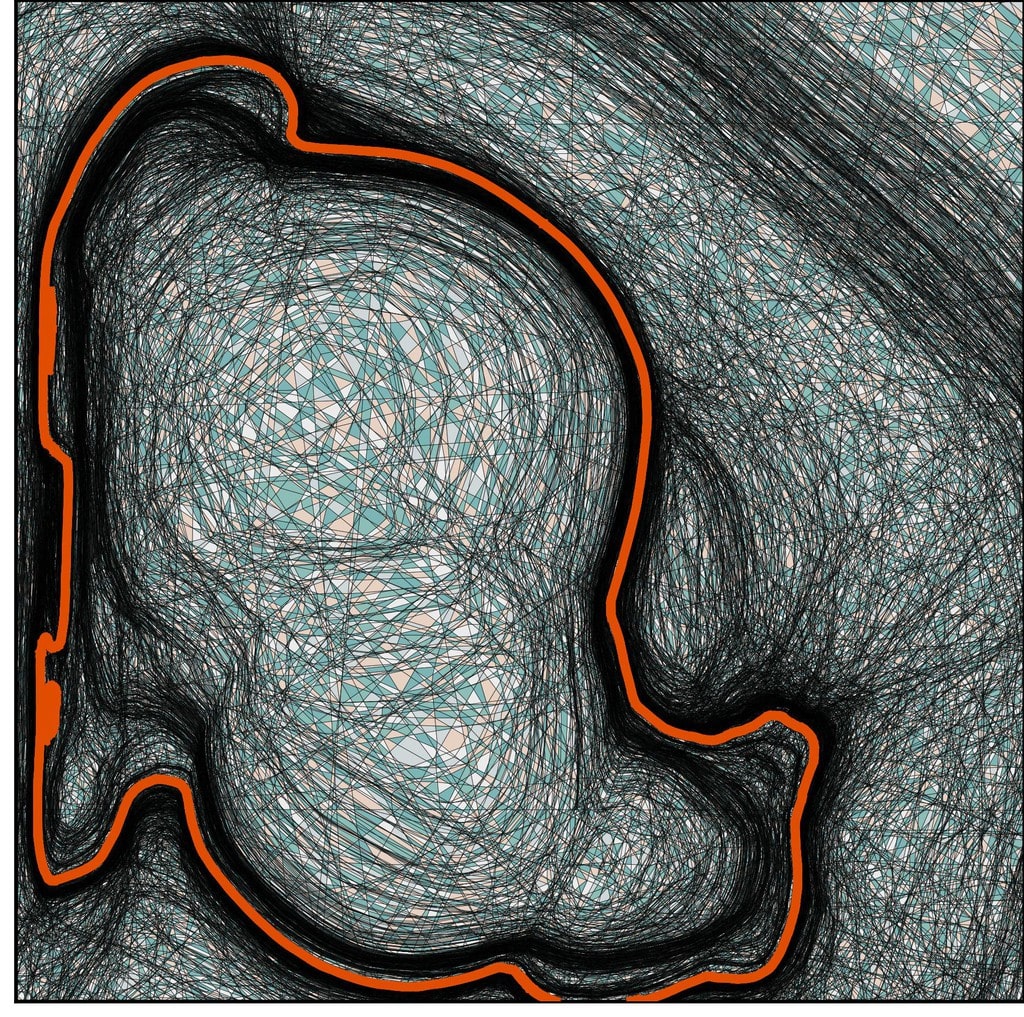}
    \caption{Exact visualization of the decision boundary and partition geometry of a 3D neural signed distance field (SDF). \textbf{(Top left)} Surface normals obtained from the learned signed distance field with annotations indicating slices used for visualization. For each of the slices, we can see the spline partition geometry of the learned SDF- each contiguous line represents a neuron, on either side of which it gets activated/deactivated. Neurons from different depths of the network create a partitioning of the input space into 'linear regions'. Here the colored lines represent the decision boundary learned by the SDF. Note that while the final neuron obtains the decision boundary, many neurons place their boundaries close to the ground truth surface to obtain the final SDF representation.}
    \label{fig:sdf3d}
\end{figure}

Interpreting the geometry of a DN is a nontrivial task since many different sets of parameters can lead to the same input-output mapping. One example is obtained by permuting the rows of $\bW^{\ell},\bb^\ell$ and the columns of $\bW^{\ell+1}$ for any two consecutive layers in a DN. Another example is to rescale $\bW^{\ell},\bb^\ell$ by some constant $\kappa$ and to rescale $\bW^{\ell+1}$ by $1/\kappa$ for a ReLU-DN \cite{phuong2020functional}; the list of such parameter manipulations preserving the underlying DN's function is an active area of research \cite{petzka2020notes}. Since one cannot trivially use the DN's parameters to describe its mapping, practitioners have relied on different solutions to interpret what has been learned by a model by looking at the activations instead of the weights of the network \cite{yosinski2015understanding,jalwana2021cameras}. Activation based interpretability methods however can be susceptible to feature adversarial attacks, i.e., adversarial attacks that don't cross the decision boundary but changes the activation \cite{ghorbani2019interpretation}. Some alternative empirical methods for model interpretation, therefore rely on sampling the decision boundary or finding the point on a model's decision boundary closest to a sample $\bx$ \cite{somepalli2022can}. Beyond interpretability, such methods find practical use in active learning \cite{locatelli2018adaptive} and adversarial robustness \cite{he2018decision}. In this setting, gradient updates are performed from an initial guess for $\bx$ based on an objective function that reaches its minimum whenever its argument lies on the model's decision boundary. While alternative and more efficient solutions have been developed, most of the progress in this direction has focused on providing more optimized losses and sampling strategies \cite{somepalli2022can,he2018decision}. 
In short, {\em there doesn't exist an exact (up to machine precision) method to compute the decision boundary of a DN.}

In this paper, we focus on DNs employing Continuous Piece-Wise Linear (CPWL) activation functions $\sigma$, such as the (leaky-)ReLU, absolute value, and max-pooling. In this setting, the entire DN itself becomes a CPWL operator, i.e., its mapping is affine within regions of a partition of its domain.
There has been previous studies dedicated to estimating the partition of such CPWL DNs and bridging empirical findings with interpretability. For example, Raghu et al. \cite{raghu2017expressive} shows that the partition density provides measures of DN expressivity, Hanin et al. \cite{hanin2019complexity} connects the DN partition density with the complexity of the learned function, Jordan et al. \cite{NEURIPS2019_robustcert} approximates the DN partition to provide robustness certificates, Zhang et al. \cite{zhang2020empirical} interprets the impact of dropout with respect to DN partitions, Balestriero et al. \cite{balestriero2022batch} proposes to improve batch-normalization to further adapt DN partitions to the data geometry, Humayun et al. \cite{humayun2022polarity,humayun2022magnet} proposes methods to control pre-trained generative network output distributions by approximating the DN partition, Chen et al. \cite{mellor21nas} proposes a neural architecture search method based on partition statistics. Despite being successful, \textit{all these methods rely on approximation of the DN partition.}

We propose \textit{SplineCam}, a sampling-free method to compute the exact partition of a DN. Our method computes the partition on two-dimensional domains of the input space, easily scales with width and depth of DNs, can handle convolutional layers and skip connections, and can be scaled to discover numerous regions as opposed to previously existing methods.
Our method also allows local characterization of the input space based on partition statistics, and enables one to tractably and efficiently sample arbitrarily many samples that provably lie on a DN's decision boundary - opening new avenues for visualization and interpretability.  
We summarize our contributions as follows:
\begin{itemize}
    \item Development of a scalable enumeration method that, given a bounded 2D domain of a DN's input space, computes the DN's input space partition (aka, linear regions) and decision boundary.
    \item Development of {\bf SplineCam} that leverages our new enumeration method to directly visualize a DN's input space partition, compute partition statistics and sample from the decision boundary.
    \item Quantitative analysis that demonstrates the ability of SplineCam to characterize the DN and compare between architectural choices and training regimes.
\end{itemize}

The \textbf{SplineCam} library, and codes required to reproduce our results are provided in our Github\footnote{\texttt{https://bit.ly/splinecam-git}}. In Suppl.~Sec.~\ref{appendix:usage}, we also demonstrate the usage of SplineCAM with example code blocks.

\section{The Exact Geometry and Decision Boundary of Continuous Piece-Wise Linear Deep Networks}

The goal of this section is to first introduce basic notations and concepts associated with CPWL DNs (Sec.~\ref{sec:background}), and then develop our method that consists of building the exact DN input space partition, and the DN's decision boundary that lives on it (Sec.~\ref{sec:method}); empirical studies based on our method will be provided in Sec.~\ref{sec:experiments}.

\subsection{Deep Networks as Continuous Piece-Wise \\ Linear Operators}
\label{sec:background}

One of the most fundamental functional form for a nonlinear function emerges from polynomials, and in particular, spline operators. In all generality a spline is a mapping which has locally degree $D$ polynomials on each region $\omega$ of its input space partition $\Omega$, with the additional constraints that the first $D-1$ derivatives of those polynomials are continuous throughout the domain, i.e., imposing a smoothness constraint when moving from one region to any of its neighbor.
\begin{figure}
    \centering
    \includegraphics[width=\linewidth]{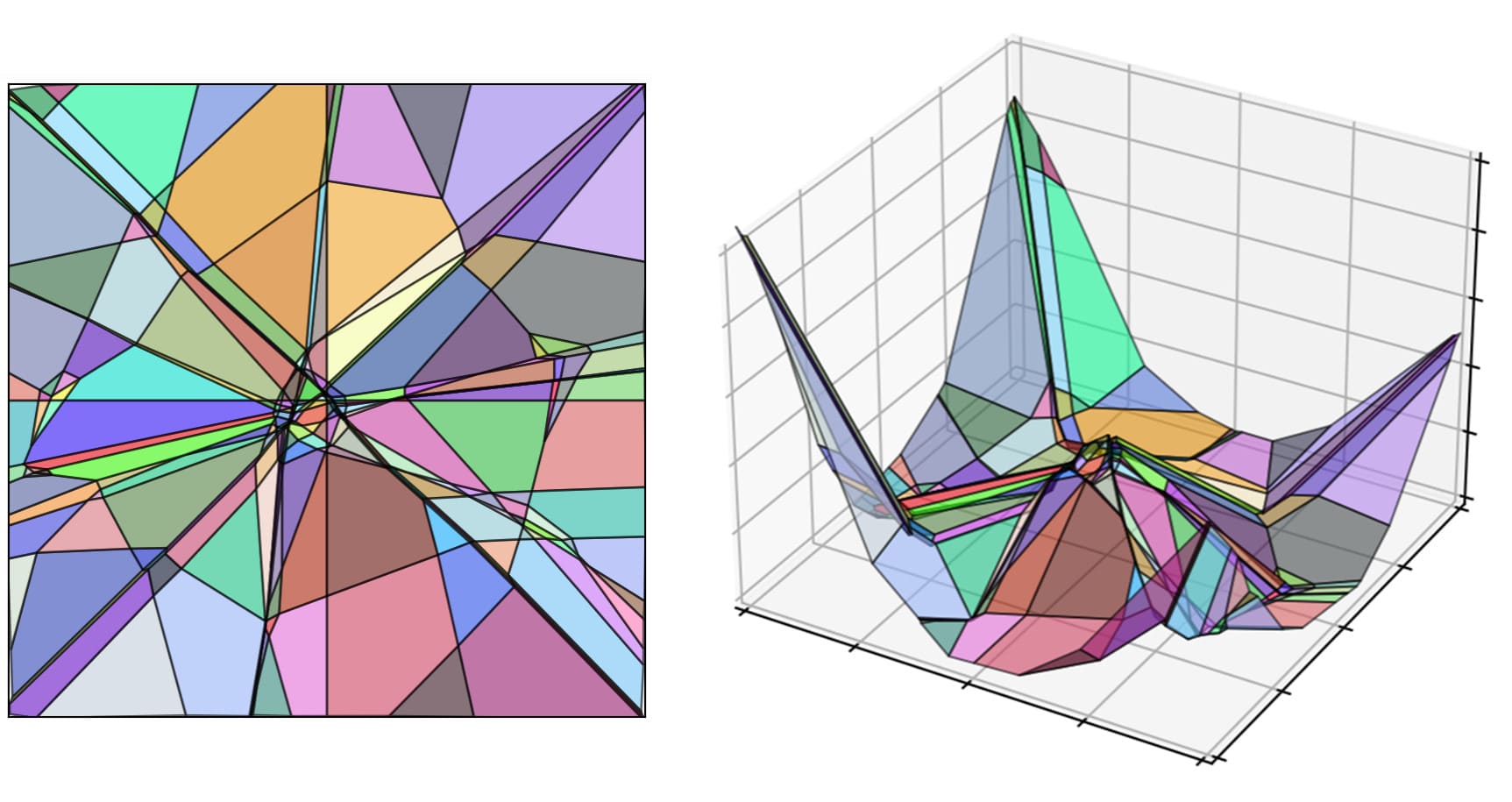}
    \caption{Visual depiction of Eq.~\ref{eq:CPA} with a toy affine spline mapping $S : \mathbb{R}^2 \rightarrow \mathbb{R}^3$. \textbf{Left.} Input space partition $\Omega$ made of multiple convex regions shown with different colors and with boundaries shown in black. \textbf{Right.} Affine spline image $Im(S)$ which is a continuous piecewise affine surface composed of the input space regions affinely transformed by the per-region affine mappings. Colors maintain correspondence from the left to the right.}
    \label{fig:2dexact}
\end{figure}

More formally and for the context of DNs we will particularly focus on affine splines, i.e., spline operators with $D=1$ and only constrained to enforce continuity throughout the domain. 

Let $\bS$ be a Deep Network (DN) with $L$ layers and parameters $\{\bW^\ell,\bb^\ell\}_{\ell=1}^{L}$. Whenever $\bS$ employs continuous piecewise affine (CPA) activation $\sigma$ at each layer, i.e., layer $\ell$ outputs are given by Eq.~\ref{eq:no_BN},
with $\bz^0$ being an input $\bx \in \mathbb{R}^S$.

\textbf{Lemma 1.} \textit{The layer $1$ to $\ell$ composition of a DN $\bS$, denoted as $\bS^{\ell}$ with output space $\mathbb{R}^\ell$, can be expressed as}

\begin{equation} 
    \bS^{\ell}(\bx)=\sum_{\omega \in \Omega}\left(\bA_{\omega}^\ell\bx+\bb_{\omega}^\ell\right)\Indic_{\{\bx \in \omega\}},\label{eq:CPA}    
\end{equation}

\textit{with indicator function $\Indic_{\{.\}}$ and per-region affine parameters given by,}

\vspace{-1em}
\begin{align}
\bA_{\omega}^\ell\hspace{-0.1cm}=& \prod_{i=1}^{\ell}\diag\left(\bq^{i}_\omega\right)\bW^{i},\\[0em]  \bb_{\omega}^\ell\hspace{-0.1cm}=&\diag\left(\bq^{\ell}_\omega\right)\bb^{\ell} \hspace{-0.1cm}+\hspace{-0.1cm}\sum_{i=1}^{\ell-1}\hspace{-0.1cm}\left( \prod_{j=i+1}^{\ell}\hspace{-0.1cm}\diag\left(\bq^{j}_\omega\right)\hspace{-0.1cm}\bW^{j}\hspace{-0.1cm}\right) \hspace{-0.1cm}\diag\left(\bq^{i}_\omega\right)\bb^{i}.
\end{align}

Here, $\bq^\ell_\omega$ is the point-wise derivative of $\sigma$ at pre-activation $\bW^\ell\bz^{\ell-1} + \bb^\ell$, and $\diag(.)$ operator given a vector argument creates a matrix with the vector values along the diagonal. As a consequence of Thm. 1 from Balestriero et al. \cite{balestriero2018spline}, $\bq^\ell_\omega$ is unique for any region $\omega \in \Omega$.

Such formulations of DNs have previously been employed to make theoretical studies amenable to actual DNs without any simplification, while leveraging the rich literature on spline theory, e.g., in approximation theory \cite{cheney2009course}, optimal control \cite{egerstedt2009control}, statistics \cite{fantuzzi2002identification} and related fields.

\subsection{Exact Computation of Their Partition and \\ Decision Boundary}
\label{sec:method}

Suppose, $\bw_i^\ell$,$\bb_i^\ell$ are the $i$-th rows of $\bW^\ell,\bb^\ell$. The following lemma provides us a framework to back-project to $\R^S$ a hyperplane $h_i^\ell \in \mathbb{R}^{\ell-1}$ from layer $\ell$ with parameters $\bw_i^\ell,\bb_i^\ell$, expressed as,
\begin{equation}
    h_i^\ell \triangleq \{\bz \in \mathbb{R}^{\ell-1} : \langle \bw_i^\ell , \bz \rangle + \bb_i^\ell = 0\ \}. \label{eq:hyperplane}
\end{equation}

\begin{algorithm}[t]
\caption{Find Partitions}
\begin{algorithmic}
\Require 2-polytope $P$ and hyperplanes $\mathcal{H} \in \mathbb{R}^2$, s.t., $\mathcal{H} \cap P \neq \emptyset$.
\Ensure $G, C$. $G = (E,N)$, where $E$ are edges and $N$ are nodes of the graph $G$. $C$ are cycles/cells/faces of $G$.
\\

\noindent
\textbf{Step 1:} Solve for $N$ = $\{h_i \cap (h_j \cup P_e)~\forall h_i,h_j \in \mathcal{H} : j \neq i \}$, where $P_e$ are edges of $P$. 
\\

\noindent
\textbf{Step 2:} For each $h_i$, sort $\{h_i \cap (h_j \cup P_e)~\forall h_j \in \mathcal{H} : j \neq i \}$ and connect in sorted sequence to obtain edges $E$. 
\\

\noindent
\textbf{Step 3:} Obtain set of cycles $C$ from graph $G$ via Alg.~\ref{alg:findCycles}.

\end{algorithmic}
\label{alg:findPartitions}
\end{algorithm}

\begin{figure*}[!t]
    \centering
    \includegraphics[width=\linewidth]{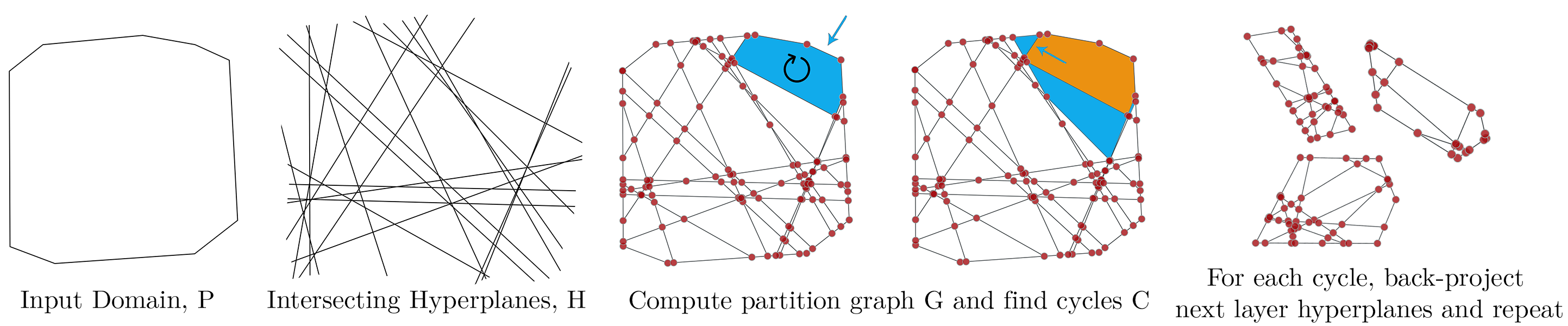}
    \caption{
    Given an input domain $P$ and a set of hyperplanes $\mathcal{H}$, SplineCam first produces a graph $G$ using all the edge and hyperplane intersections (as in Alg.~\ref{alg:findPartitions}). To find all the cycles in the graph, we select a boundary edge $e_s$ (blue arrow), do a breadth-first search (BFS) to find the shortest path through the graph between vertices of $e_s$ and obtain the corresponding cycle (blue). The edges obtained via BFS are enqueued and the search is repeated for each.
    Each non-boundary edge is allowed to be traversed twice, once from either direction (see Alg.~\ref{alg:findCycles}). Once new regions are found, we back-project deeper layer hyperplanes, compute partition graphs, and repeat.}
    \label{fig:region_find}
\end{figure*}

\textbf{Lemma 2.} \textit{Given a hyperplane $h_i^\ell \in \mathbb{R}^{\ell-1}$,
it can be projected onto the tangent space of region $\omega \in \Omega$ in $\mathbb{R}^S$
as,}
\begin{equation}
    proj_\omega(h_i^\ell) = \{ \bx \in  \mathbb{R}^S : \langle \bw_i^\ell, \bA_\omega^{\ell-1} \bx + \bb_\omega^{\ell-1} \rangle + \bb_i^\ell = 0\}. \label{eq:backproject}
\end{equation}

The proof of Lem.~2 is direct since $\bz^{\ell}_i = \langle \bw_i^\ell, \bA_\omega^{\ell-1} \bx + \bb_\omega^{\ell-1} \rangle + \bb_i^\ell$, with $\bz_i^\ell$ the i-th element of layer $\ell$ activation.

\textbf{Theorem 1.} \textit{Let, $\bS$ is a binary classifier DN therefore with a single output neuron. In $\mathbb{R}^{L-1}$, the decision boundary is the hyperplane $h^{L}_1$.
The decision boundary in $\mathbb{R}^S$ is therefore $\cup_{\omega \in \Omega} \{ proj_\omega(h_1^L) \cap \omega \}$.}

Thm.~1 can be proven by repeatedly applying Lem.~2 to back-project the hyperplane $h_1^L$ for all $\omega \in \Omega$. 
While the above is general, we want to compute $\Omega$ on a 2-polytope $P \in \mathbb{R}^S$ for tractability \cite{agarwal1990partitioning} and ease of visualization. 

\textbf{SplineCam.} Let's denote the partition in the input space formed by the composition of layers $1$ to $\ell$ as $\Omega^\ell$. Using Alg.~\ref{alg:findPartitions}, SplineCam starts by partitioning $P$ into $\Omega^1$ via  hyperplanes $h_i^1$ from layer $\ell=1$, the first layer. Then for each $\omega \in \Omega^1$, we use Lem.~1 and Lem.~2 to obtain $proj_\omega(h_i^2)$ for layer two. Therefore, we use Alg.~\ref{alg:findPartitions} on each region to obtain a finer partition. For a given polytopal region $\omega$, the target of Alg.~\ref{alg:findPartitions} is to first compute the graph $G$ formed via intersections between the edges of $\omega$ and a set of hyperplanes $\mathcal{H}$. Next, via Alg.~\ref{alg:findCycles}, it finds the unique cycles in $G$ (also referred to as circuits or faces for a planar graph). By repeating  Alg.~\ref{alg:findPartitions} for each region $\omega \in \Omega^1$, we can obtain $\Omega^2$. We repeat this process for all layers up to $\ell=L$ to obtain the final partition.
We have provided pseudocode for the search algorithm in python script in Suppl.~\ref{appendix:usage} List.~3.

\begin{algorithm}[t]
\caption{Find cycles}
\begin{algorithmic}
    
\Require{$G = (E,N)$ an undirected graph, $P_e \subset E$ boundary edges and starting edge $e_s \subset P_e$.}
\Ensure{$C$ cycles.}

\State{Initialize $C = \emptyset, E_q = \emptyset$}

\State{$G' = $bidirectional$(G)$ }\Comment{connect edges both ways}

\State{APPEND $E_q \leftarrow e_s$} \Comment{append to end of queue}

\State{REMOVE $e_s$ from $G'$}

\While{$E_q \neq \emptyset$}

\State{POP $e \leftarrow E_q$} \Comment{get from top of queue}

\State {REMOVE $e'$ from $G'$} \\ \Comment{$e'$ is $e$ with its direction inverted}




\State{$E_c=$ bfs$(G',v_e,v_s)$} \\ \Comment{$v_s,v_e$ are start and end vertices of $e$} \\ \Comment{$E_c$ are edges forming shortest path from $v_e$ to $v_s$}

\State{APPEND $E_q \leftarrow e_c', \forall e_c \in E_c$ if $e_c \notin P_e$}

\State {REMOVE $e_c$ from $G'$, $\forall e_c \in E_c$}

\State{REMOVE $e_c'$ from $G'$, $\forall e_c \in E_c$ s.t. $e_c \subset P_e$}

\State{APPEND $E_c \leftarrow e'$} \Comment{append edge to form cycle}

\State{APPEND $C \leftarrow E_c$}

\EndWhile

\end{algorithmic}
\label{alg:findCycles}
\end{algorithm}

\noindent
\textbf{Scalability and Complexity.} SplineCam is scalable, all the SplineCam operations can be vectorized except for the search algorithm, which finds cycles in a given graph $G$. For the vectorized operations, we can trade-off time complexity with space complexity by allocating more memory, preferably on GPU. For Alg.~\ref{alg:findCycles}, scaling requires distributing sets of $(\omega,\mathcal{H})$ across CPU threads.
Given a set of $n$ intersecting hyperplanes and a 2-polytope $P$, the operation of Alg.~\ref{alg:findPartitions} to find the partition reduces to an arrangement of lines problem. Therefore, the number of intersections, edges and cycles $\leq \mathcal{O}(n^2)$ \cite{agarwal1990partitioning}. As the number of hyperplanes $n\rightarrow \infty$, the expected number of edges per cycle $\approx \mathcal{O}(1)$. We can also see this in Table.~\ref{tab:my-table}, where we see that regardless of the architecture or training dataset, the average number of edges per region converges to $4$. Therefore, the average case complexity of Alg.~\ref{alg:findCycles} as $n \rightarrow \infty$ is also of the order $\mathcal{O}(n^2)$. In Suppl. Fig.~\ref{fig:complexity_singleMLPlayer}, we present the wall time required for SplineCam for a randomly initialized single layer MLP with variable width and input dimensionality. For an MLP with width $1000$, input dimensionality $8002$, and $8$M params, it takes SplineCam $134s$ to find $132K$ regions. We present in Suppl. Fig.~\ref{fig:mlp_partition} partition visualization for such a setting. For deeper networks, e.g., $138$M parameter VGG16 pre-trained on tinyImagenet, SplineCam takes $\sim7$ minutes to find 1K regions. We also provide statistics and visualizations for deeper networks, in Fig.~\ref{fig:tinyimagDA} and Suppl. Materials.

The methods closest to SplineCam in the literature are by Yuan et al. \cite{ijcai2022LP} that uses an exponential complexity linear programming based algorithm to compute the DN partitions and Gamba et al. \cite{gamba2022all} a method that computes the intersection of partition boundaries with one-dimensional lines connecting pairs of training samples. SplineCam computes the partition of 2D input domains and can compute the per region affine functions as well. \textit{SplineCam is the first exact method that is fast, scalable and computes the partition on 2D slices for a wide array of architectures.}


\begin{figure*}[h]
    \centering
    \begin{minipage}{0.135\linewidth}
    \centering
    Original image
    \end{minipage}
    \begin{minipage}{0.135\linewidth}
    \centering
    Layer 1
    \end{minipage}
    \begin{minipage}{0.135\linewidth}
    \centering
    Layer 2
    \end{minipage}
    \begin{minipage}{0.135\linewidth}
    \centering
    Layer 3
    \end{minipage}
    \begin{minipage}{0.135\linewidth}
    \centering
    Layer 4
    \end{minipage}
    \begin{minipage}{0.135\linewidth}
    \centering
    Layer 5
    \end{minipage}
    \begin{minipage}{0.135\linewidth}
    \centering
    Reconstruction
    \end{minipage}\\
    \includegraphics[width=\linewidth]{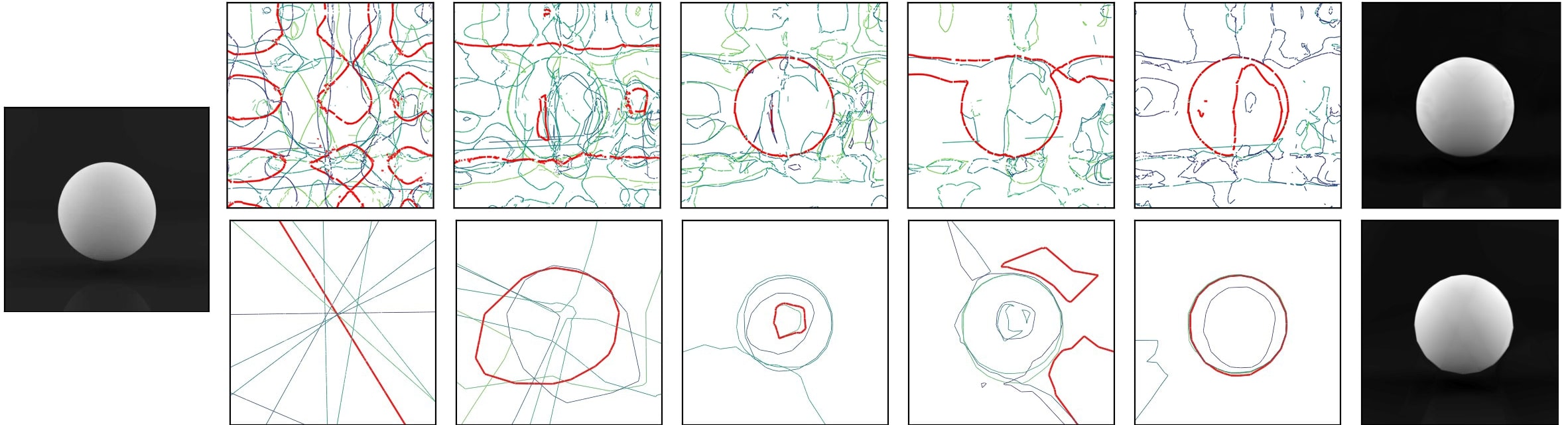}
    \caption{SplineCam visualizations of neurons from different layers of an MLP trained with \textbf{(top)} and without \textbf{(bottom)} periodic position encoding on a 2D image fitting task. All the neurons are visualized in the input space, color coded by the same color, and one neuron from each layer is highlighted in red. The trained MLP has a width of $10$ and depth of $5$ and has ReLU activations for every layer. For the positionally encoded (PE) network, boundaries of some neurons seem to be periodically repeating in the input domain, significantly increasing the number of unique $\omega$ where the ReLU is active. \textit{The increased weight sharing, i.e., same weights/neurons being used
    to represent/fit non-contiguous parts of the learned function, could be a possible reason for improved convergence of PE MLP \cite{nowlan1992simplifying}}.}
    \label{fig:posenc}
\end{figure*}

\begin{figure}[h]
    \centering
    \includegraphics[width=.3\linewidth]{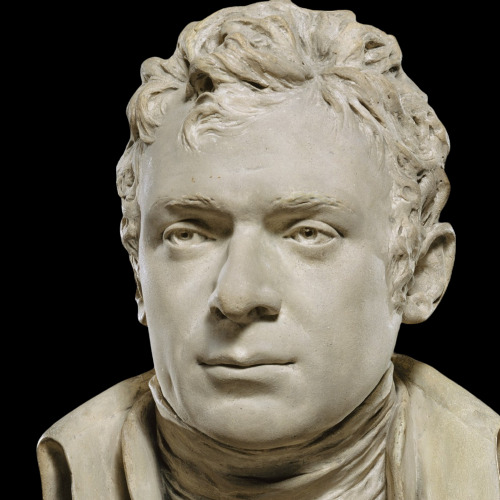}
    \includegraphics[width=.3\linewidth]{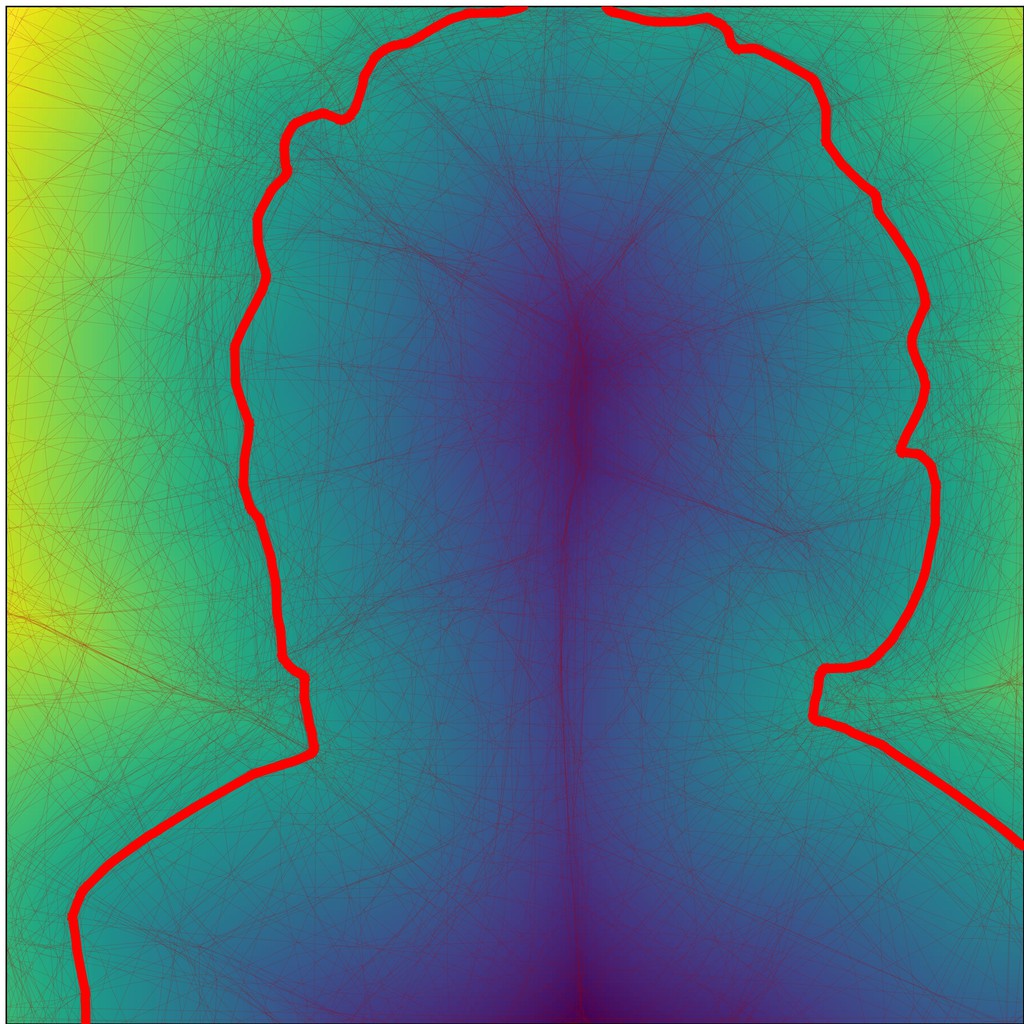}
    \includegraphics[width=.3\linewidth]{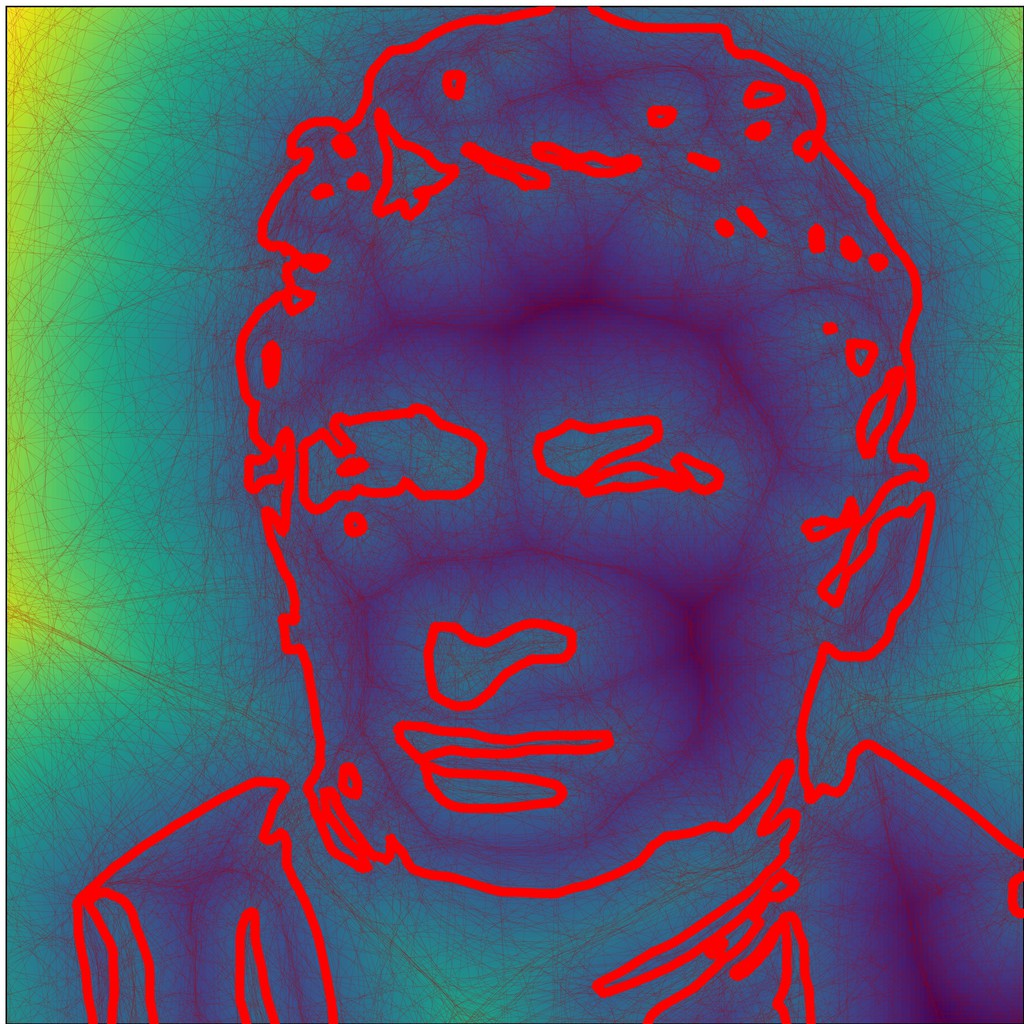}
    \includegraphics[width=.45\linewidth]{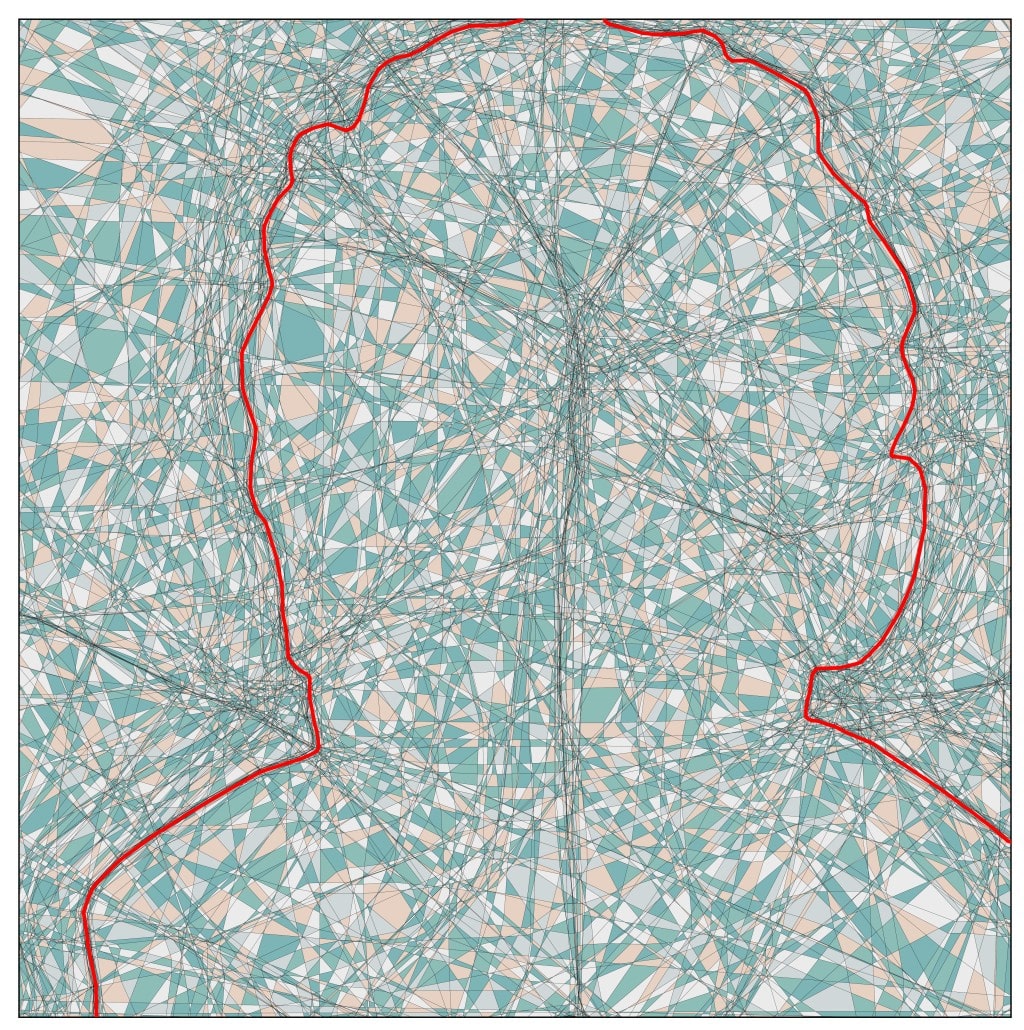}
    \includegraphics[width=.45\linewidth]{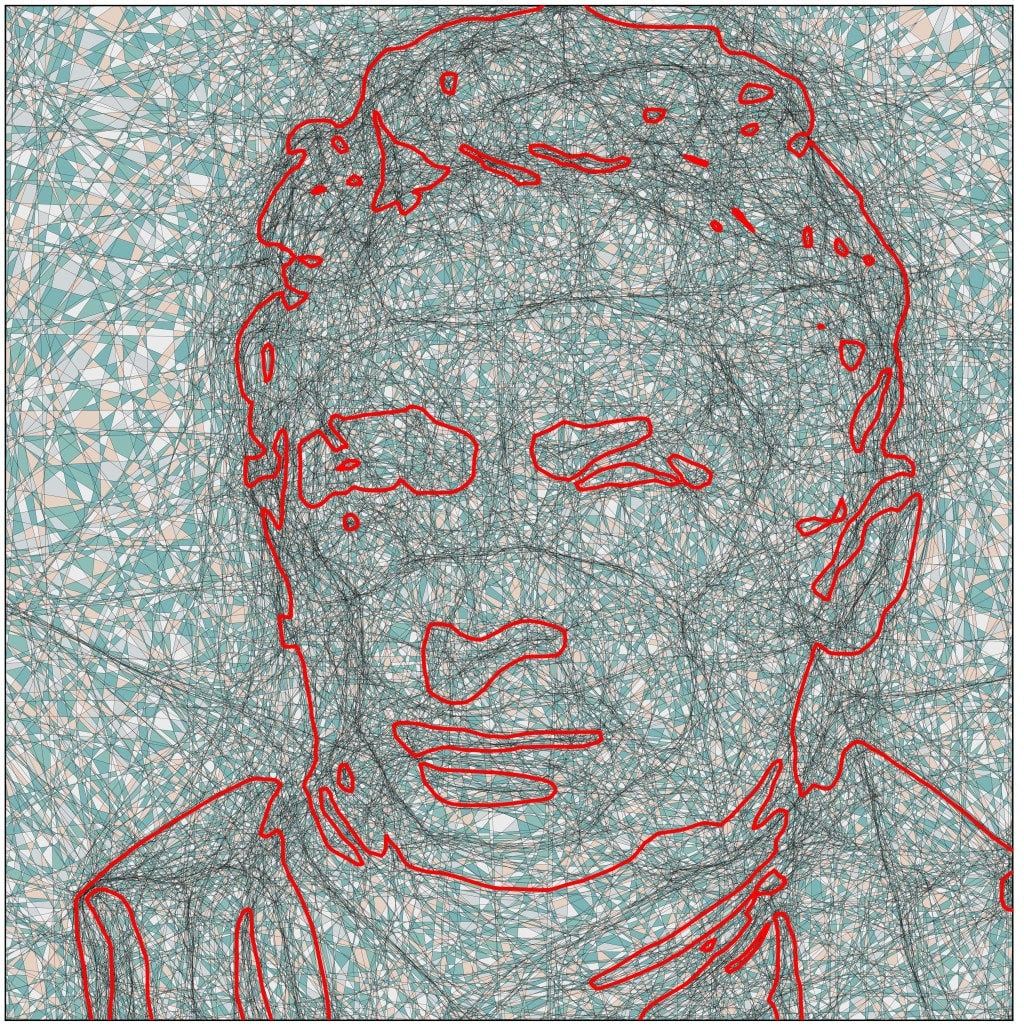}
    \caption{Visualization of the decision boundary and partition geometry of a 2D neural SDF with width $256$ and depth $6$. A single training image is thresholded at $0.01$ and $0.5$ to create two signed distance fields (\textbf{top-middle and top-right}), on which an MLP is trained. 
    We use SplineCam to obtain the analytical zero level set (decision boundary, in red) and also visualize the partition geometry \textbf{(bottom)}. Note that even with identical architecture, the partition density differs significantly based on the task complexity. \vspace{-1em}  }
    \label{fig:sdf2d}
\end{figure}

\section{Visualizing and Understanding Implicit Neural Representations}

We start our journey into the geometry of DNs by looking at implicit neural representations (INRs). INRs are generally multi-layer perceptrons (MLPs) that are trained to produce a continuous mapping from 1D/2D/3D signal coordinates to the intensity of the signal at those coordinates. They are pervasive in applications like 3D view synthesis \cite{mildenhall2021nerf} and inverse problems \cite{sun2021coil}. The low input dimensionality, and interpretability of ground truth functions make INRs a good setting to qualitatively validate SplineCam.
There also exists a lack of theoretical understanding for current INR practices \cite{yuce2022structured}. 
For example, while ReLU MLPs were primarily used in NeRF \cite{mildenhall2020nerf}- one of the most popular INR applications- the current practice has moved towards using periodic activations to encode of the input coordinates and following up with a ReLU MLP. In this section, we look at the effect of periodic encodings and visualize the geometry of the regions learned by INRs.

\subsection{Decision Boundary of Signed Distance \\ Functions} 

A signed distance function (SDF) is an implicit continuous representation of a surface or boundary; the output of an SDF is the signed distance of an input from the boundary represented by the function. The zero level set of an SDF therefore denotes the surface or boundary of the function. Training an INR as a signed distance function is essentially a regression task, where a ground truth distance field is fit by the model to implicitly learn a continuous boundary. We train a 2D and 3D SDF and visualize the analytical zero level set,  à la decision boundary, using our method, and provide the spline partitioning learned by the functions in Fig.~\ref{fig:sdf3d} and Fig.~\ref{fig:sdf2d}.

To train an INR as a 2D SDF, we take the image as in Fig.~\ref{fig:sdf2d} from the MetFaces \cite{karras2020analyzing} dataset and threshold it at $.001$ and $.05$ to create two binary images. Each binary image is used to create separate ground truth SDFs, one with a simpler boundary separating the background (ESDF) and another with a more convoluted boundary (HSDF). We train an identical ReLU-MLP architecture with width $256$ and depth $6$ on both ESDF and HSDF. In Fig.~\ref{fig:sdf2d} we present the analytical decision boundary of the SDF overlaid on the ground truth signed distance field for both ESDF and HSDF. While the network capacity remains the same for both, we can see that the spline partition of the two figures vary dramatically, with a finer partition and higher region density for the HSDF task compared to ESDF. For the harder HSDF task, creating significantly more regions allows the network to learn the curvature of the decision boundary. \textit{This indicates that harder tasks may utilize more of the network parameters compared to easier tasks.}

For the 3D SDF task, we train a leaky-ReLU-MLP with width $256$ depth $6$ on a Stanford Bunny SDF, and present in Fig.~\ref{fig:sdf3d} the normal map of the learned SDF, as well as the spline partition and decision boundary on three 2D slices, $\{x=0,y=0,z=0\}$. What is noticeable here is that, apart from the final layer hyperplane (final output neuron weights), many neurons from the deeper layers of the network also place their boundaries near the zero level set of the SDF. Meaning, while the decision boundary is denoted by the output neuron, there are multiple neurons that learn the surface boundary. \textit{This is indicative of a hierarchical nature of signal fitting by INRs.}

\subsection{The Effect of Positional Encoding on INRs}

\begin{table*}[t]
\centering
\caption{Statistics of the spline partitions formed by fully connected (MLP) and convolutional (CONV) neural networks. For each dataset, the same 2D slice and input domain is used to find the partition regions. Convolutional neural networks have a significantly higher number of regions with lower volume compared to MLPs even with less parameters.
}
\begin{tabular}{@{}ccrrrrr@{}}
\toprule
Architecture & Dataset & \multicolumn{1}{c}{Parameters} & \multicolumn{1}{c}{Avg. Reg. Vol} & \multicolumn{1}{c}{\begin{tabular}[c]{@{}c@{}}Avg. Number\\ of Vertices\end{tabular}} & \multicolumn{1}{c}{Ecc.} & \multicolumn{1}{c}{\begin{tabular}[c]{@{}c@{}}Number of\\ Regions\end{tabular}} \\ \midrule
\multirow{2}{*}{MLP} & MNIST & 44,860 & 3.144e-4 & 4 & 102e7 & 318 \\
 & Fashion-MNIST & 44,860 & 4.991e-4 & 4 & 36e7 & 1364 \\
\multirow{2}{*}{CONV} & MNIST & 39,780 & 1.134e-5 & 4 & 17e7 & 8814 \\
 & Fashion-MNIST & 39,780 & 3.54e-5 & 4 & 14e7 & 28222 \\ \bottomrule
\end{tabular}
\label{tab:my-table}
\end{table*}

\begin{figure}[t]
    \centering
    \includegraphics[width=0.44\linewidth]{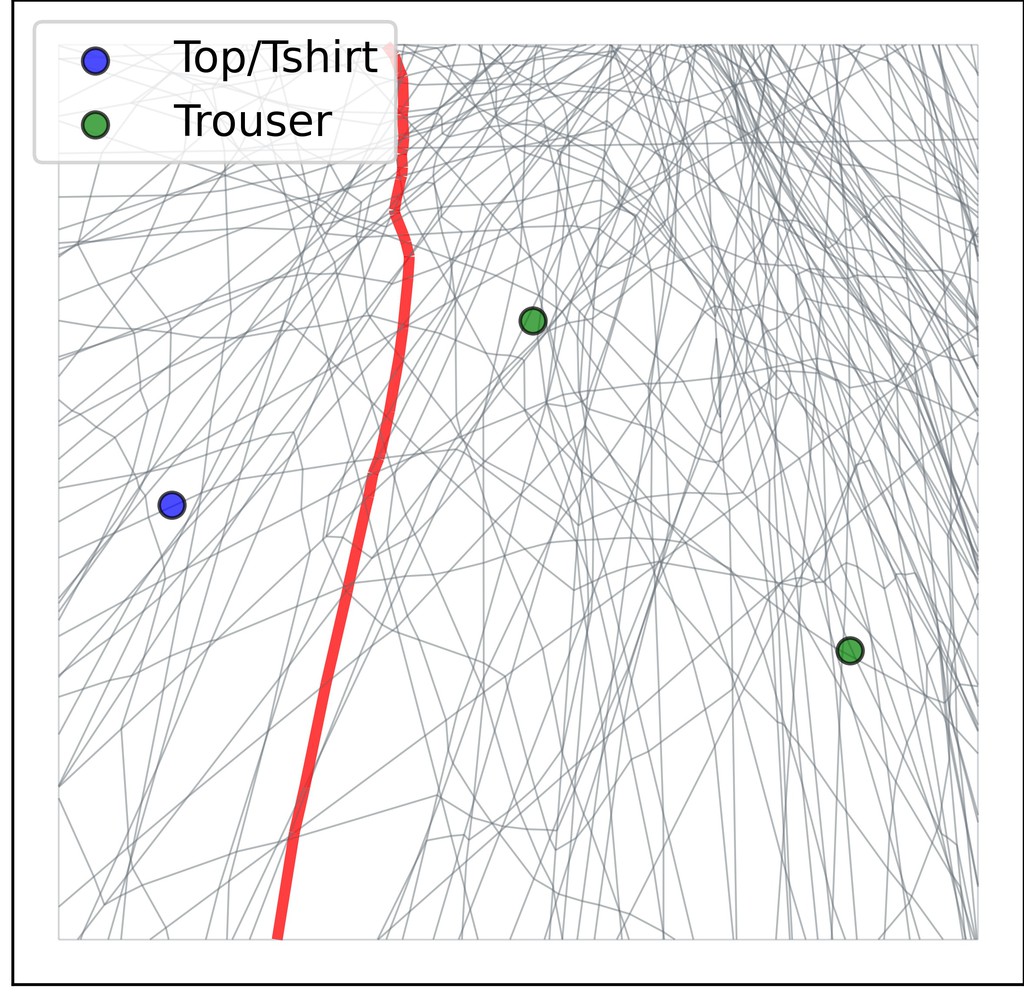}
    \includegraphics[width=0.44\linewidth]{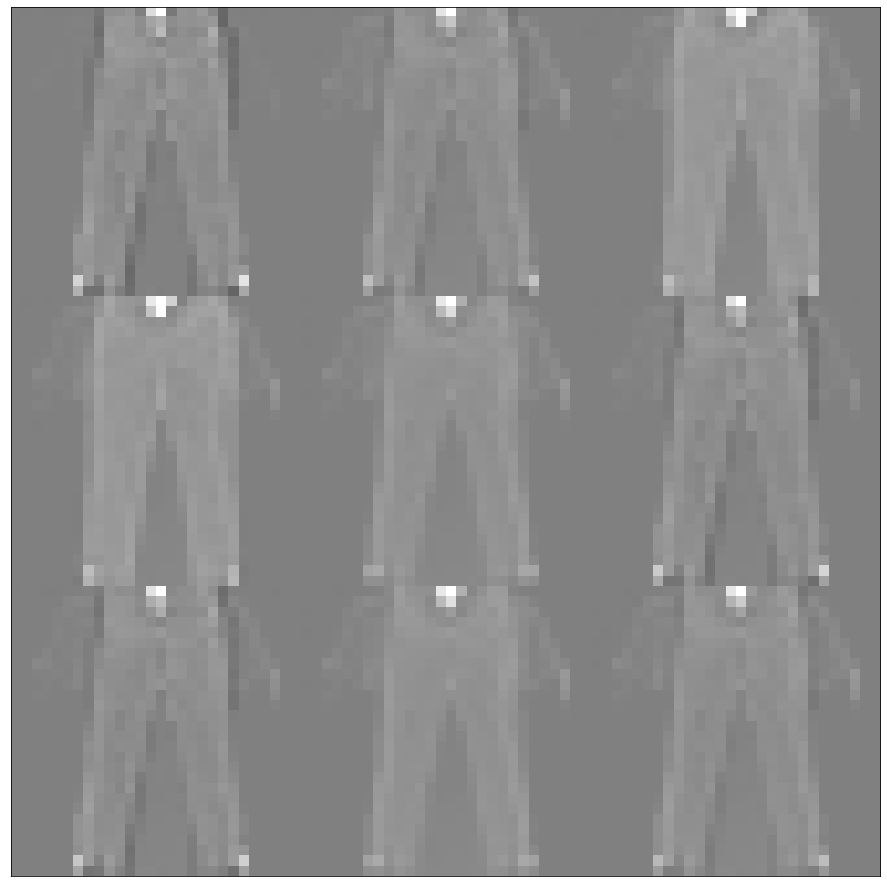}
    \includegraphics[width=0.44\linewidth]{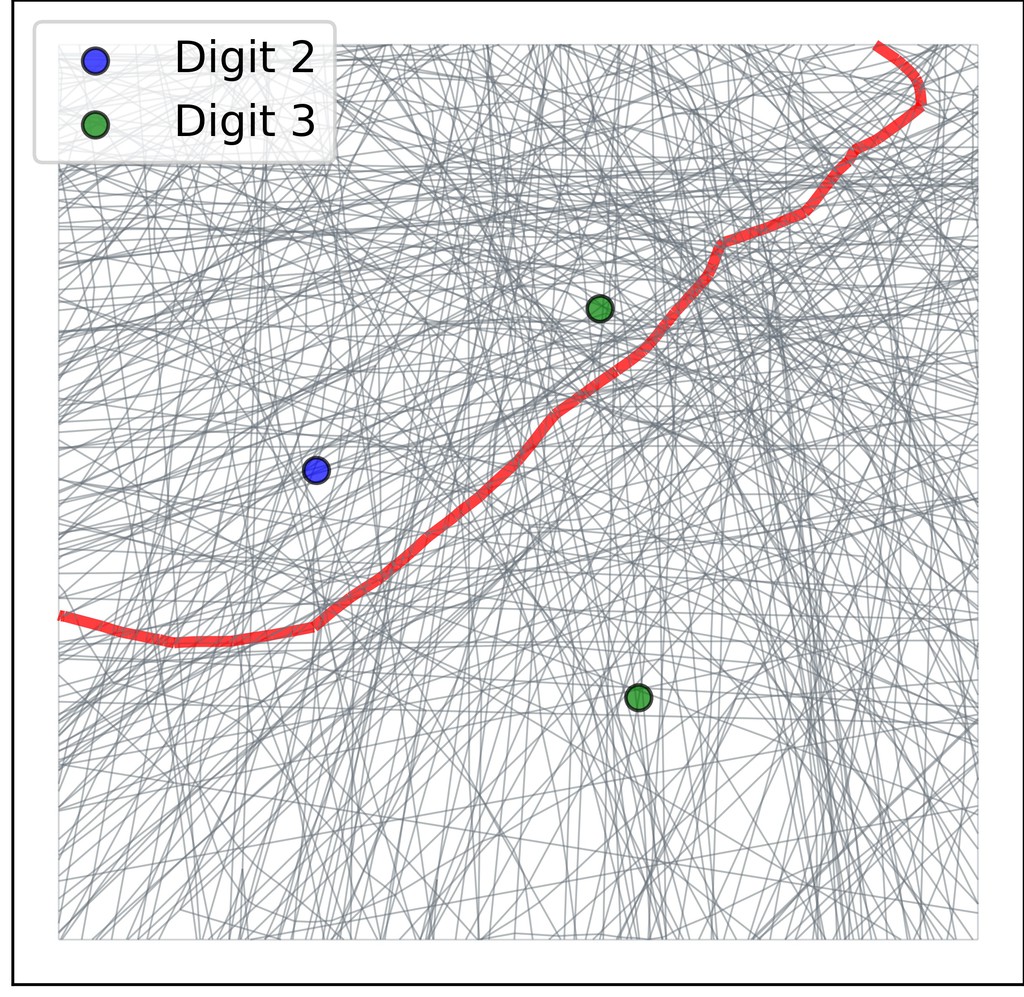}
    \includegraphics[width=0.44\linewidth]{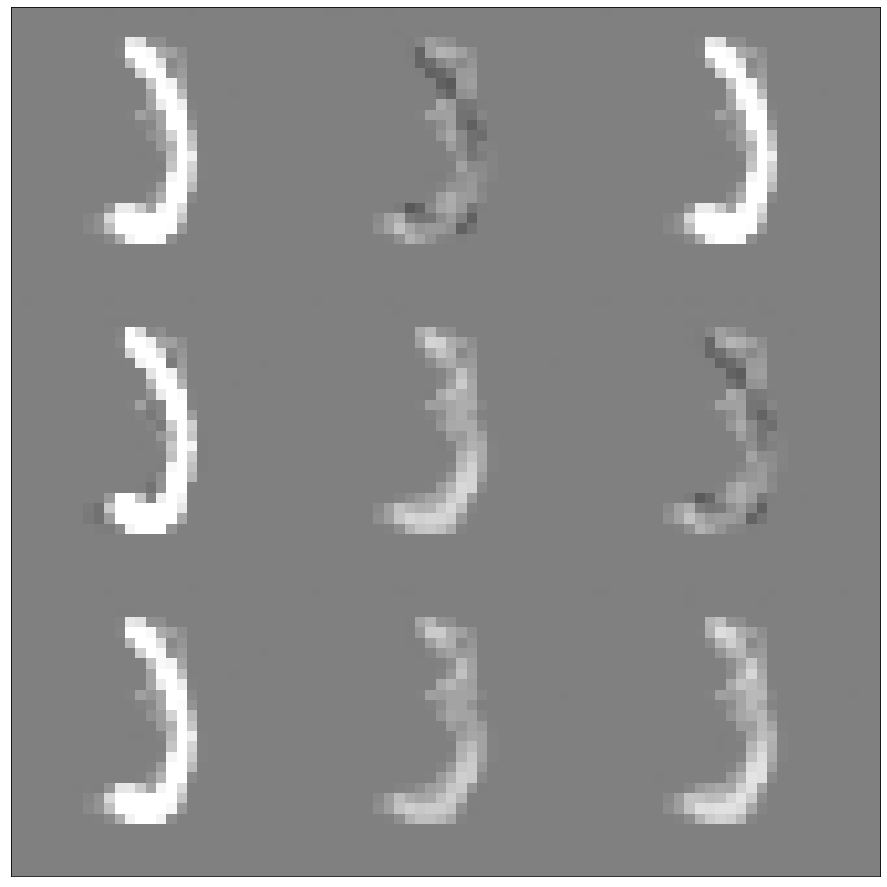}
    \caption{\textbf{
    (Top Left)} Decision boundary visualization for an MLP of width 50 and depth 3 trained on fashion-MNIST. Red lines represent the learned decision boundary while black lines represent the spline partition of the network. \textbf{(Top Right)} Samples from the decision boundary between classes Top and Trouser. The samples are ambiguous with distinguishable attributes present from both classes, indicating a good local decision function. \textbf{(Bottom Left)} SplineCam visualization for a CNN trained on MNIST, with two convolutional layers and one hidden fully connected layer of width 50. One of the digit 3 samples is misclassified by the network as digit 2. \textbf{(Bottom Right)} Samples from the decision boundary between digits 2 and 3 of MNIST. Some of the samples clearly look like the digit three, indicating that the decision boundary here could be sub-optimal. \vspace{-1em} }
    \label{fig:boundary}
\end{figure}


\begin{figure*}[h]
    \centering
    \includegraphics[width=.18\linewidth]{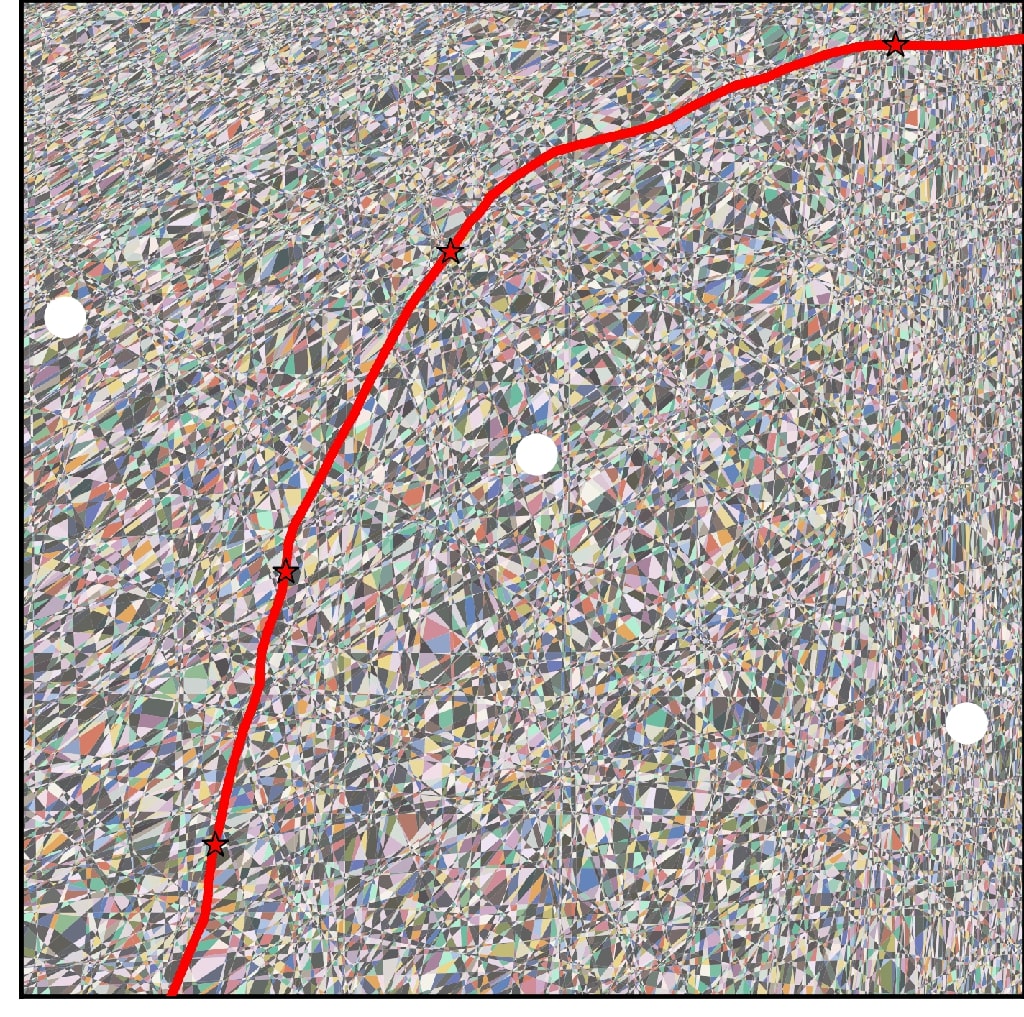}
    \includegraphics[width=.18\linewidth]{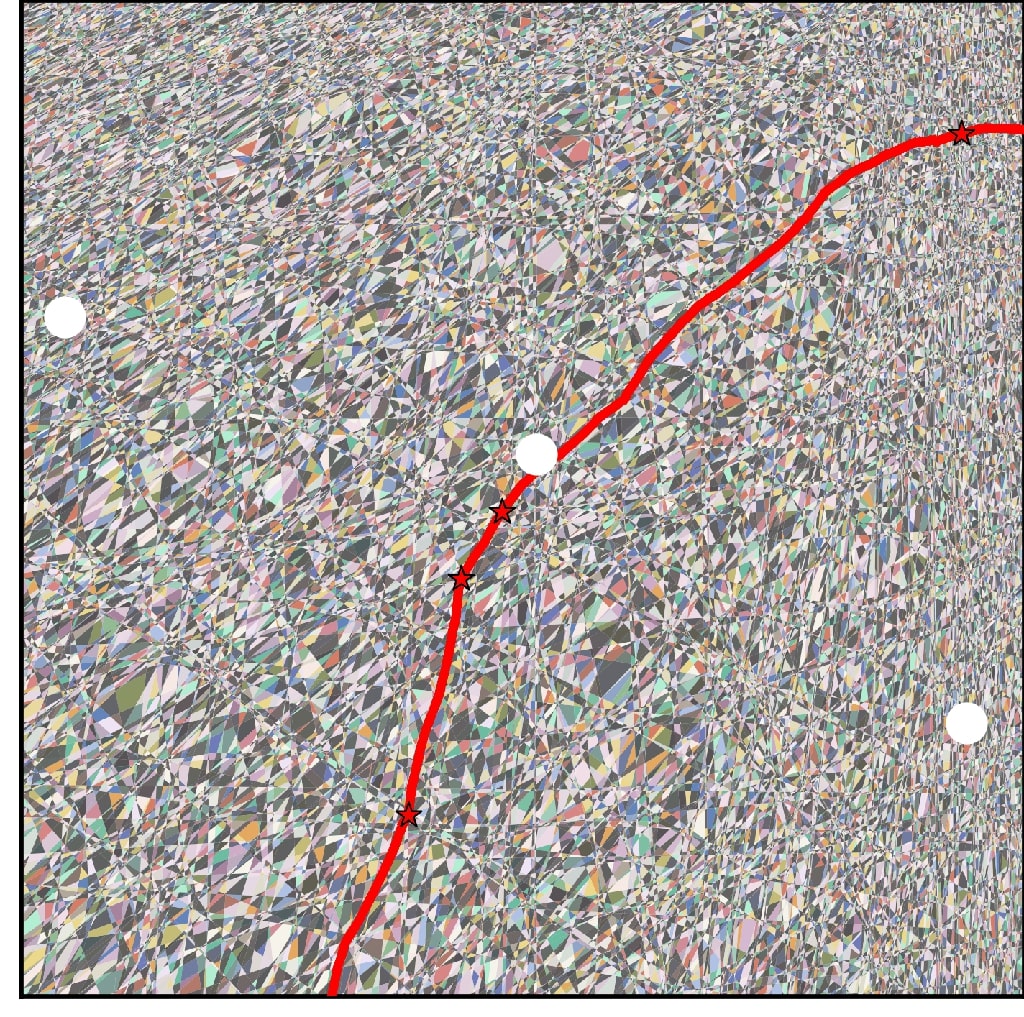}
    \includegraphics[width=.18\linewidth]{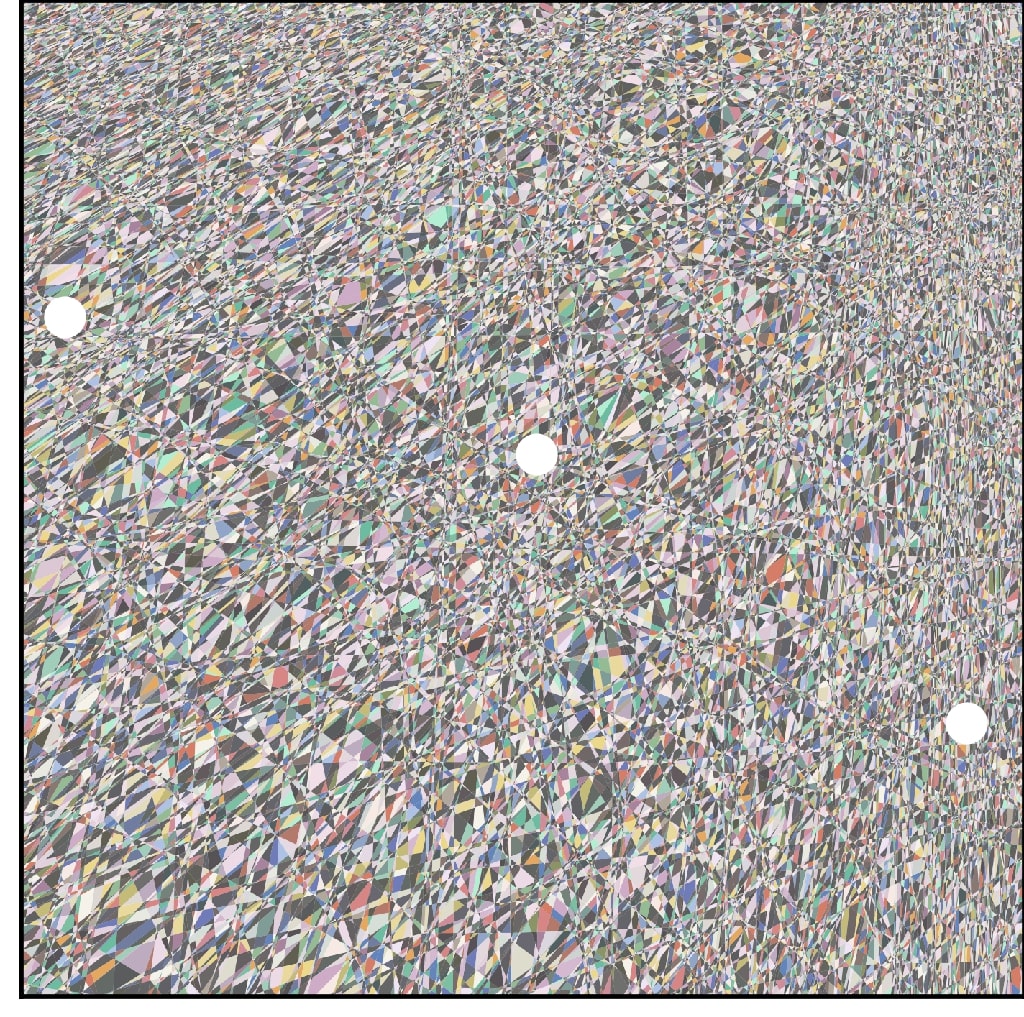}
    \includegraphics[width=.18\linewidth]{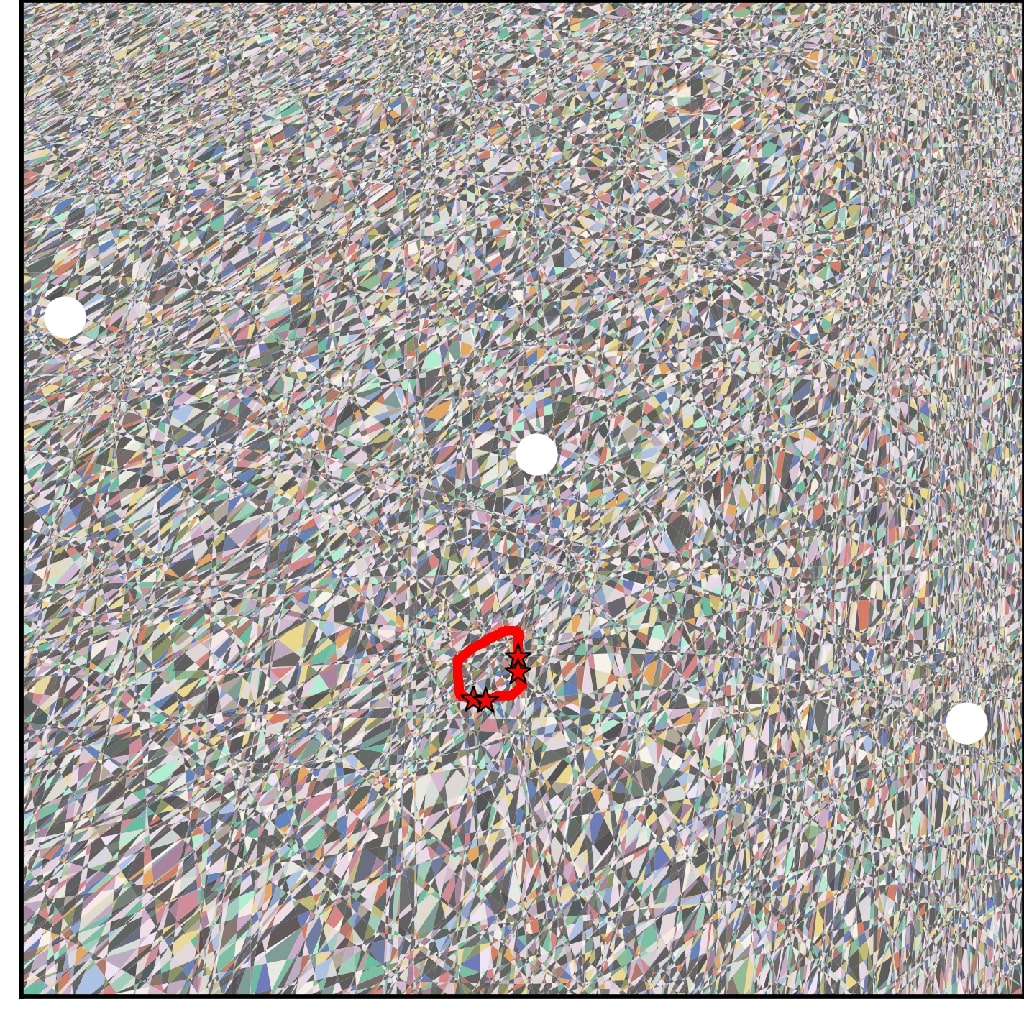}
    \includegraphics[width=.18\linewidth]{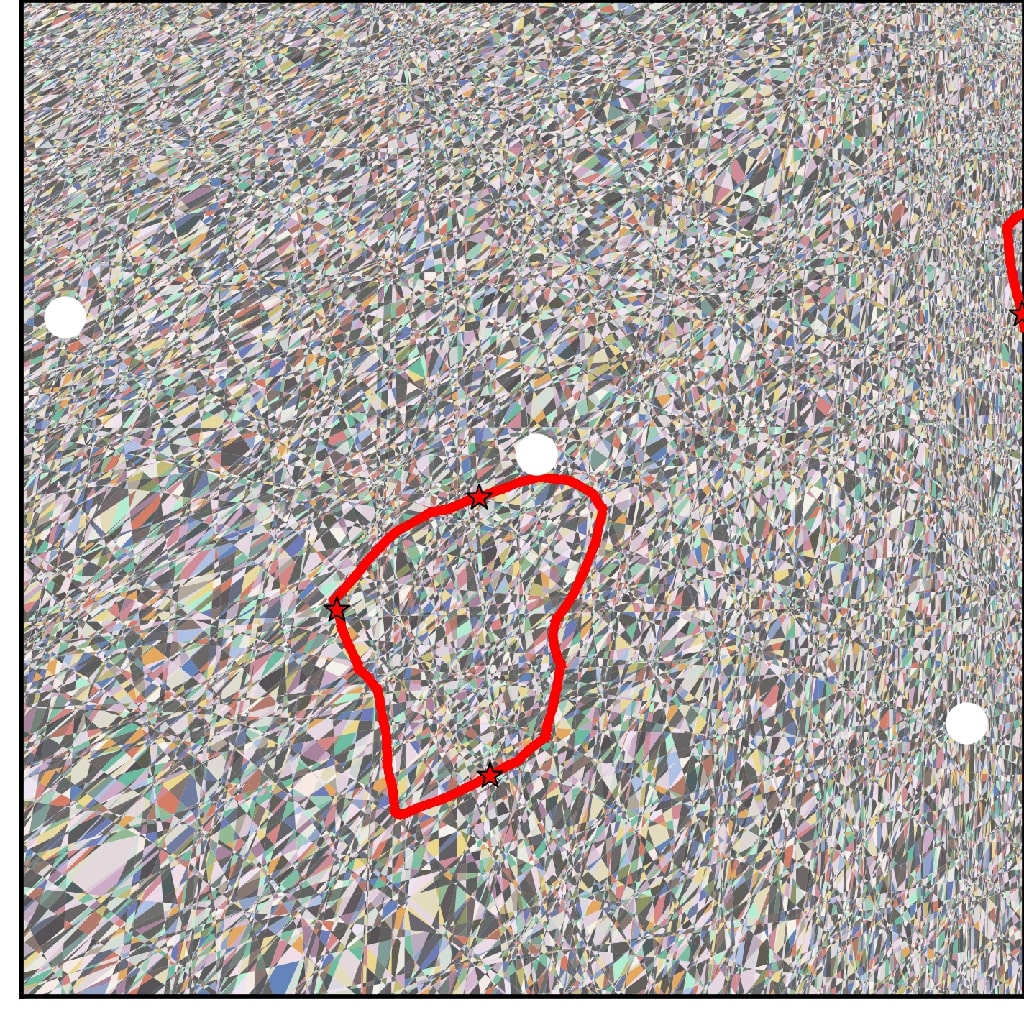}\\
    \includegraphics[width=.18\linewidth]{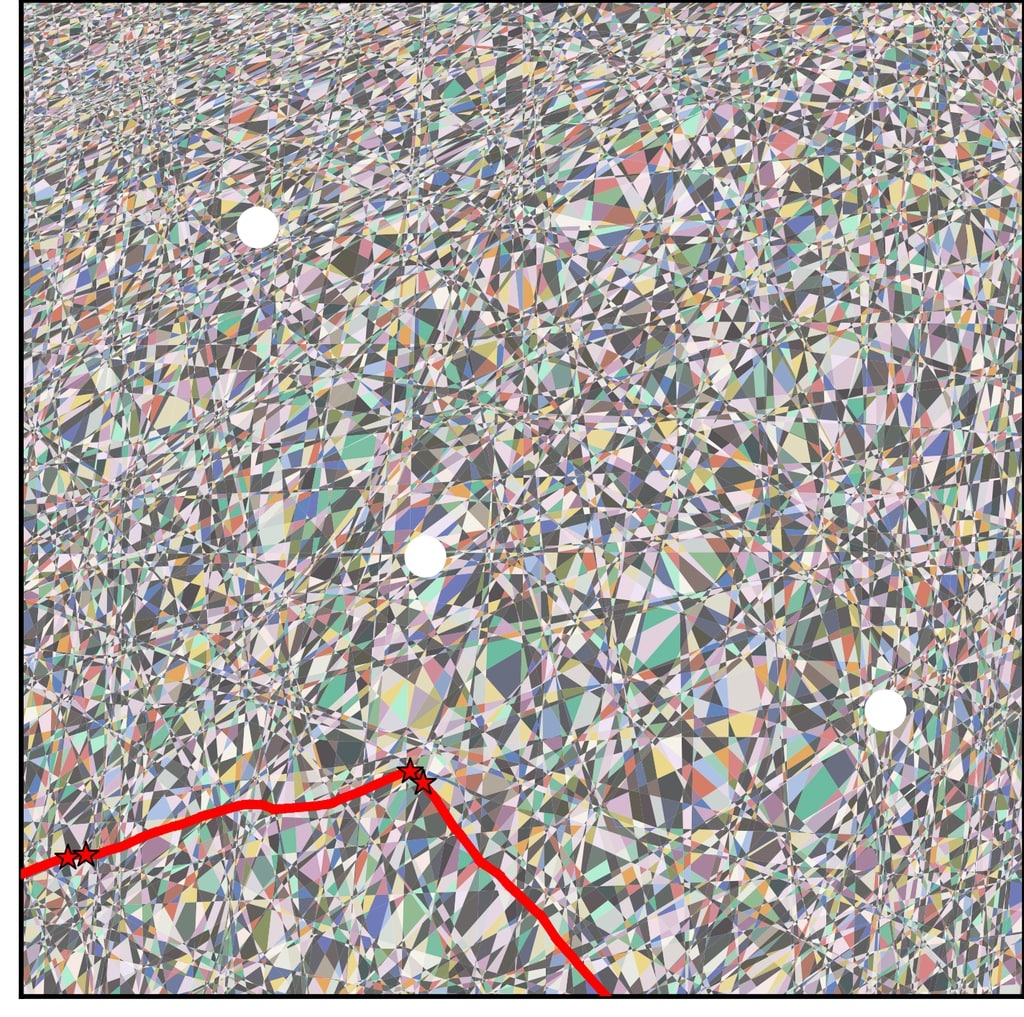}
    \includegraphics[width=.18\linewidth]{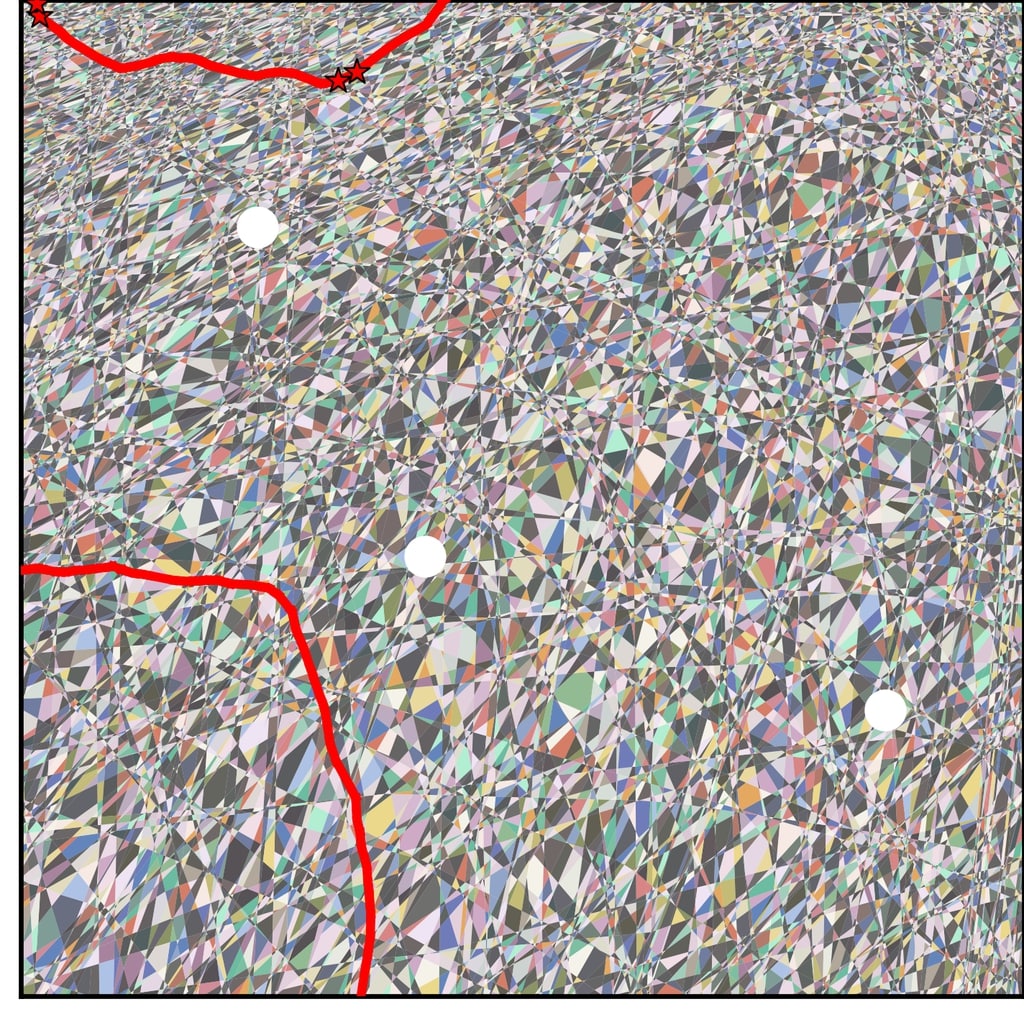}
    \includegraphics[width=.18\linewidth]{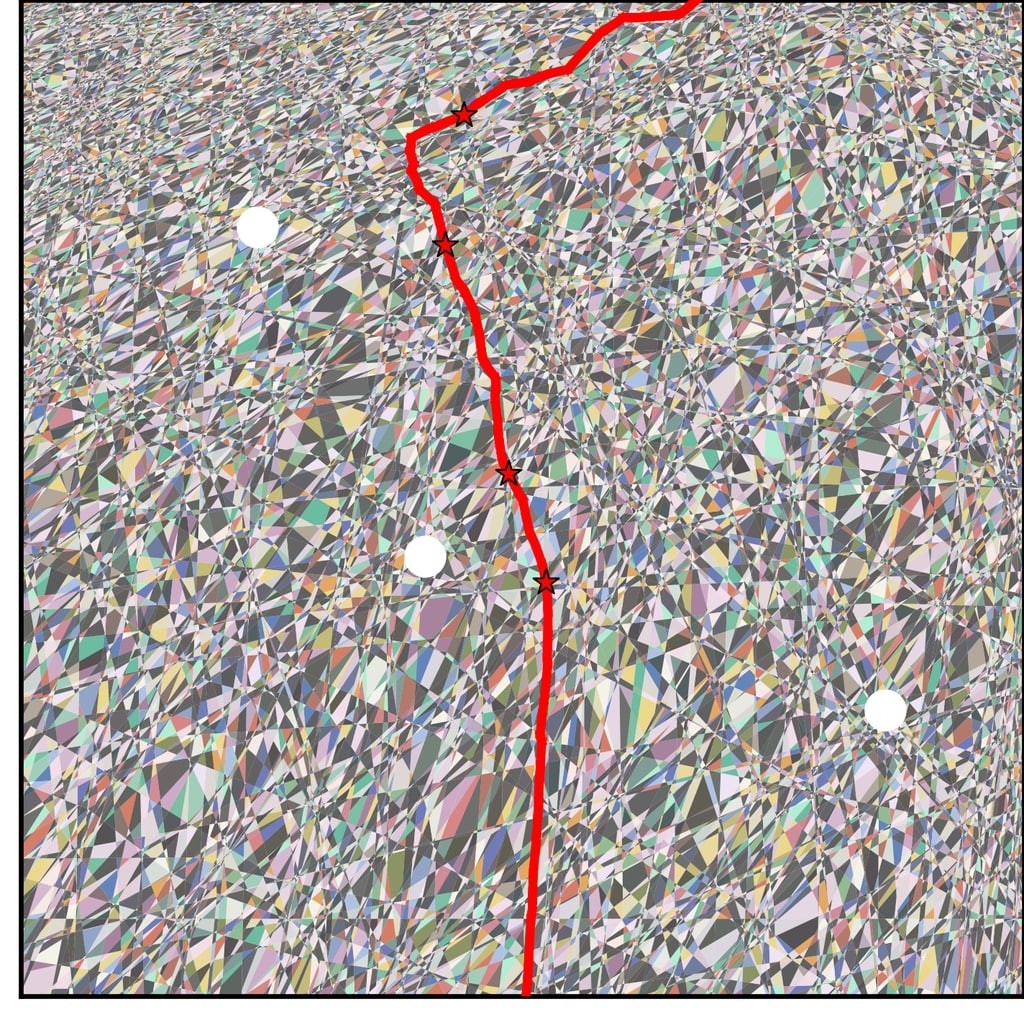}
    \includegraphics[width=.18\linewidth]{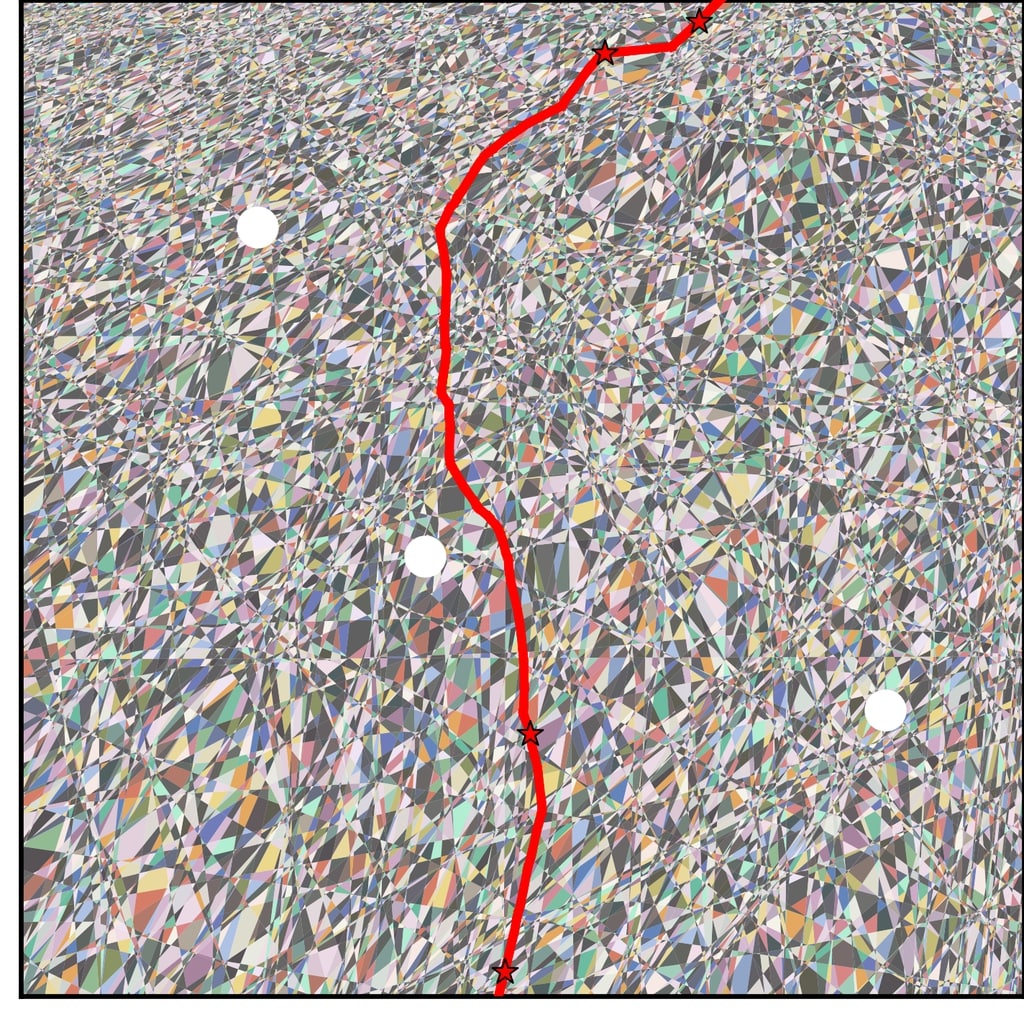}
    \includegraphics[width=.18\linewidth]{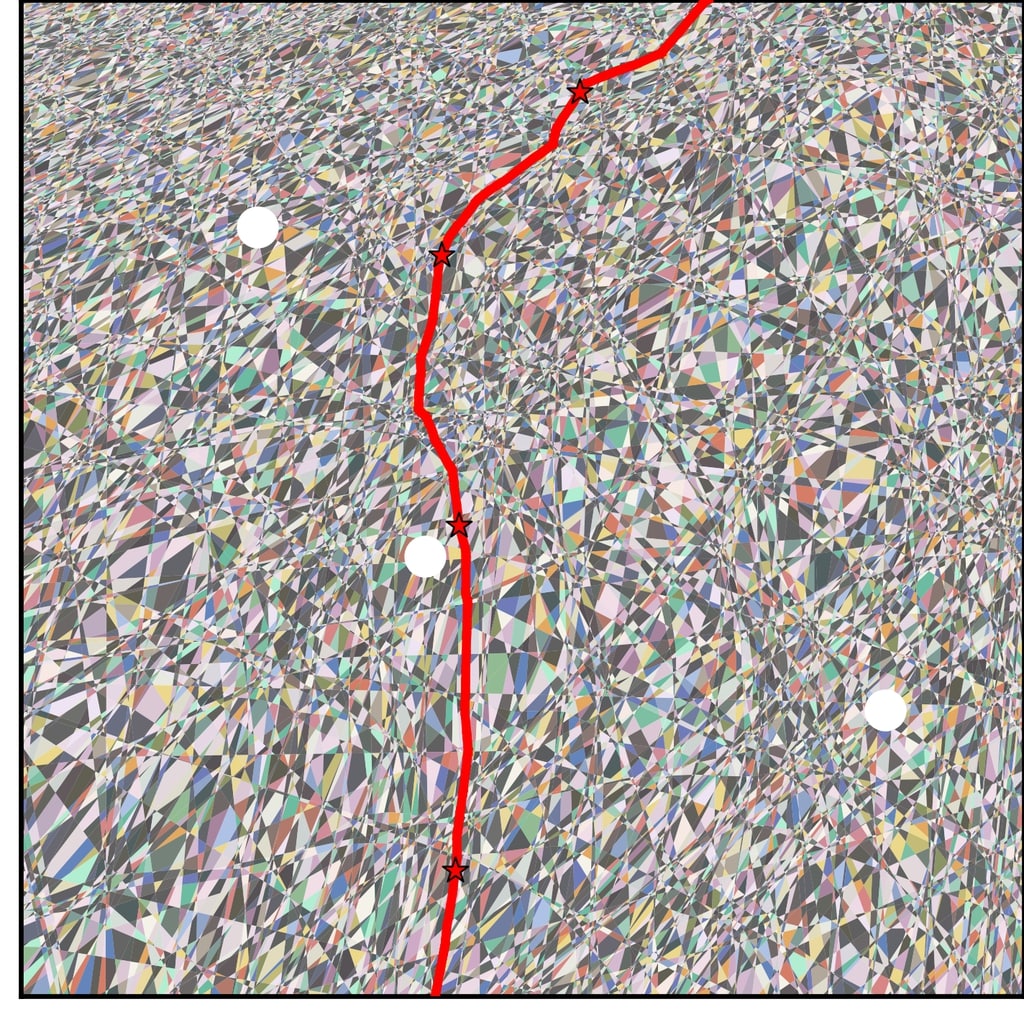}
    \caption{Decision boundary of a $5$ layer convnet during training epochs \{50,100,150,200,300\} ({\bf columns}) trained on a binary classification task to classify between Egyptian cat and Tabby Cat classes of tinyimagenet. White points are images from the dataset that are used to determine the $2D$ input domains. We can observe that between epochs 50 and 100, not much change occurs to the decision boundary ({\bf red line}). However, between epoch 100 and 150, a sudden change occurs for both cases. Especially for top row, the decision boundary moves out of the input domain. Following that, between epochs $200$ and $300$ the boundary stays roughly identical until the end of training. In Suppl. Fig.~\ref{fig:cnn_withrand} we present the change of partition statistics at different train, test and off-manifold locations, while training a CNN on CIFAR 10. In Suppl. Sec.~\ref{appendix:evolution} we present further discussions.}
    \label{fig:dbimgcat}
\end{figure*}

Empirical evidence shows that INRs trained with periodically encoded (PE) coordinates instead of input space coordinates, are able to fit the input signal better with faster convergence rates. While initially inspired by its use in transformer networks there has been very less theoretical investigation of how positional encoding affects learning. 
We train a 2D INR to regress the grayscale intensity of an image, for given pixel coordinates (Fig.\ref{fig:posenc}). We use a ReLU MLP as the INR backbone, with width $10$ and depth $5$, and visualize the partition induced with/without a periodic position encoding front-end. Since SplineCam works with affine nonlinearities only, we use a piecewise approximation of sine/cosine while training, with 5 linear regions for each period of the sine/cosine. We see that using this encoding has negligible effect on performance compared to using continuous cosine and sine functions for encoding. In Fig.~\ref{fig:posenc} we present a layerwise visualization where we separately show for each layer the neurons in the input space. We also highlight in red, the boundary of a single neuron from each layer.

For the PE network, boundaries of some neurons seem to be periodically repeating across the input domain. This is due to the periodic wrapping of space induced by PE, i.e., the input domain is wrapped around in the embedded space between $[-1,1]$ for each dimension, which gets cut by subsequent ReLU hyperplanes. The effect of periodicity is most evident for the first layer hyperplanes, as can be seen from the highlighted neuron in Fig.~\ref{fig:posenc}. Such repetition of neurons across different parts of the input space, \textit{significantly increases the number of regions and weight sharing across input space}, which could be a possible reason for improved convergence \cite{nowlan1992simplifying}.
We also see a layerwise learning of the boundary of the sphere, indicating that multiple neurons across different depths coordinate to complete the final regression task. The absence of some neurons from the input space domain also shows that not all neurons actively partake in the regression task. For example, while for the first layer of the non-PE network, all $10$ hyperplanes intersect the input domain, for the last layer only $4$ of the $10$ neurons intersect the input domain. \textit{This shows how different neurons create redundancy by remaining active/inactive for all possible inputs observed during training.}

\section{How Training Decisions Impact Your Spline}

Recall from \cref{sec:background} that any DN with CPWL nonlinearities is a CPWL mapping or affine spline.
Affine splines have been widely studied and many of their properties, e.g. number of regions in their partition, are known to convey information on function complexity \cite{hanin2019complexity}. In this section, we explore the effect of different training choices, e.g., architecture, data-augmentation, on DN partition.

\begin{figure*}[t]
    \hspace{4.5em} Avg. Region Volume \hspace{5em} Number of Regions \hspace{5em} Avg. Region Eccentricity \hfill\\ 
    \centering
    \includegraphics[trim=0 0 0 1.5em, clip, width=.28\linewidth]{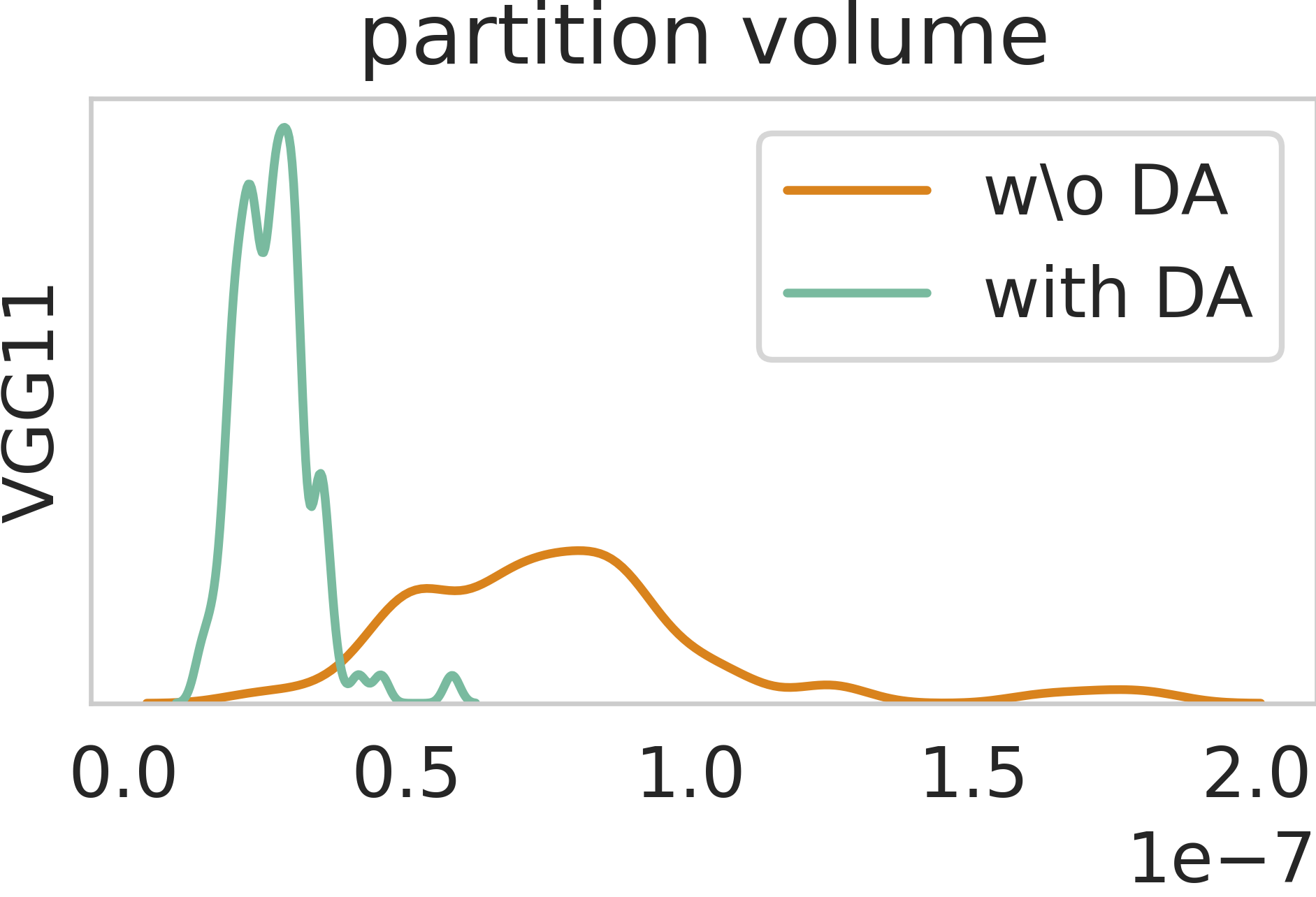}
    \includegraphics[trim=0 0 0 1.5em, clip, width=.28\linewidth]{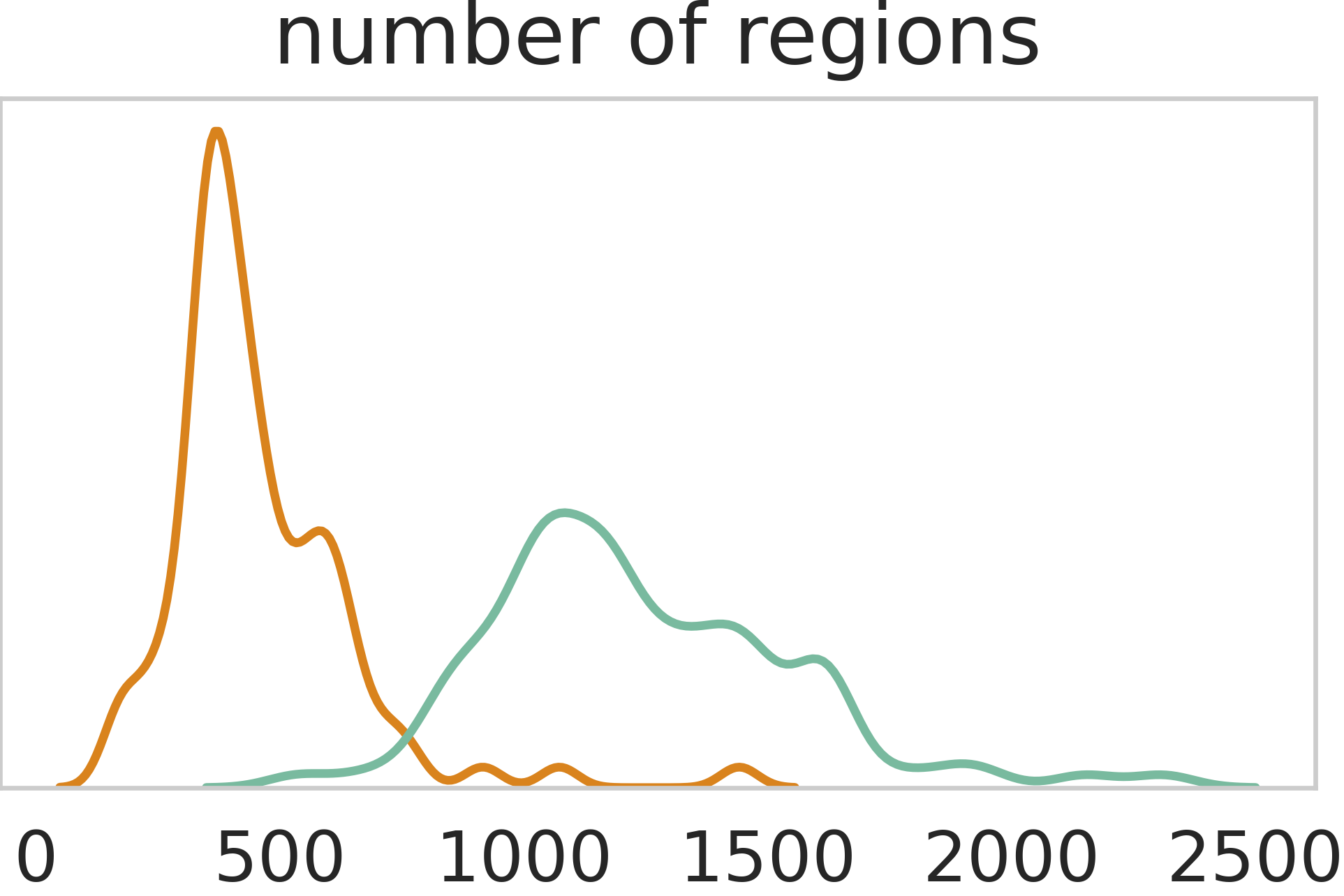}
    \includegraphics[trim=0 0 0 3em, clip, width=.28\linewidth]{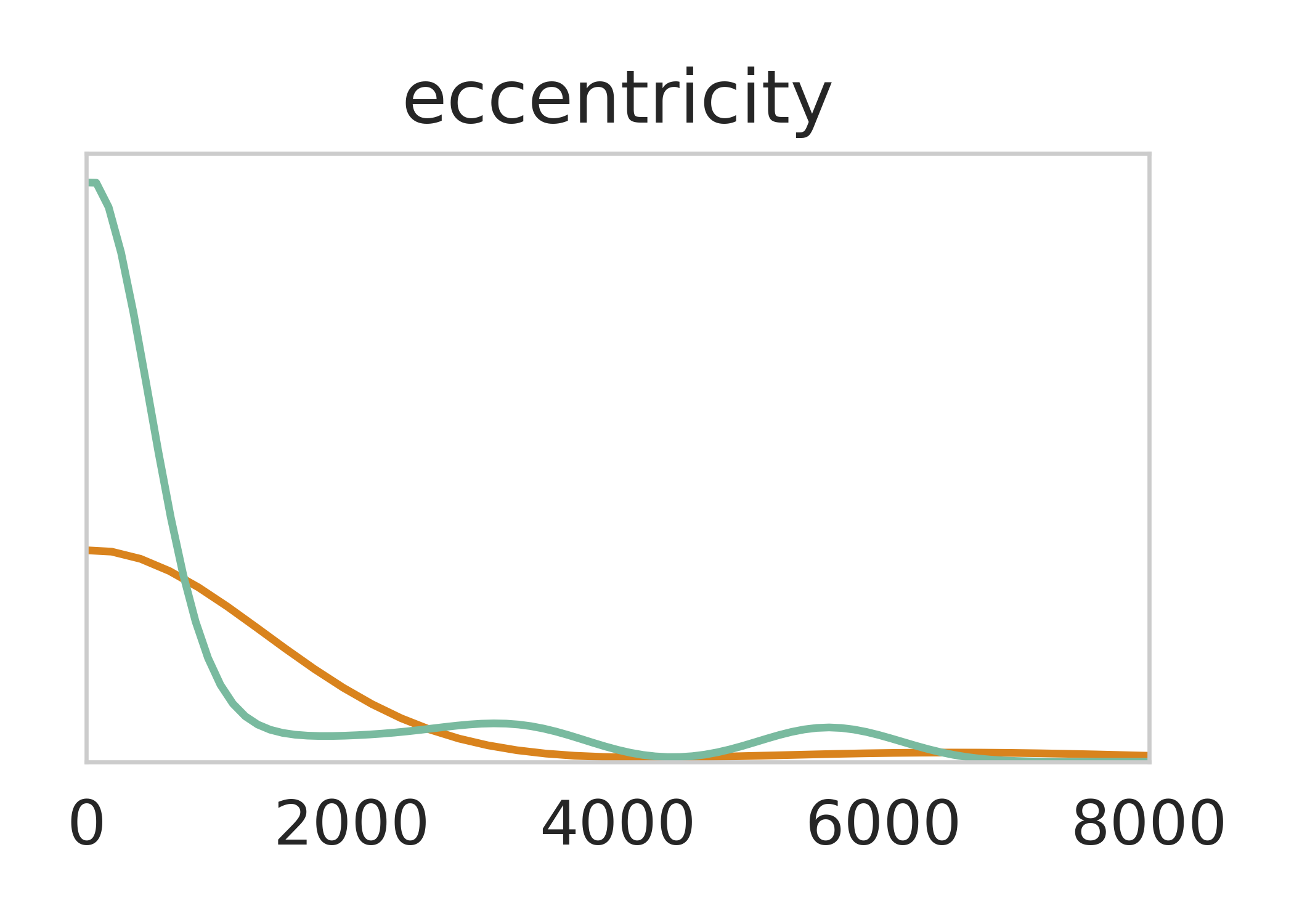}
    \includegraphics[trim=0 0 0 1.5em, clip, width=.28\linewidth]{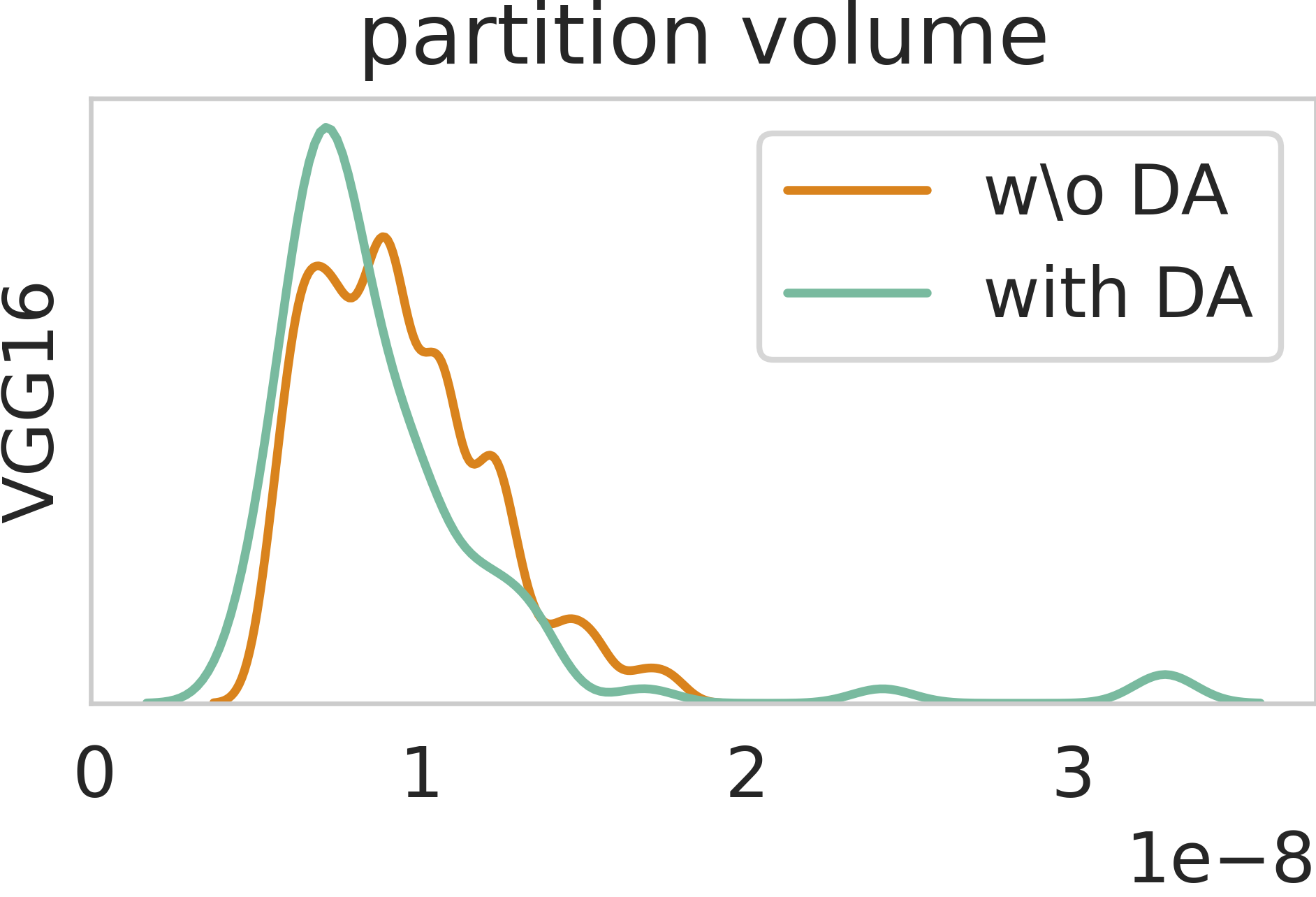}
    \includegraphics[trim=0 0 0 1.5em, clip, width=.28\linewidth]{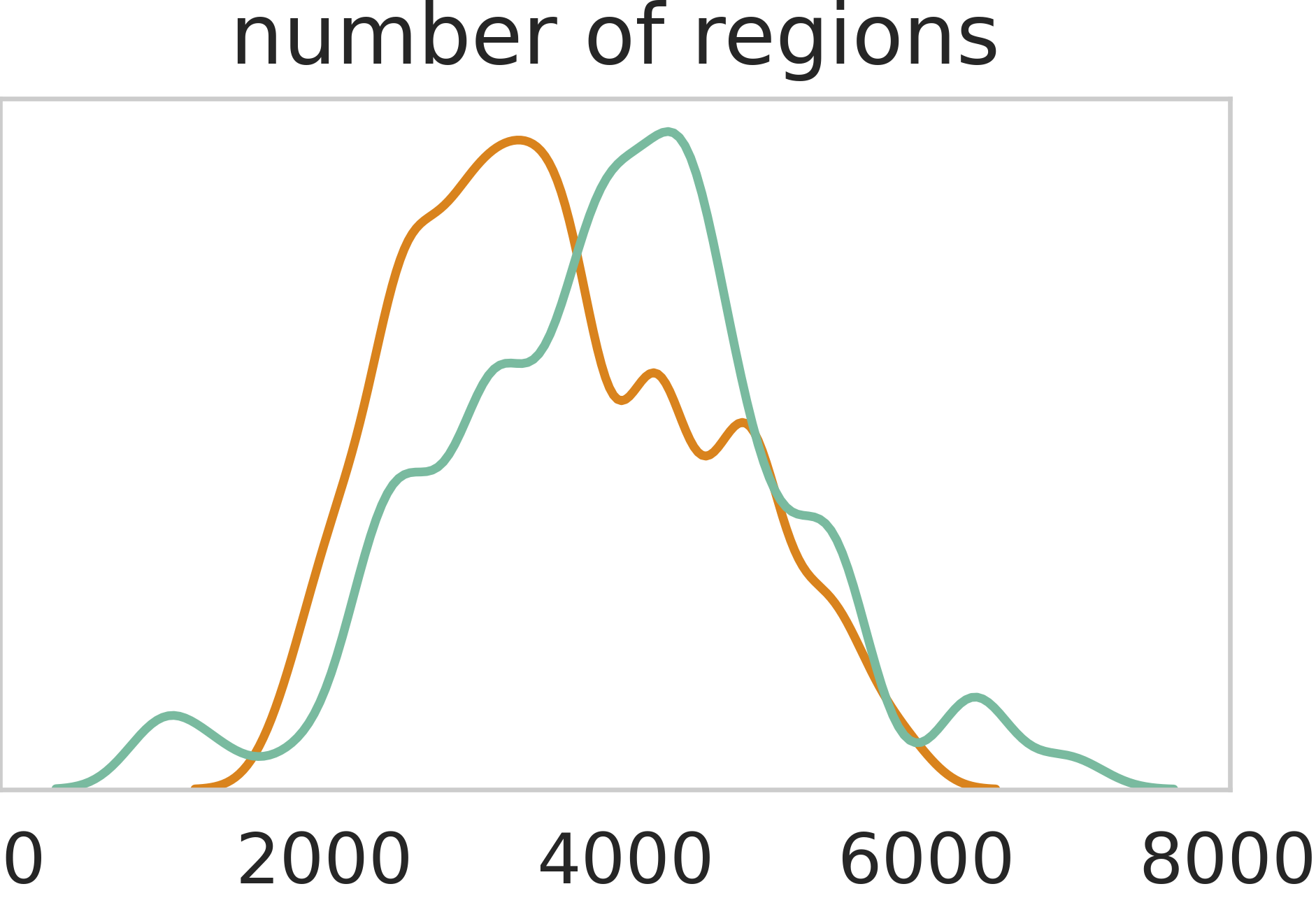}
    \includegraphics[trim=0 0 0 3em, clip, width=.28\linewidth]{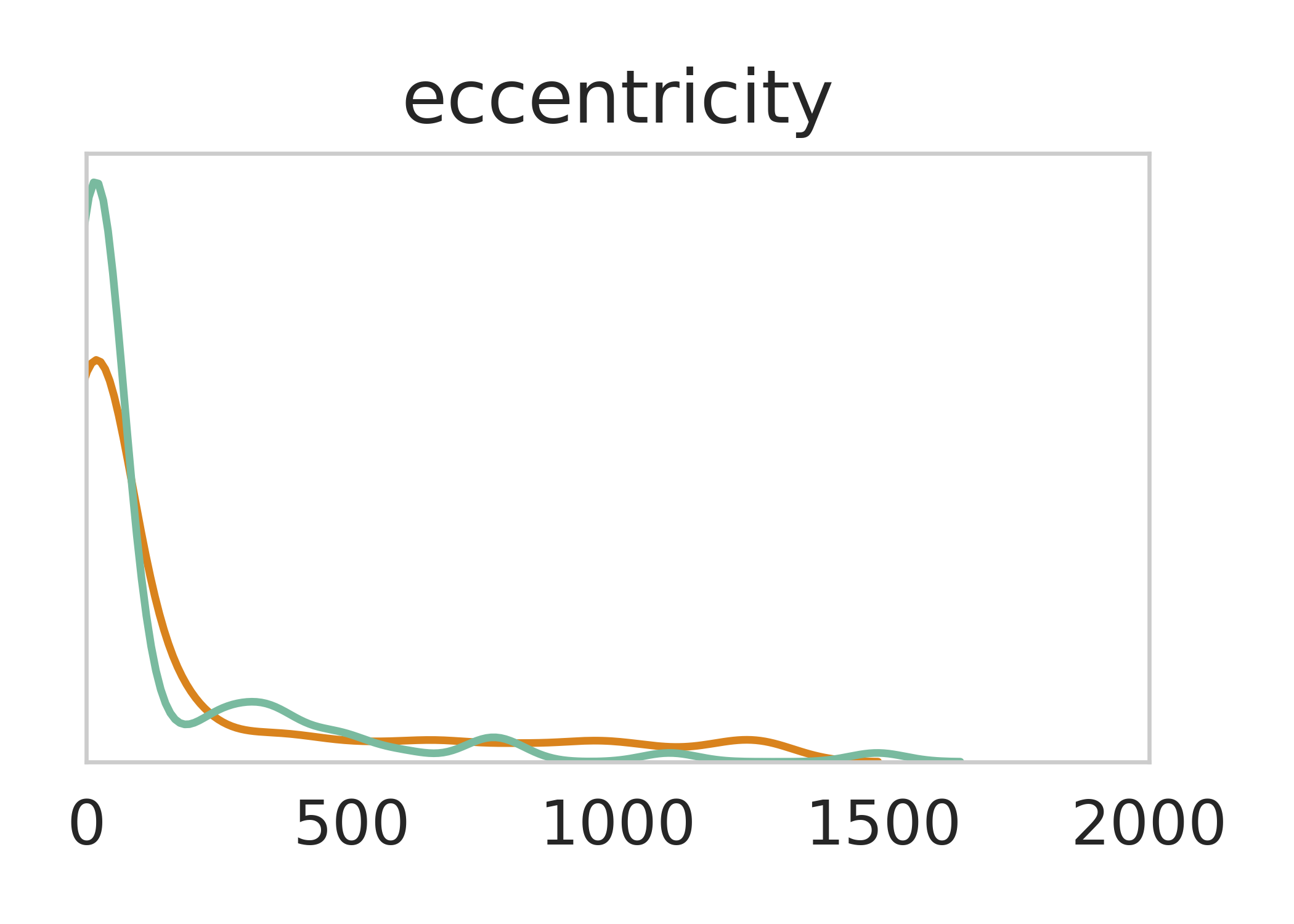}
    \caption{Average partition statistics for 90 tinyimagenet test samples with and without DA training for VGG11 and VGG16. The average volume and number of regions are indicative of partition density whereas eccentricity is indicative of the shape of the regions. For VGG11 a distinct difference in the  statistics can be visualized between DA and non-DA training. DA training significantly increases the partition density at test points, which is indicative of better generalizability. On the other hand, the difference reduces for VGG16 while the overall region density increases. This is expected behavior since the VGG16 has significantly more parameters. For both case, the DA models acquired a similar accuracy of $~51\%$ on the tinyimagenet-200 classification task. In Suppl. Fig.\ref{fig:mnist_mlp_sweep_anr},\ref{fig:mnist_mlp_sweep_ARV},\ref{fig:cifar_cnn_sweep_ARV},\ref{fig:cnn_cifar_sweep_anr} we present partition statistics at random training and test set samples, while varying architecture and hyperparameters.}
    \label{fig:tinyimagDA}
\end{figure*}

\begin{figure}[ht]
    \centering
    \includegraphics[trim=.8em .3em .5em 1em, clip, width=0.85\linewidth]{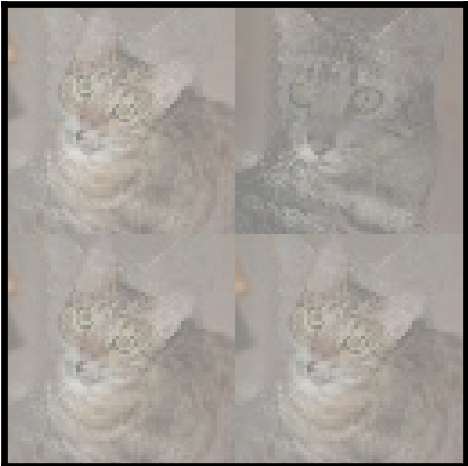}
    \caption{Images from the decision boundary of a CNN trained to perform binary classification between Tabby Cat and Egyptian Cat classes of tinyimagenet. Larger sets of samples from the decision boundary along with partition visualizations are provided in the Suppl. Fig.~\ref{fig:cat_partition}. Samples from the decision boundary are necessarily linear combinations of the samples used to determine the 2D plane for SplineCam. The weights of the linear combination are determined by the geometry of the decision boundary.}
    \label{fig:decboundim}
\end{figure}

\subsection{Impact of Architecture on Partitions Properties}
\label{sec:experiments}

In this section we explore the impact of the DN's architecture on the partition. To start off, we look at the partitions induced by MLPs and CNNs when trained on fashion-MNIST and MNIST datasets. 
We quantify the characteristics of the partitions via the following measures- Average Region Volume (ARV), Average Number of Vertices (ANV), Number of Regions (NR), and Average Region Eccentricity (ARE) which is defined as the ratio of the max and min pairwise distance between vertices. For a given dataset, we fix the input domain and switch between convolutional and fully connected architectures to draw emphasis on the effect of the convolutional layers. We see that in convolutional architectures, there is a significantly higher number of partition regions formed. This is intuitive since convolutional architectures repeat the same set of parameters across sets of dimensions via a circulant weight matrix. This increases the number of cuts by the same neurons in the input space, demonstrating higher complexity for the same number of parameters \cite{montufar2014number}. In Tab.~\ref{tab:my-table} we present partition statistics comparisons between architectural choices and datasets. We see that the eccentricity and volume of the polytopes are significantly smaller for convolutional architectures compared to fully connected architectures.
These can also be visualized in Fig.~\ref{fig:boundary}. In Suppl. Fig.\ref{fig:cnn_withrand} we present the variation of partition statistics with training, for training points, test points and points off the data manifold. We see that partition density increases for sample on the data manifold regardless of being on training or testing.

\subsection{Data-Augmentation}

Data-Augmentation is a ubiquitous technique that has led to great improvements into DN performance \cite{fawzi2016adaptive}. It is still unclear what is the impact of DA onto the DN's mapping. In fact, while explicit regularizers of DA has been found \cite{hernandez2018data,https://doi.org/10.48550/arxiv.2202.08325} and while many empirical studies have emerged \cite{wang2019implicit}, none truly pinpoint what is different between DNs trained with DA and DNs trained without.

In order to provide a first quantitative understanding of what actually changes within a DN when DA is applied, we propose to rely on SplineCam. In Fig.~\ref{fig:tinyimagDA} we provide the ARV, NR, and ARE for spline partitions computed from randomly oriented 2D domains centered on 90 tinyImageNet test samples. 
For each sample, computation of partition statistics take $\sim7$mins for VGG11. We see that partition statistics vary significantly between DA and non-DA for VGG11 but not as significantly for VGG16. We provide partition visualizations for VGG11 in Suppl. Fig.~\ref{fig:VGG11_dec_bound}. Region volumes can vary within a given 2D input domain as well as between input domain orientations.
In Suppl. Sec.~\ref{appendix:random_orientation} we present SplineCam partition statistics for random orientations at different training points and discuss the variability of statistics between orientations. 

\section{Conclusions}

In this paper, we have developed SplineCam, the first provably exact method to compute the geometry of the input-space partition of a DN based on CPWL nonlinearities. 
We have demonstrated SplineCam's ability to visualize a large-scale DN's decision boundary, obtain samples that are provably on the boundary, and characterize DNs based on their partition statistics. 
SplineCam and its underlying theory open up several new avenues of exploration for practical neural networks, including quantifying the quality of initialization at training data points during transfer learning, improving INR initialization, and visualizing partition dynamics for different training strategies, to name a few.



\vspace{-0.2cm}
\section*{Acknowledgements}
\vspace{-0.2cm}
{\small Humayun and Baraniuk were supported by NSF grants CCF-1911094, IIS-1838177, and IIS-1730574; ONR grants N00014-18-12571, N00014-20-1-2534, and MURI N00014-20-1-2787; AFOSR grant FA9550-22-1-0060; and a Vannevar Bush Faculty Fellowship, ONR grant N00014-18-1-2047.}

{\small
\bibliographystyle{ieee_fullname}
\bibliography{refs}
}
\clearpage

\appendix

\setcounter{page}{1}

\twocolumn[
\centering
\Large
\textbf{SplineCam: Exact Visualization of Deep Neural Network Geometry and Decision Boundary} \\
\vspace{0.5em}Supplementary Materials \\
\vspace{.5em}
{\small  Codes available at \href{https://bit.ly/splinecam-git}{SplineCAM Github}\\Google Colab demo
\href{https://bit.ly/splinecam-demo}{https://bit.ly/splinecam-demo}\\

}
\vspace{.5em}
] %
\appendix

We provide the following supplementary materials (SMs) as support of our theoretical and empirical claims. This SM is organized as follows. 

\cref{appendix:maso} provides the necessary background results for SplineCam and elaborates on how any deep neural network or implicit neural representation function with piecewise contiuous affine activations are max affine splines.
\cref{appendix:implementation} provides further implementation details, extending from Sec.~\ref{sec:method} and including hardware and software requirements.

\cref{appendix:complexity} elaborates on the computational complexity of the method. We present experiments assessing the time complexity of SplineCam varying the width of a single layer MLP and varying the volume of the input domain for VGG11. In \cref{appendix:extra} we present new experiments, first \cref{appendix:evolution} discusses the change of partition characteristics with training epochs and while varying architecture parameters (e.g., width for MLP and number of filters for a CNN). We also present the change of characteristics across different parts of the input space, e.g., around training samples, test samples and regions off the data manifold. In \cref{appendix:random_orientation} we present quantitative results on the variation of partition statistics while varying the orientation of the 2D input domain of interest. We see that the variation is considerably low between random orientation, showing that a single 2D slice can possibly be good enough to characterize the partition in different parts of the input space. Finally, in \cref{appendix:usage} we expand on the usage of SplineCam and present code blocks as explainers for how the SplineCam framework operates.

\begin{figure}[]
    \centering
    \includegraphics[width=\linewidth]{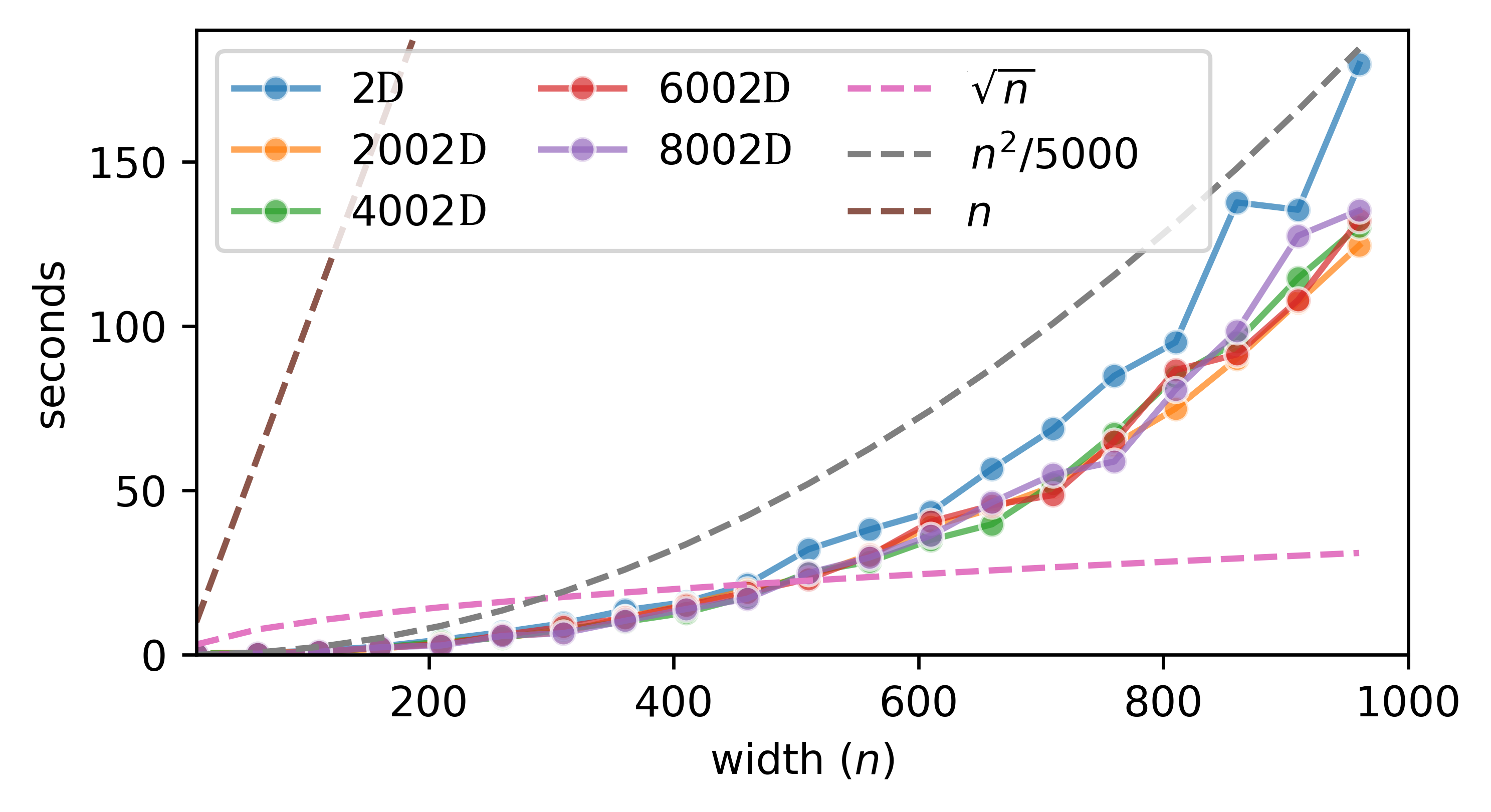}
    \includegraphics[width=\linewidth]{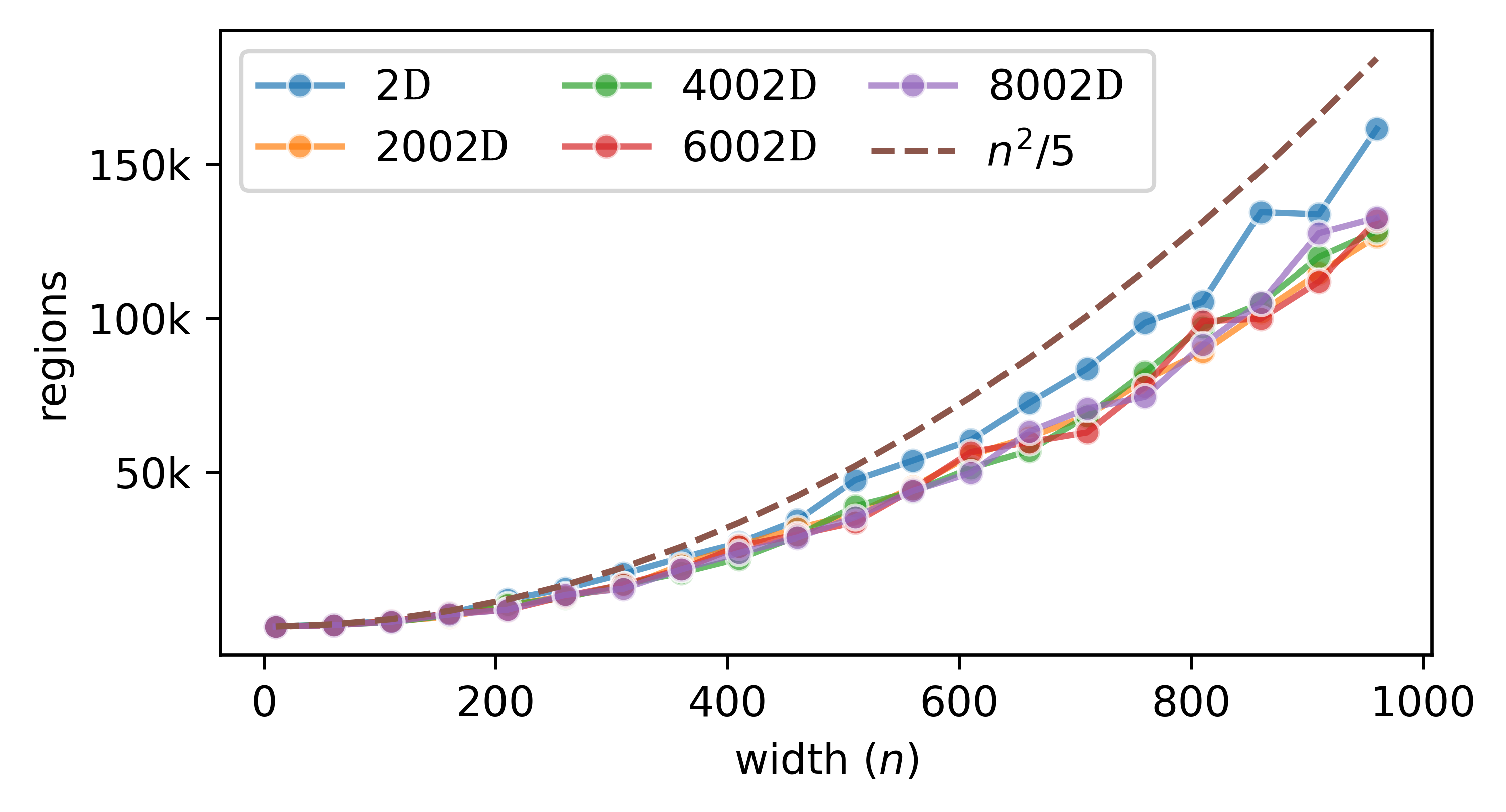}
    \caption{Time complexity of computing the partition regions \textbf{(Top)} and growth of the number of regions with width \textbf{(Bottom)}, for a single layer randomly initialized MLP with varying width ($n$) and input dimensionality. For all the input dimensions, we take a randomly oriented, square 2D domain centered on the origin, with an area of $4$ units.
    Note that with increasing input dimensionality, we get reduced number of hyperplane intersections with our 2D domain of interest, therefore we can see a slight reduction in the wall time required and also the number of regions.}
    \label{fig:complexity_singleMLPlayer}
\end{figure}

\section{Background on Continuous Piecewise Affine Deep Networks}
\label{appendix:maso}
A {\em max-affine spline operator} (MASO) concatenates independent {\em max-affine spline} (MAS) functions, with each MAS formed from the point-wise maximum of $R$ affine mappings
\cite{magnani2009convex,hannah2013multivariate}. For our purpose each MASO will express a DN layer and is thus an operator producing a $D^{\ell}$ dimensional vector from a $D^{\ell-1}$ dimensional vector and is formally given by
\begin{align}
{\rm MASO}(\bv;\{\bA_r,\bb_r\}_{r=1}^R) = \max_{r=1,\dots,R} \bA_{r}\bv + \bb_{r},\label{eq:MASO}
\end{align}
where $\bA_r \in \mathbb{R}^{D^{\ell}\times D^{\ell-1}}$ are the slopes and 
$\bb_r \in \mathbb{R}^{D^{\ell}}$ 
are the offset/bias parameters and the maximum is taken coordinate-wise. For example, a layer comprising a fully connected operator with weights $\bW^{\ell}$ and biases $\bb^{\ell}$ followed by a ReLU activation operator corresponds to a (single) MASO with $R=2, \bA_1=\bW^{\ell},\bA_2=\mathbf{0}, \bb_{1}=\bb^{\ell}, \bb_{2}=\mathbf{0}$.
Note that a MASO is a {\em continuous piecewise-affine} (CPA) operator \cite{wang2005generalization}.

The key background result for this paper is that {\em the layers of DNs constructed from piecewise affine operators (e.g., convolution, ReLU, and max-pooling) are MASOs}
\cite{balestriero2018spline}:
\begin{align}
    \exists R \in \mathbb{N}^{*}, \exists \{\bA_r,\bb_r\}_{r=1}^{R}  \\
    \text{ s.t. }{\rm MASO}(\bv;\{\bA_r,\bb_r\}_{r=1}^R)=g^{\ell}(\bv), \forall \bv \in \mathbb{R}^{D^{\ell-1}},\label{eq:MASO_layer}
\end{align}
making any Implicit Neural Representation or even Deep Generative Networks a composition of MASOs.
\\
The CPA spline interpretation enabled from a MASO formulation of DNs provides a powerful global geometric interpretation of the network mapping based on a partition of its input space $\R^S$ into polyhedral regions and a per-region affine transformation producing the network output. The partition regions are built up over the layers via a {\em subdivision} process and are closely related to Voronoi and power diagrams \cite{balestriero2019geometry}.

\begin{figure}[h]
    \centering
    \includegraphics[width=\linewidth]{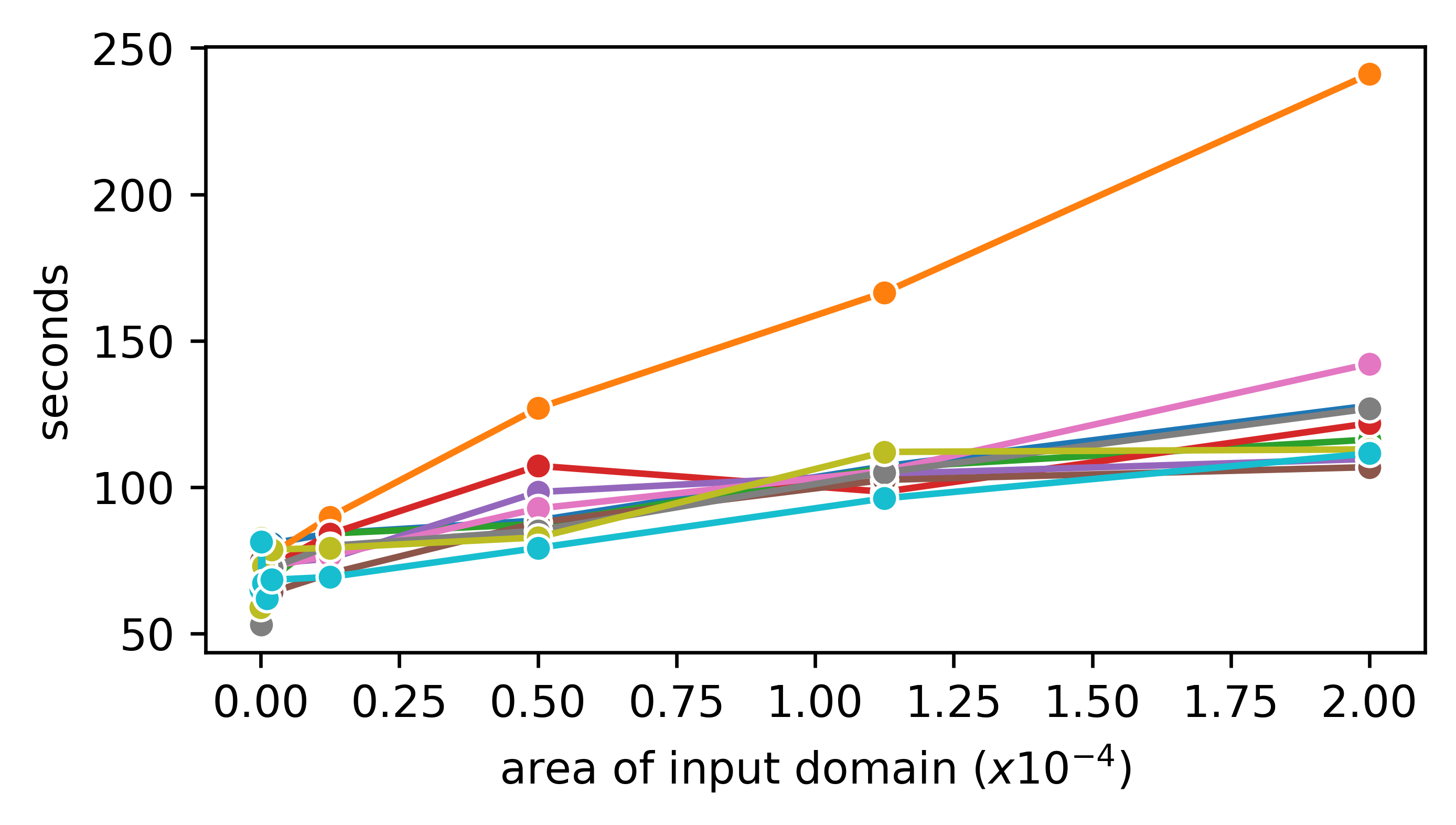}
    \includegraphics[width=\linewidth]{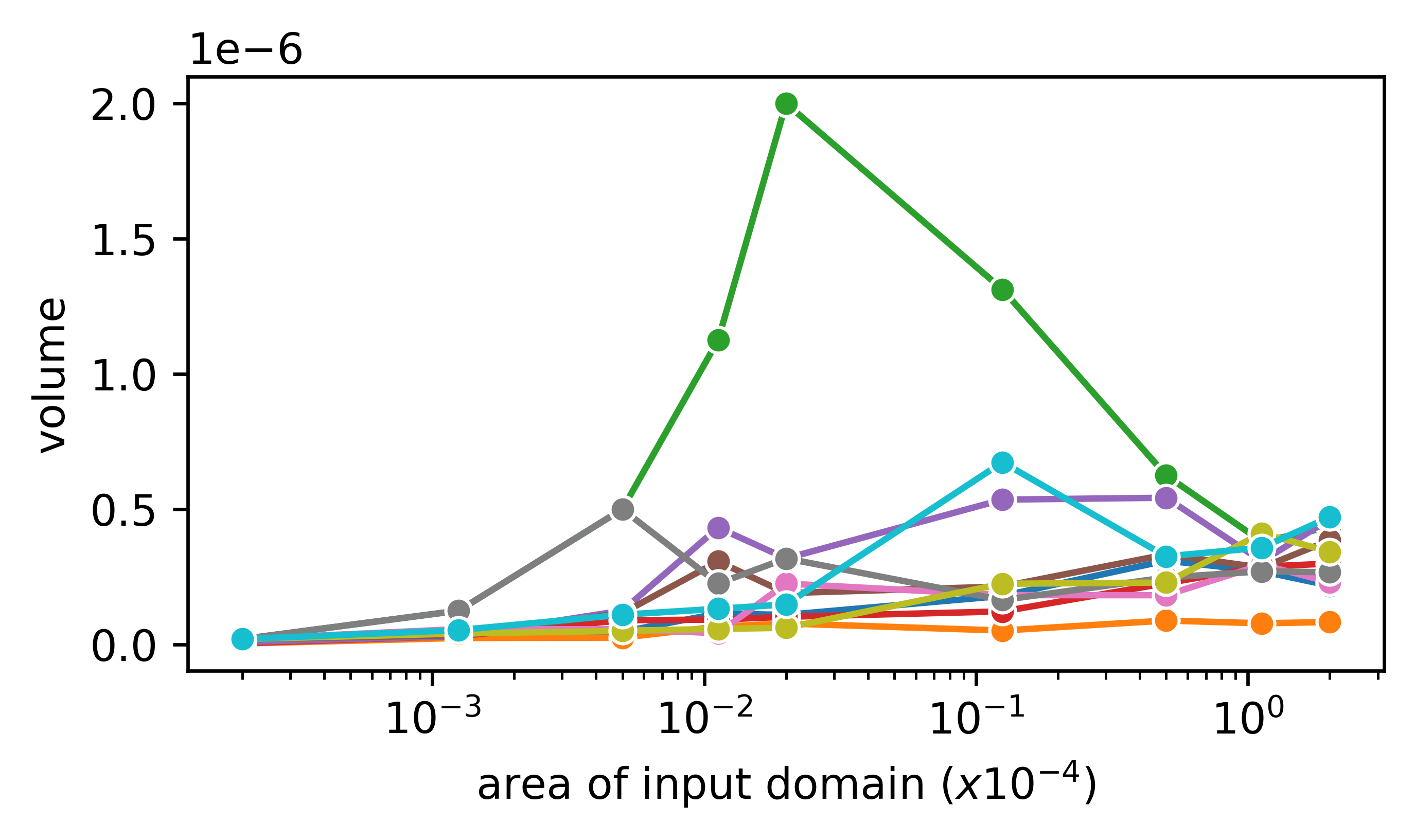}
    \caption{Time complexity of computing the partition regions \textbf{Top} and average region volume (ARV) \textbf{Bottom} while increasing the area of the input domain, for a VGG11 trained on tinyimagenet-200. Each line corresponds to one of $10$ training samples anchored on which the 2D input domains are defined, with random orientations. We see that the required time scales linearly with the size of the input domain. Apart from that, we see that as the neighborhood size is increased for some samples, the ARV increases and then decreases again. This could indicate that while the ARV is small for smaller neighborhoods around training samples while farther away larger regions appear and cause the transient behavior in the curves.}
    \label{fig:complexity_area}
\end{figure}

\section{Implementation Details}
\label{appendix:implementation}

In Sec.~\ref{sec:method} we provide a summary of how SplineCam works. In this section we provide details about SplineCam implementation and algorithm.

SplineCam is implemented using Pytorch \cite{pytorch} and Graph-tool \cite{peixoto_graph-tool_2014}. All the linear algebra operations are performed using Pytorch and are scalable using GPUs. The region finding operation on the other hand is single threaded. This can be a bottleneck in cases, for example, with DNNs that have more than one layer. In this case, distributing regions across threads, as pairs of hyperplane-sets and a polygon, can result in significant speedups.

Since finding the regions involve solving systems of linear equations, most of the opearations in SplineCam require \textit{double}
precision. This can introduce significant memory bottlenecks, especially for large convolutional layers with multiple channels of input and output; for such layers the size of the corresponding Toeplitz matrix representation becomes significantly large with large number of input and output channels. For this reason, we always store such weight matrices as sparse matrices and query rows only when required. We also replace max pool layers with average pool layers for simplicity; in our experiments involving VGG16 and VGG11, we replace the maxpool after the first conv with an average pool and use strided convolutions for deeper layers. Unless specified, we always consider square 2D domains in the input. While characterizing we use the terms volume and area of the regions interchangeably.

\begin{figure*}[t!]
    \centering
    \includegraphics[width=.9\linewidth]{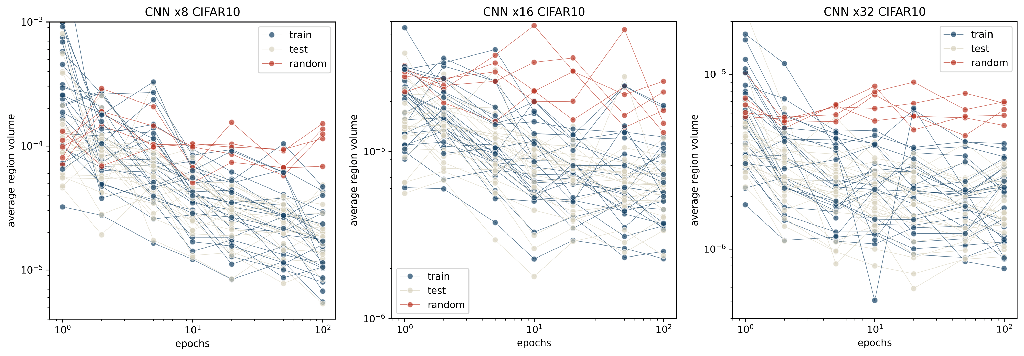}
    \caption{Evolution of average region volume (ARV) for $55$ different randomly oriented square domains of the input space. Out of the $55$ domains, $25$ are centered on CIFAR10 training samples, $25$ on test samples and $5$ on random locations in the input space. We train a $8$ layer CNN as detailed in Suppl.~\ref{appendix:evolution}. We notice that as training progresses, the ARV around points on the data manifold gets reduced. For points off the manifold ARV reduces as well, for CNN x8 CIFAR10 it reduces significantly while for CNN x32 it doesn't. In all cases, the off manifold ARV remains larger than the on manifold ARV at 100 training epochs.}
    \label{fig:cnn_withrand}
\end{figure*}

In Suppl.~\ref{appendix:usage}, we provide details on how SplineCam can be used, with modular codes showing how it computes the partition in a layerwise fashion. We also provide a pseudocode for the search algorithm we have proposed to find regions given a graph formed via polygon-hyperplane intersections. All the reported computation times were evaluated on a setup consisting of 8x NVIDIA QUADRO RTX 8000 48GB GPUs and 2x Intel Xeon Gold 5220R.

\section{Computational Complexity}
\label{appendix:complexity}

In this section we present two sets of experiments that we have performed to assess the computational complexity of SplineCam.

\textbf{Varying width for a single layer MLP.} As we have discussed earlier and have elaborated in Suppl.~\ref{appendix:usage}, SplineCam performs region wise partition operations, which can be parallelized across sets of regions. To assess the per region performance of SplineCam, we do the following experiment. We take an uninitialized 1 layer ReLU-MLP with $n$ neurons and $D$ input dimensionality. The number of neurons is equal to the number of hyperplanes that would partition space. We vary $n \in \{10...980\}$ and also vary $D \in \{2,2002,4002,6002,8002\}$ and plot in Fig.~\ref{fig:complexity_singleMLPlayer} the wall time in seconds, required to compute the partition. As the input domain, we consider a randomly oriented 2D domain centered on the origin with $4sq$ unit area. We see that the computation time complexity is upper bounded by $\mathcal{O}(n^2/5000)$ within the range of $n$ in question. With increasing $D$ we see a reduced number of hyperplane intersections, therefore we see a reduction in required time.

\textbf{Effect of the area of the input domain.} For this experiment, we take a VGG11 model and increase the size (area) of the input domain. We present the wall time vs area plot in Fig.~\ref{fig:complexity_area}. We can see that increasing the area (or volume) of the input region, monotonically increases the required time for computing the partition, in approximately a linear fashion. Even though increasing the area of the input domain should increase the number of first layer intersections, we see that the effect of that remains linear within the range. Note that, we can also scale the area of the input domain by breaking the input domain into multiple 2D polygonal regions of equal area and using separate threads/gpus to perform computation. This way we can also parallelize the partition computation and scale across memory instead of time. 

\section{Extra Experiments}
\label{appendix:extra}

\subsection{Evolution of partition statistics while training}
\label{appendix:evolution}
\begin{figure}[!t]
    \centering
    Train \hspace{8em} Test
    \includegraphics[width=.9\linewidth]{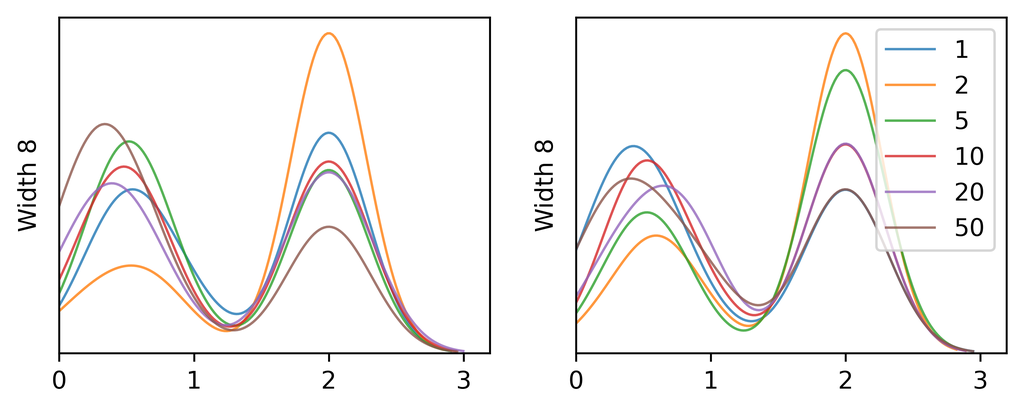}
    \includegraphics[width=.9\linewidth]{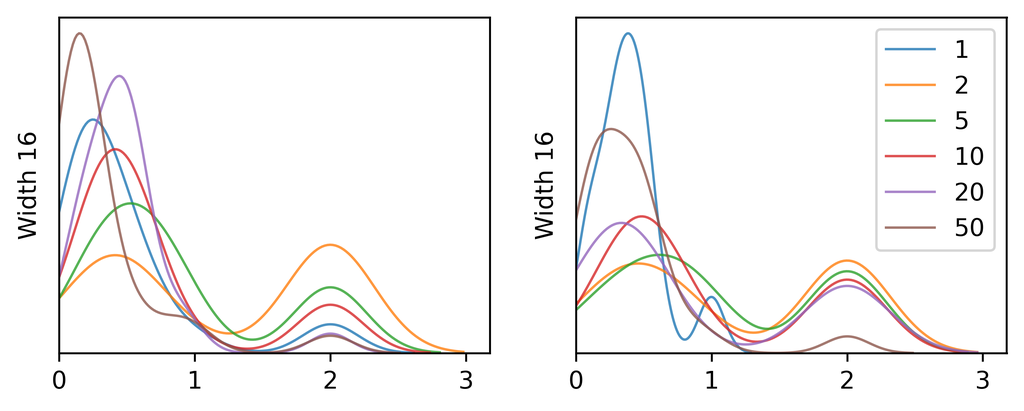}
    \includegraphics[width=.9\linewidth]{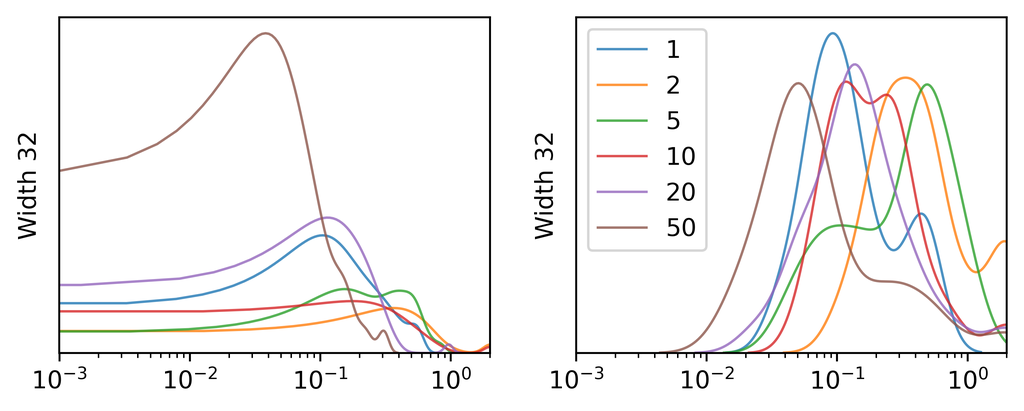}
    \includegraphics[width=.9\linewidth]{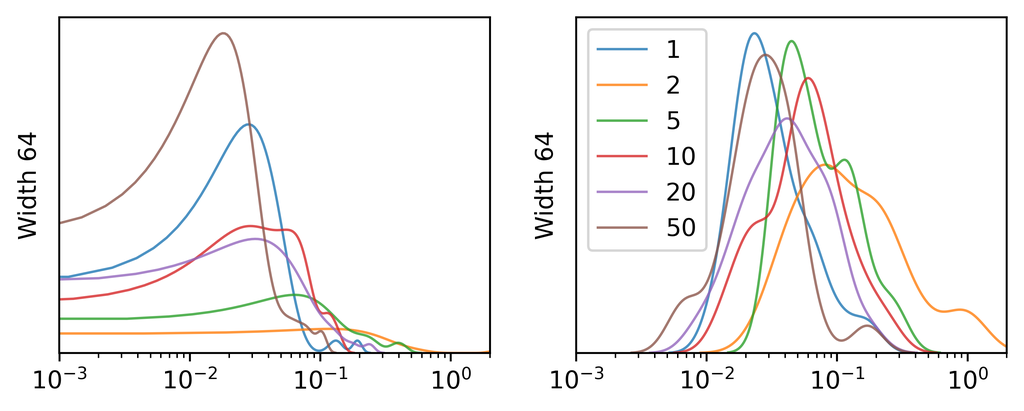}
    \includegraphics[width=.9\linewidth]{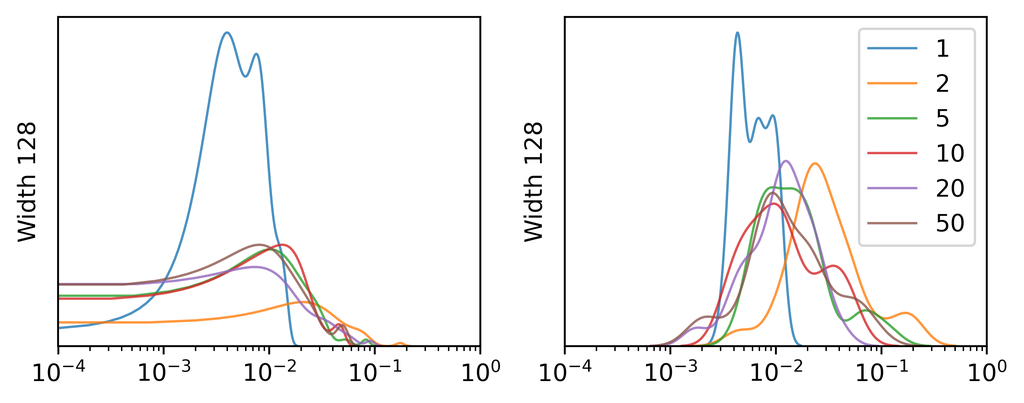}
    \caption{Distribution of \textit{average region volume} (ARV) across training epochs, in the neighborhood of $25$ train \textbf{(Left)} and $25$ test \textbf{(Right)} samples from MNIST. We train an MLP with depth $5$ and vary its width between $\{8,16,32,16,128\}$. ARV is considerably large and bimodal for smaller widths; as network width is increased ARV becomes smaller and unimodal, with long smaller volume tails for training samples. This discrepancy between training and test samples, possibly indicates memorization.}
    \label{fig:mnist_mlp_sweep_ARV}
\end{figure}

\begin{figure}[!t]
    \centering
    Train \hspace{8em} Test
    \includegraphics[width=.9\linewidth]{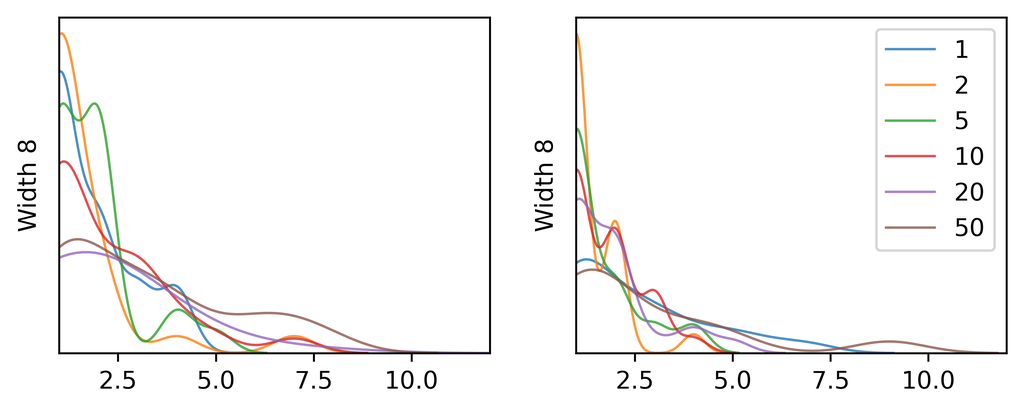}
    \includegraphics[width=.9\linewidth]{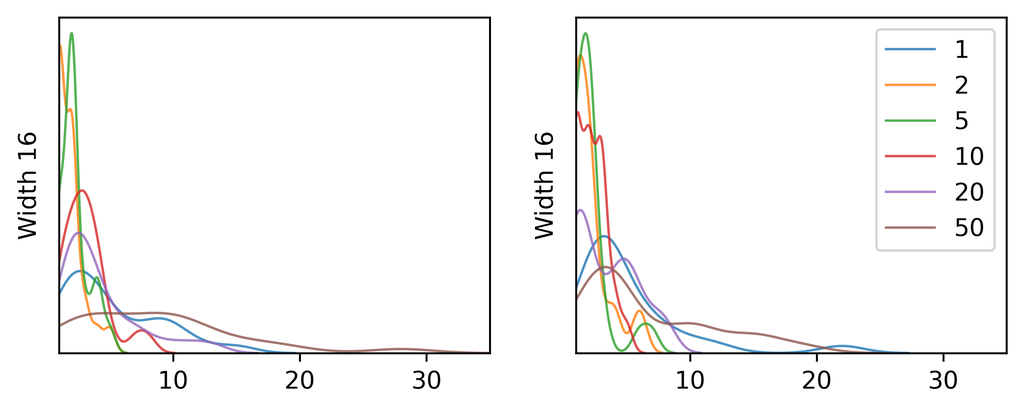}
    \includegraphics[width=.9\linewidth]{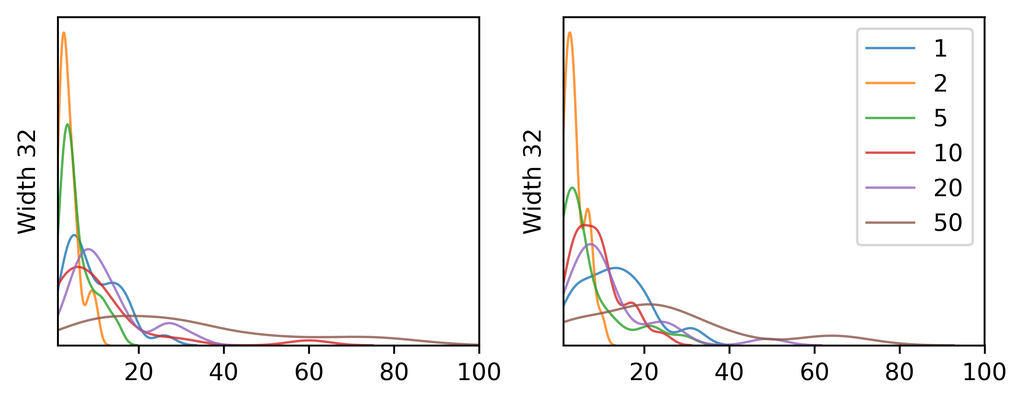}
    \includegraphics[width=.9\linewidth]{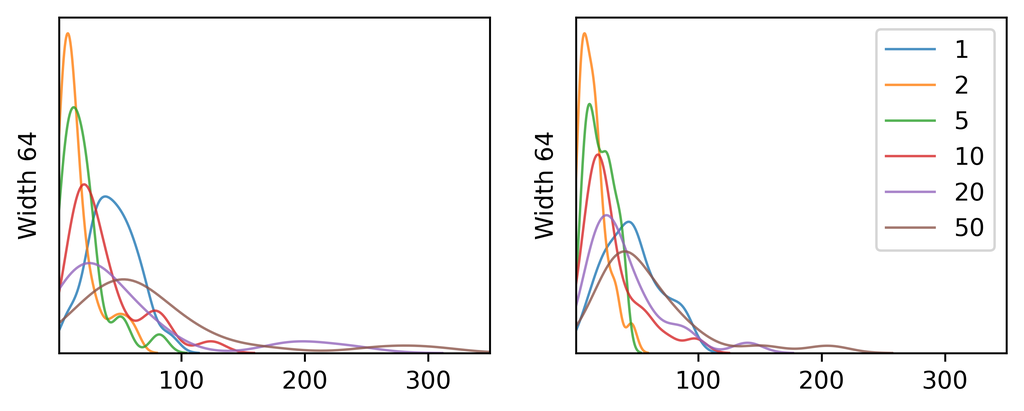}
    \includegraphics[width=.9\linewidth]{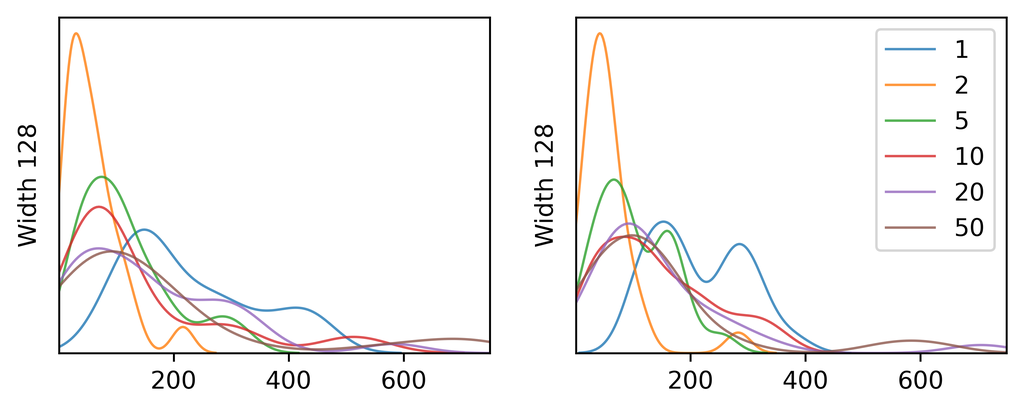}
    \caption{Distribution of \textit{number of regions} (NR) across training epochs, in the neighborhood of $25$ train \textbf{(Left)} and $25$ test \textbf{(Right)} samples from MNIST. We train an MLP with depth $5$ and vary its width between $\{8,16,32,16,128\}$. NR is small for smaller widths and increases significantly as the network width is increased. For larger networks, the distributions have large NR tails. We can also see a shift towards lower NR right after epoch 1 and a slow shift of the distribution towards larger NR as training progresses.}
    \label{fig:mnist_mlp_sweep_anr}
\end{figure}

\textbf{MLP trained on MNIST.} For this experiment we train an MLP with depth $5$ and width $\in \{8,16,32,16,128\}$. We train the MLPs for $50$ training epochs on the MNIST dataset, and evaluate the partition statistics via SplineCam, for $25$ training and $25$ validation samples of randomly selected from MNIST. We present average region volume (ARV) distributions per training epoch in Fig.~\ref{fig:mnist_mlp_sweep_ARV}. The first thing to notice is that for smaller width networks, ARV is bimodal across all epochs, while for width 64 and onwards, the mode with higher volume regions vanishes. ARV also shifts towards lower volumes as the width of the networks are increased. While training samples tend to have lower ARV, the lower ends of the distributions differ between training and test samples; for widths 32, 64, and 128 we see distinct low ARV tails which are not visible for the test samples. This shows that for some of the training samples, the partition regions of the network are smaller, indicating that the model has more representation capacity for such sample neighborhoods \cite{montufar2014number}. This could be a possible indication for memorization of some training samples. Another thing to notice is that during the first epoch, the avg partition volumes are significantly lower. As training progresses, first ARV shifts to the right (larger) and then slowly shifts to the left. As we increase width, the starting ARV of the network becomes small in general compared to the ARV for the last training epoch. In Fig.~\ref{fig:mnist_mlp_sweep_anr} we also present the distributions of the number of regions (NR) in the neighborhood of the same samples used for Fig.~\ref{fig:mnist_mlp_sweep_ARV}.

\begin{figure}[!t]
    \centering
    \includegraphics[width=\linewidth]{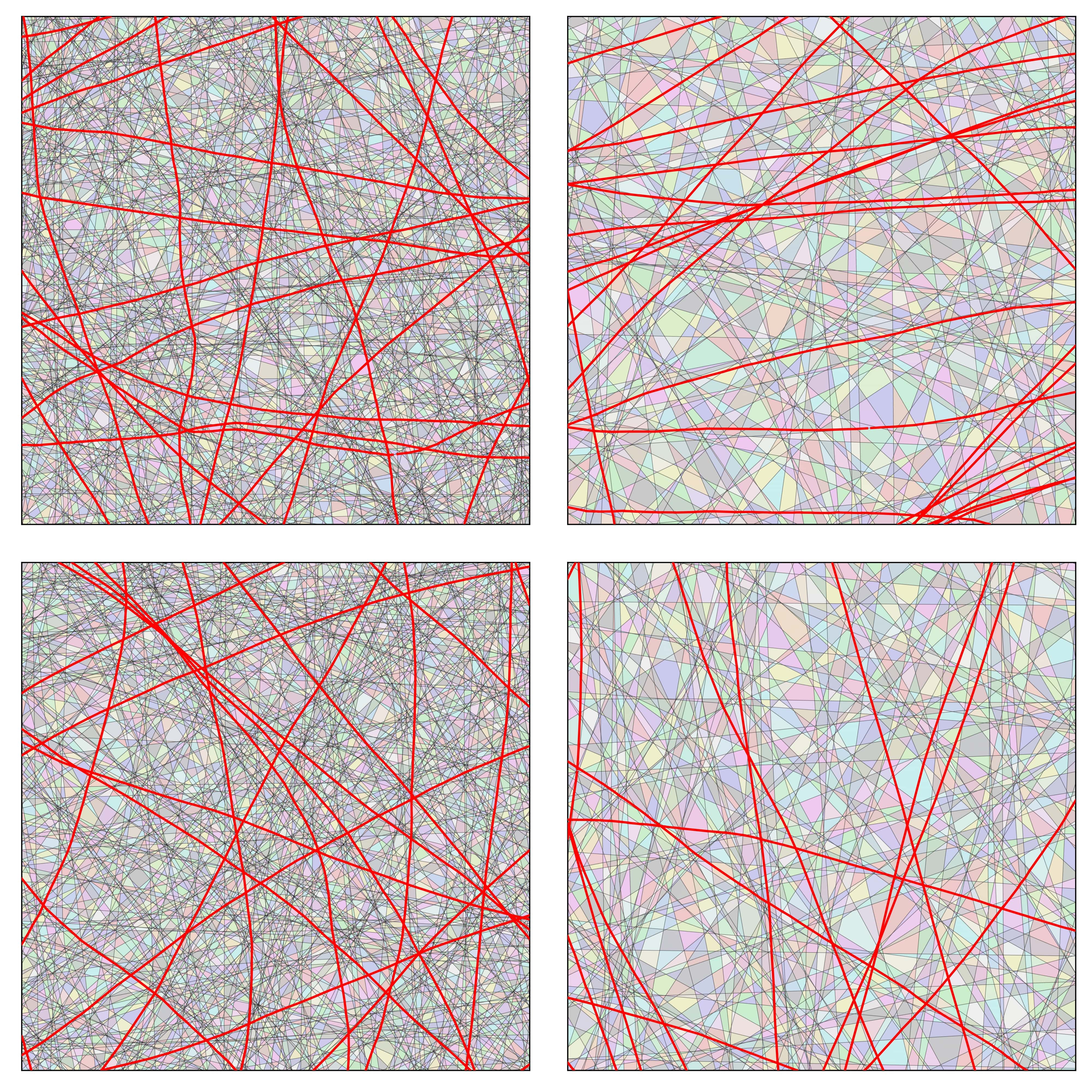}
    \caption{Partition visualization for a VGG11 model trained on TinyImagenet. We take 10 samples from the training set for which the model posteriors have the lowest entropy, and present here neighborhoods for two samples with high partition density \textbf{(Left)} and two samples with low partition density \textbf{(Right)}. We also highlight in red, the sets of points for which any two classes from the dataset has equal probability. We see that the denser partition regions have more such lines compared to sparser regions, suggesting correlations with generalization \cite{raghu2016expressive}}
    \label{fig:VGG11_dec_bound}
\end{figure}

\textbf{CNN trained on CIFAR10.} For this experiment, we train an $8$ layer CNN with $6$ convolutional layers and $2$ fully connected layers. The number of filters for the convolutional layers are set as $\{\lfloor \ell/2 \rfloor \times mul\ : \ell = 1...6\}$, where $\lfloor . \rfloor$ is the floor operation, and $mul \in \{8,16,32\}$ is a width multiplier. We see that, similar to Fig.~\ref{fig:mnist_mlp_sweep_ARV}, the ARV gets reduced with increased width. For training, we can see longer tails towards lower ARV, indicating denser regions near some training samples. One thing that is noticeable here is that contrary to MLPs, the ARV of neighborhoods near training samples monotonically decrease in most cases for the CNNs. This could be due to the complexity of the task, CIFAR10 classification being a harder task compared to MNIST, the region density is required to be significantly higher compared to early training.

We also visualize in Fig.~\ref{fig:cnn_withrand} the evolution of ARV with training epochs for CNNs with different width multipliers. We visualize for $25$ training, $25$ test and $5$ random samples, a randomly oriented 2D neighborhood. We notice that as training progresses, the ARV around points on the data manifold gets reduced. For points off the manifold ARV reduces as well, for CNN x8 CIFAR10 it reduces significantly while it doesn't as much for CNN x32. In all cases, the off manifold ARV remains larger than the on manifold ARV at 100 training epochs.

\subsection{Variation of region statistics between random orientations}
\label{appendix:random_orientation}
For this experiment we take 5 random training samples from TinyImagenet and calculate partition statistics for 20 randomly oriented 2D domains with area $0.005$, centered on each sample. To assess the variability between different 2D domains we first look at the region volume (RV) statistic for the partition generated by a VGG11 model. Region volumes can vary both for a given 2D input domain and between different input domains. The maximum in-domain RV standard deviation over 20 different orientations is $\{7.3955e-07, 2.2665e-07, 3.0617e-06, 1.0149e-06, 2.2171e-06\}$ for each sample. The between orientation ARV standard deviation is $\{5.5993e-08, 7.4666e-09, 1.3462e-07, 3.9948e-08, 1.2935e-07\}$ for each sample, which is an order smaller than the in-domain RV standard deviation. This is an indication that SplineCam statistics for a single 2D slice could possibly be accurate enough to not require multiple 2D slices, even for high dimensional inputs.

\subsection{Extra Figures}
In the following section we present some figures complementing the experiments done above.

\clearpage
\newpage

\clearpage
\newpage

\begin{figure}[!t]
    \centering
    Train \hspace{8em} Test
    \includegraphics[width=.9\linewidth]{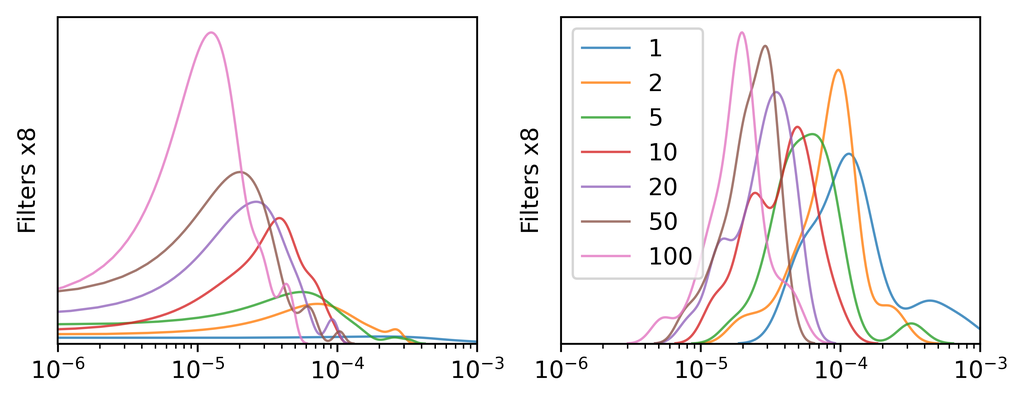}
    \includegraphics[width=.9\linewidth]{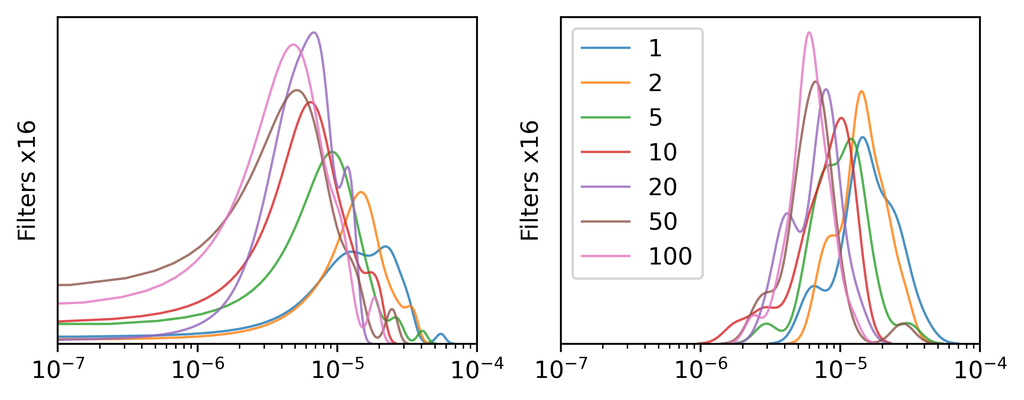}
    \includegraphics[width=.9\linewidth]{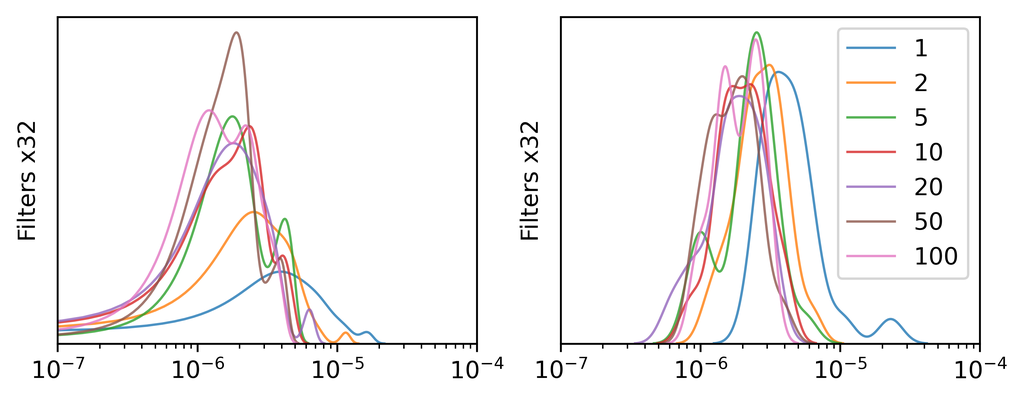}
    \caption{Distribution of \textit{average region volume} (ARV) across training epochs, in the neighborhood of $25$ train \textbf{(Left)} and $25$ test \textbf{(Right)} samples from CIFAR10. We train a CNN with $6$ convolutional layers and $2$ fully connected layers. The number of filters for the layers are set as $\{\lfloor \ell/2 \rfloor \times mul\ : \ell = 1...6\}$, where $\lfloor . \rfloor$ is the floor operation, and $mul \in \{8,16,32\}$ is a width multiplier. We see that, similar to Fig.~\ref{fig:mnist_mlp_sweep_ARV}, the ARV gets reduced with increased width. For training, we can see longer tails towards lower ARV, indicating that for some training samples the regions become very small. For both train and test, with increasing number of epochs, the ARV distribution mean shifts towards lower ARV.}
    \label{fig:cifar_cnn_sweep_ARV}
\end{figure}

\begin{figure}[!t]
    \centering
    Train \hspace{8em} Test
    \includegraphics[width=.9\linewidth]{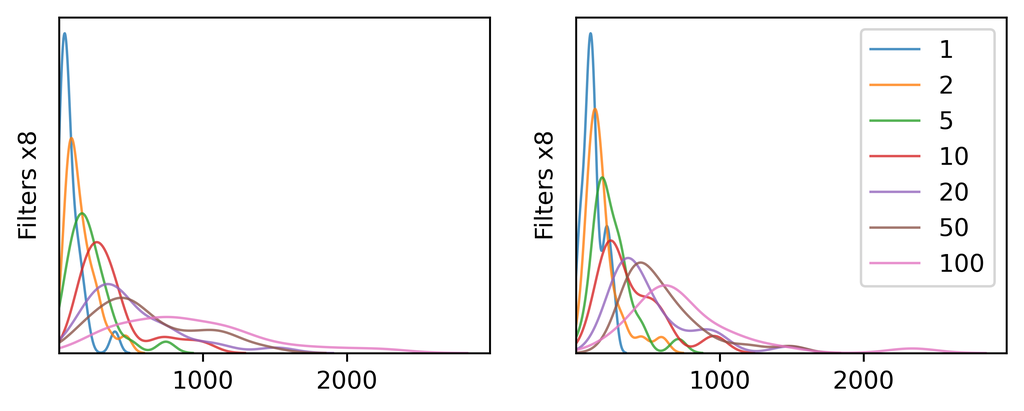}
    \includegraphics[width=.9\linewidth]{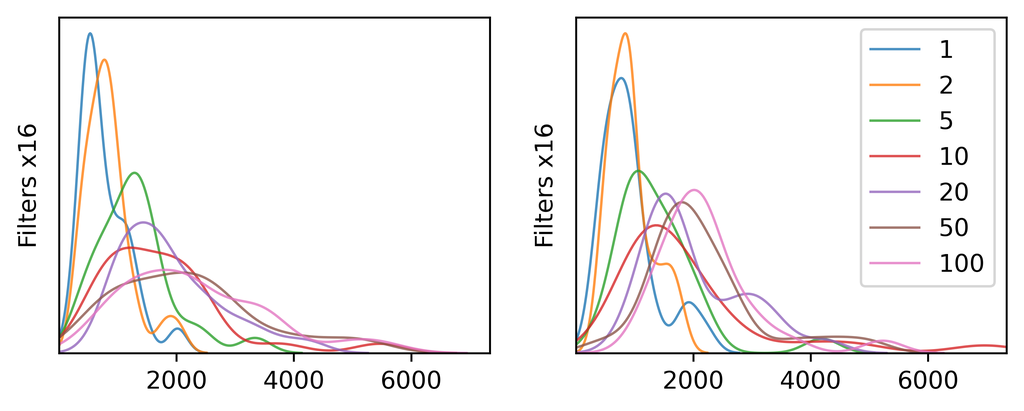}
    \includegraphics[width=.9\linewidth]{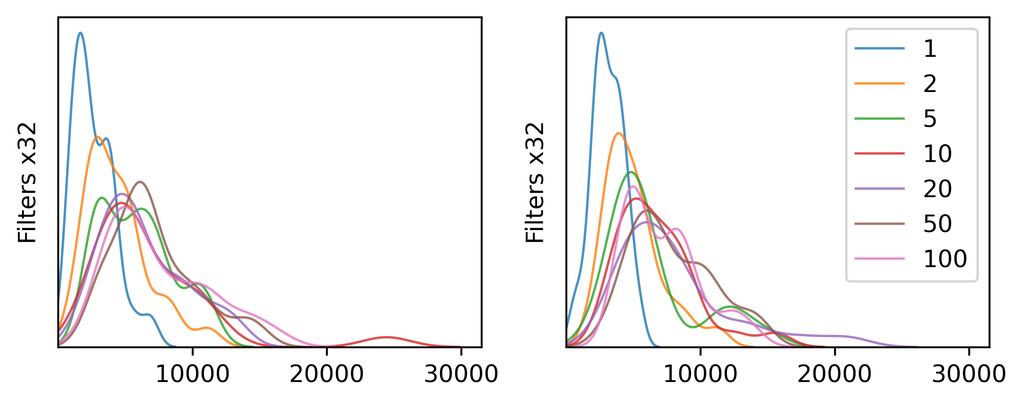}
    \caption{Distribution of \textit{number of regions} (NR) across training epochs, in the neighborhood of $25$ train \textbf{(Left)} and $25$ test \textbf{(Right)} samples from CIFAR10. We train a CNN with $6$ convolutional layers and $2$ fully connected layers. The number of filters for the layers are set as $\{\lfloor \ell/2 \rfloor \times mul\ : \ell = 1...6\}$, where $\lfloor . \rfloor$ is the floor operation, and $mul \in \{8,16,32\}$ is a width multiplier. We see that, similar to Fig.~\ref{fig:mnist_mlp_sweep_anr}, the NR significantly increases with increased width. For training, we can see longer tails towards higher NR, indicating that for some training samples there is high region density in the neighborhood. For both train and test, with increasing number of epochs, the NR distribution mean shifts towards higher NR.}
    \label{fig:cnn_cifar_sweep_anr}
\end{figure}

\clearpage
\newpage

\begin{figure*}
    \centering
    \includegraphics[width=\linewidth]{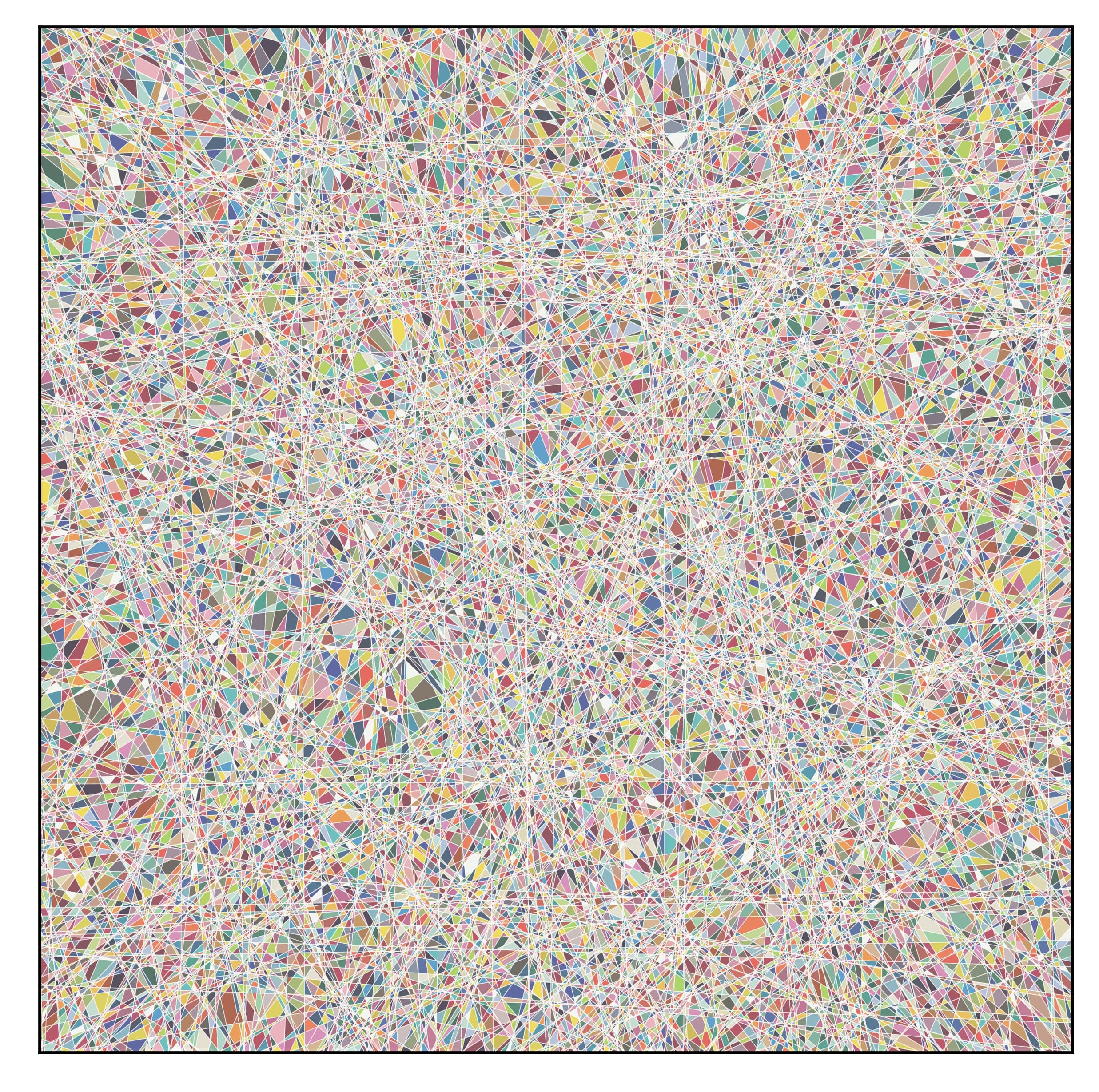}
    \caption{Partition visualization of a randomly initialized single layer MLP with 1000 hyperplanes and input dimensionality of 8002 for a randomly oriented 2D square domain centered on the origin. The partition contains $132569$ regions and takes $134s$ to compute.}
    \label{fig:mlp_partition}
\end{figure*}

\clearpage
\newpage

\begin{figure*}
    \centering
    \includegraphics[width=\linewidth]{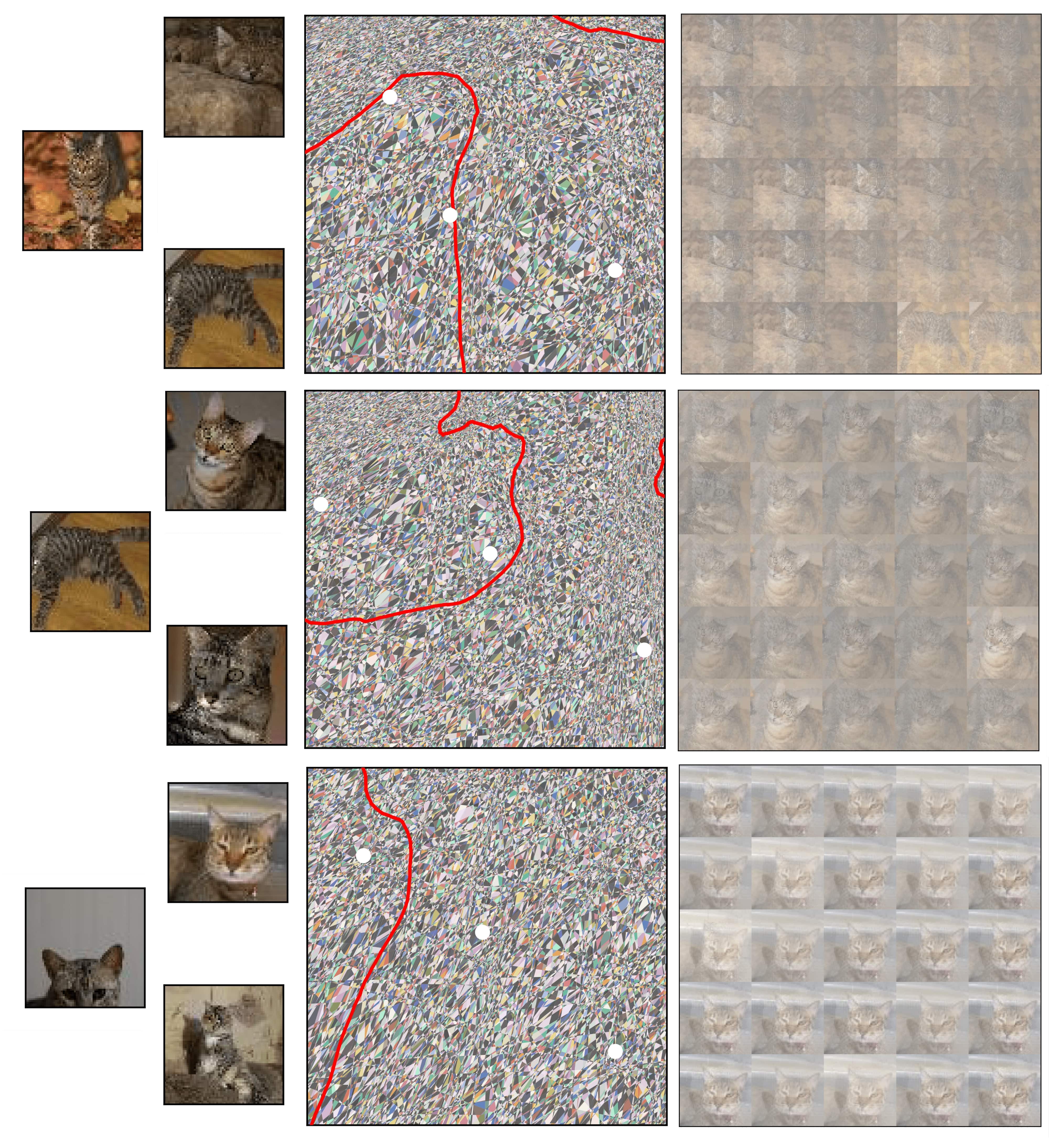}
    \caption{Partition visualization of a convolutional neural network, trained for binary classification of tinyimagenet Tabby Cat and Egyptian Cat classes. \textbf{Left} Samples that are used as anchor points to determine the 2D slice with an area of $450$ units. \textbf{Middle} The partitioning of input space induced by the model as well as the decision boundary (in red). \textbf{Right} Randomly sampled points from the decision boundary. Samples from the decision boundary visually represent a linear combination of the three anchor samples, while the weights are determined by the non-linear decision boundary. For example, for the top and bottom rows, samples from the boundary look biased towards two of the three anchor points. Computing the partition regions take 7.46 mins, 13.96 mins and 5.02 mins respectively, with each of the partitions containing 82817, 119895, and 60455 regions. The number of regions is positively correlated with the curvature of the decision boundary in the neighborhood.}
    \label{fig:cat_partition}
\end{figure*}

\begin{figure*}
    \centering
    \includegraphics[width=.9\linewidth]{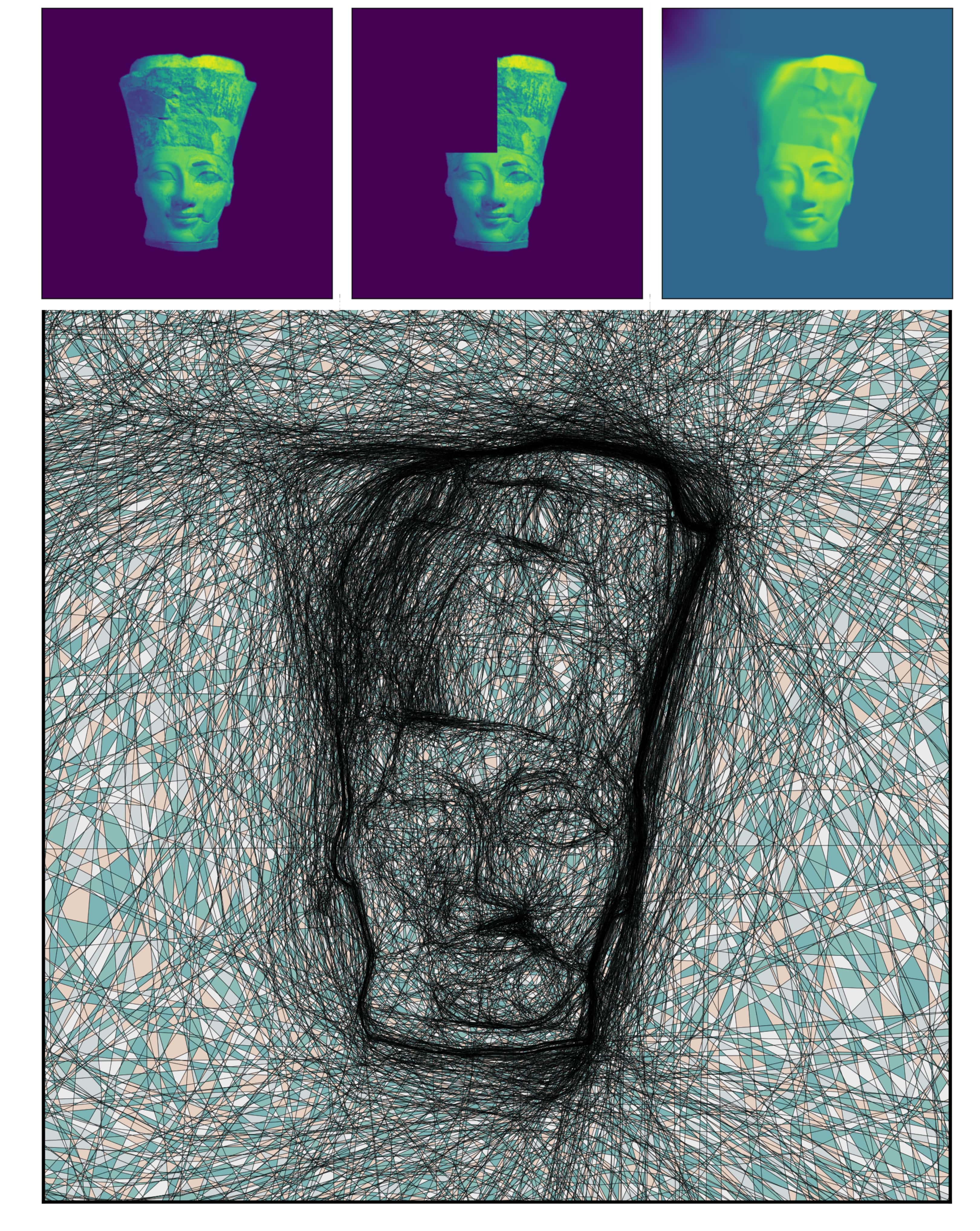}
    \caption{Training and visualizing the partition of a 2D INR trained on an inpainting task. Top row shows the original training image, the training image with a section cropped out and the predictions of a trained MLP with width 256 and depth 6. The MLP, via implicit regularization, learns a smooth surface in place of the discontinuity. Note how a lot of neurons are placed in the cropped region, and also around the foreground boundaries, to allow more curvature while fitting.}
    \label{fig:inpaint}
\end{figure*}

\clearpage
\newpage

\section{Usage of SplineCam}
\label{appendix:usage}

We provide SplineCam as a python toolbox that can wrap any given Pytorch \cite{pytorch} \textit{sequential} network, containing a set of supported modules. 

To begin, first we have to define a 2D input space region of interest (ROI). The region of interest is can be a polytopal region at the input of the network defined via vertices. Since all the region finding operations are performed in 2D, we also require an orthogonal projection matrix that projects vectors from the input space on to the 2D ROI. Following this we can use the SplineCam library to wrap a given model. 

\begin{lstlisting}[language=Python,
caption=Wrapping a model with splinecam,
label=code:wrapper]

import torch
import splinecam

## given torch model and domain (ROI) as a list of vertices

T = splinecam.utils.get_proj_mat(domain)

model.cuda()
model.eval()
model.type(torch.float64)

print('Wrapping model with SplineCam...')
NN = splinecam.wrappers.model_wrapper(
    model,
    input_shape=model.input_shape,
    T = T,
    dtype = torch.float64,
    device = 'cuda'
)

## check .forward() and matmul operation equivalence
print('Verifying wrapped model...')
flag =  NN.verify()
print('Model.forward and matmul equivalence check', flag)
assert flag

#
\end{lstlisting}

SplineCam supports custom layers as well. Each SplineCam layer requires a submodules to return the weights, the intersection pattern and activation pattern of the layer. We refer the reader to our codes for details. The wrapped SplineCam model contains a verification method, to ensure that the affine operations and the forward operations (which can be different from the affine operation based on implementation, e.g., convolution) of the model result in the same value for random inputs.

SplineCam can take any set of  2D domains as ROI, with corresponding projection matrices. This allows SplineCam to be used to visualize the partition for piecewise linear subspaces in the input space. The following example shows how SplineCam computes the partition in a layerwise fashion.

\begin{lstlisting}[language=Python,
caption=Modular code for computing spline partition,
label=code:wrapper2]

## for a given list of polygons in 2D and corresponding projection matrices

Abw = T
out_cyc = poly

for current_layer in range(1,len(NN.layers)):
    
    ## given a set of 2D regions and a target layer, find all new regions in 2D
    out_cyc,out_idx = splinecam.graph.to_next_layer_partition(
        cycles = out_cyc,
        Abw = Abw,
        NN = NN,
        current_layer = current_layer,
    )
    
    ## acquire region centroids
    means = splinecam.utils.get_region_means(out_cyc)
    
    ## pass each region centroid to next layer
    means = NN.layers[:current_layer].forward(means)
    
    ## get activation mask for each region    
    q = NN.layers[current_layer].get_activation_pattern(means)

    ## query network weights
    Wb =  NN.layers[current_layer].get_weights()

    ## calculate affine parameters per region
    Abw = splinecam.utils.get_Abw(
            q = q,
            Wb = Wb.to_dense(),
            incoming_Abw = Abw) 

#

\end{lstlisting}

One of the key algorithms in the \texttt{to\_next\_layer\_partition(.)} function is the search algorithm that allows us to find cycles from a given graph. The following codeblock presents a pseudocode of our heuristic breadth first search method.

\begin{lstlisting}[language=Python,
caption=Pseudocode function for finding cycles given a undirected graph,
label=code:wrapper3]

from graph_tool import topology

def find_cycles(V=input_graph,start_edge=input_edge):
    '''
    Given a undirected graph and a starting edge find cycles from that edge
    '''
    
    ## Convert V to a bidirectional graph. This allows us to control
    ## the number of traversals for each edge
    V = convert_to_bidirectional(V)
    edge_q.append(start_edge)
    
    ## if edge is a boundary edge
    V.remove_edge(start_edge)
    
    out_cycles = []  
    
    for e in edge_q:
        
        remove_q = []
        vertices = []
        vertex_id = []
        
        ## if no way out of v0 or no way in for v1, continue
        if not (V.get_in_degrees(e.vertex1)>1
            ) and (
            V.get_out_degrees(e.vertex0)>1):
            continue
        
        ## if the edge doesn't exist, continue 
        if V.edge(e) is None and V.edge(e.vertex1,e.vertex0) is None:
            continue
      
        ## add opposite path to removal queue
        remove_q.append(V.edge(e.vertex1,
                    e.vertex0))) 
        
        ## bfs
        vs,es = topology.shortest_path(V,
                 source=e.vertex0,
                 target=e.vertex1,
                )
                            
        out_cycles.append([V.vertex_index[each] for each in vs])
   
        for new_e in es:
        
            ## if boundary edge remove edges in both directions
            if V.ep['layer'][new_e] == -1:
                remove_q.append(new_e)
                remove_q.append(V.edge(new_e.vertex1,new_e.vertex0))
            
            else:
                ## remove only one direction and append to queue
                remove_q.append(new_e)
                edge_q.append(new_e)
        
        for each in remove_q:
            V.remove_edge(each)

    return out_cycles

\end{lstlisting}

\end{document}